\documentclass{article}
\pdfoutput=1  %

\newif\ifarxiv
\arxivtrue
\arxivtrue

\ifarxiv
  \usepackage[preprint]{neurips_2026}
\else
  \usepackage{neurips_2026}
\fi

\usepackage[utf8]{inputenc}
\usepackage[T1]{fontenc}
\usepackage{amsmath,amssymb,amsfonts}
\usepackage{amsthm}
\usepackage{mathtools}
\usepackage{algorithm}
\usepackage[noEnd]{algpseudocodex}
\usepackage[inline]{enumitem}
\usepackage{booktabs}
\usepackage{tabularx}
\usepackage{colortbl}
\usepackage{multirow}
\usepackage{graphicx}
\usepackage{hyperref}
\usepackage{url}
\usepackage{microtype}
\usepackage{xcolor}
\IfFileExists{.minted-finalize}{%
  \usepackage[finalizecache,cachedir=minted-cache]{minted}%
}{%
  \ifarxiv
    \usepackage[frozencache,cachedir=minted-cache]{minted}%
  \else
    \usepackage[cache=true,cachedir=minted-cache]{minted}%
  \fi
}
\usepackage{tikz}
\usepackage{pgfplots}
\pgfplotsset{compat=1.18}
\usepackage{marginalia}
\usepackage{needspace}

\definecolor{lstbg}{RGB}{248,248,250}
\setminted{
  style=tango,
  fontsize=\scriptsize,
  breaklines=true,
  breakanywhere=true,
  linenos=true,
  numbersep=3pt,
  xleftmargin=1.2em,
  numberblanklines=false,
  frame=none,
  bgcolor=lstbg,
  autogobble,
  baselinestretch=1.0,
}

\usetikzlibrary{positioning, arrows.meta, shapes, fit, backgrounds, patterns}
\usepackage{caption}
\usepackage{cleveref}
\let\etoolboxforlistloop\forlistloop
\usepackage{autonum}
\let\forlistloop\etoolboxforlistloop
\makeatletter
\autonum@generatePatchedReferenceCSL{Cref}
\makeatother
\crefname{listing}{Listing}{Listings}
\Crefname{listing}{Listing}{Listings}
\crefname{appendix}{Appendix}{Appendices}
\Crefname{appendix}{Appendix}{Appendices}
\crefname{ALG@line}{line}{lines}
\Crefname{ALG@line}{Line}{Lines}

\usepackage{wrapfig}

\newif\ifcompact
\compacttrue  %

\usepackage{algmarginfix}

\title{Archimedean Copula Inference via Taylor-Mode AD}

\author{
  Cambridge Yang\\
  Independent\\
  \texttt{research@camyang.com}
  \And
  Dongdong Li\\
  Harvard Medical School\\
  \texttt{dongdongli@hsph.harvard.edu}
}

\begin{document}
\raggedbottom

\maketitle

\addtocontents{toc}{\protect\setcounter{tocdepth}{-1}}

\begin{abstract}
  No existing nested Archimedean copula tool handles all three of
  (a) arbitrary per-variable (right-)censoring in survival analysis,
  (b) arbitrary nesting trees, and (c) exact parameter gradients.
  Existing implementations handle only bivariate problems,
  low dimensional (i.e., $d \leq 10$) cases,
  two layers of nesting, or only hand-derived copula nestings.
  We present \textsc{acopula}, a JAX-native framework that, given any
  Archimedean generator---classical or neural---evaluates exact
  nested-copula likelihoods and parameter gradients under arbitrary
  censoring masks in polynomial time.
  The mechanism is polynomial powering of Taylor-mode automatic
  differentiation output, which replaces per-family hand-derived
  partial Bell polynomial tables with a single differentiable
  computation that any user-defined generator can drive.
  We conduct extensive simulations to verify the correctness of \textsc{acopula}.
  We then demonstrate
  (a) per-variable censoring on $85{,}229$ MIMIC-IV ICU admissions
  in high dimensions with $d{=}53$, fit by both classical Archimedean families and
  nested neural Archimedean copulas;
  (b) an 11-sector hierarchical model on S\&P~500 daily returns at
  $d{=}98$;
  (c) family-agnostic censored MLE across ten families, five of them
  with no prior implementation, on a retinopathy study; and
  (d) a ${\sim}650\times$ per-density speedup over R's \texttt{nacLL}
  at $d{=}35$, scaling quadratically to $d{=}8{,}000$.
\end{abstract}

\section{Introduction}
\label{sec:intro}

Hierarchical multivariate data with partial observation is the
default in modern survival, financial, and clinical risk settings.
Patients in a survival study with paired organs may never experience
the event during follow-up~\citep{huster1989modelling}.
Stock returns cluster by industry sector and tail-co-move during
crashes.
Intensive care unit (ICU) lab measurements group by clinical panel,
and the analyst right-censors each lab independently at hospital
discharge~\citep{johnson2023mimic}.
Each setting demands a model that captures the hierarchy with
interpretable parameters---tail dependence and concordance---and
accommodates an arbitrary censoring pattern per observation.

\paragraph{Archimedean copulas.}
A {\em copula} $C: [0,1]^d \to [0,1]$ is a multivariate distribution
on the unit cube with $d$ uniform marginals; by Sklar's
theorem~\citep{sklar1959fonctions}, every joint distribution on
$\mathbf{u} = (u_1, \ldots, u_d) \in [0,1]^d$ factors uniquely into
its one-dimensional marginals and a copula that captures their
dependence.
This decomposition lets the analyst pair domain-specific marginal
models---e.g., Weibull~\citep{weibull1951statistical} for survival times,
GARCH~\citep{bollerslev1986garch} for returns---with an interpretable
dependence structure whose parameters directly encode tail behaviour
and concordance.
      {\em Nested Archimedean copulas}~\citep{mcneil2008sampling,hofert2012density}
extend copulas to hierarchical dependence by composing generator
functions in a tree: variables under the same internal node share
stronger dependence than variables across branches.

\paragraph{The inference barrier.}
Fitting a nested copula requires its density---a mixed
partial derivative of the copula cumulative distribution function (CDF), one differentiation per observed
variable---through a tree of composed
generators~\citep{hofert2012density}; three obstacles block existing tools.
\begin{itemize}[leftmargin=*]
      \item {\em Per-variable censoring.}
            We consider per-variable right-censoring under the survival setting.
            Each subset of censored variables changes which mixed partial
            derivative of the copula CDF defines the likelihood, so a $d{=}53$
            ICU dataset with thousands of distinct censoring patterns
            requires the framework to handle every per-observation
            mask $\delta \in \{0,1\}^d$ uniformly, where $\delta_j = 1$
            marks variable $j$ as observed.
            Prior tools cover bivariate problems~\citep{sun2020copulacenr,emura2018analysis}
            or shared censoring~\citep{liu2024hacsurv}; none reach $d > 30$ per-variable.
      \item {\em Arbitrary nesting and any generator.}
            R's \texttt{copula} package~\citep{hofert2011sampling}
            hard-codes per-family Stirling-number tables for five
            generators and supports only two nesting layers; adding a
            sixth family---or any neural-based generator such as the Deep
            Archimedean Copula family~\citep{ling2020deep,liu2024hacsurv}---demands fresh hand derivation.
      \item {\em Exact automatic parameter gradients.}
            Without automatic gradients, fitting a $d{=}98$ model relies on
            poorly-scaling derivative-free or finite-difference methods,
            or Monte Carlo approximators~\citep{hofert2012estimators,ng2021generativearchimedean} that are consistent but biased with finite samples.
\end{itemize}

\paragraph{Our approach.}
We present \textsc{acopula}, a JAX-native framework that resolves all
three obstacles.
The user specifies at minimum a copula generator---classical or
neural---and \textsc{acopula} computes the nested copula likelihood
for any tree shape and any censoring mask, with exact parameter
gradients automatically.
The mechanism is polynomial powering of Taylor coefficients:
Taylor-mode AD~\citep{griewank2008evaluating,radul2020you} returns
the higher-order derivatives of each generator composition in one
forward pass, and raising those coefficient vectors to successive
powers via convolution recovers the partial Bell polynomials that
the density formula of \citet{hofert2012density} demands without
per-family hand derivation, for any generator satisfying the standard
$d_c$-monotonicity nesting condition~\citep{mcneil2008sampling}.

\paragraph{Contributions.}
We make the following contributions:
\begin{enumerate}[leftmargin=*]
      \item A {\em family-agnostic, gradient-based likelihood inference algorithm} for
            nested Archimedean copulas: given only a classical or
            neural generator (e.g., {\em Deep Archimedean Copulas}%
            ~\citep{ling2020deep,liu2024hacsurv}), the framework
            derives densities and gradients automatically.
      \item {\em Per-variable censoring} as a first-class operation:
            the likelihood algorithm handles arbitrary per-observation
            masks $\delta \in \{0,1\}^d$ directly through its core
            recurrence, lifting the bivariate-only or
            single-shared-time restrictions of prior censoring-aware
            tools.
      \item A {\em complexity result}: $O(d^3)$ worst-case for fixed-depth
            trees and $O(d^2)$ for $\sqrt{d}${\em -decomposed} fixed-depth trees, where
            every non-root subtree holds at most $O(\sqrt{d})$ uncensored
            leaves; validated empirically up to $d{=}8{,}000$.
      \item An {\em open-source JAX implementation}, \textsc{acopula},
            with a simple API for nested-copula density and gradients,
            plus a per-edge $d_c$-monotonicity diagnostic for the
            nesting-validity condition.
      \item As a byproduct, conditional-distribution sampling~\citep{hofert2013preprint} of these copulas
            becomes practical, previously blocked by the high-order derivatives; due to space, we defer to \Cref{app:rosenblatt}.
\end{enumerate}

\paragraph{Related work.}
\label{sec:related-work}
We position \textsc{acopula} against four neighbouring lines of work,
each summarised below; \Cref{tab:capability-comparison} aggregates the
resulting capability gaps in tabular form.
\begin{table}[htb]
  \centering\scriptsize
  \setlength{\tabcolsep}{3pt}
  \renewcommand{\arraystretch}{0.9}
  \caption{Capability comparison across nested-copula and copula-survival
    tools (R \texttt{copula}~\citep{hofert2011sampling,hofert2012estimators},
    HAC~\citep{okhrin2014hac},
    Copulas.jl~\citep{laverny2024copulas},
    HACSurv$^\dagger$~\citep{liu2024hacsurv},
    DCSurvival~\citep{foomani2023copula},
    CopulaCenR~\citep{sun2020copulacenr}, and the Vine implementation
    \texttt{rvinecopulib}~\citep{nagler2019rvinecopulib}). Tested $d$ reports
    each tool's largest successful attempt at a nested real-data example, including in this work.}
  \label{tab:capability-comparison}
  \definecolor{oursblue}{HTML}{B8D4FB}
  \begin{tabular}{l ccccccc >{\columncolor{oursblue}}c}
    \toprule
                                              & R \texttt{copula}                & HAC                              & Copulas.jl                       & HACSurv$^\dagger$                & DCSurvival                       & CopulaCenR                       & Vine                             & \textbf{Ours} \\
    \midrule
    \textbf{User-defined classical generator} & $\times$                         & $\times$                         & \cellcolor{oursblue}$\checkmark$ & N/A                              & $\times$                         & $\times$                         & N/A                              & $\checkmark$  \\
    Deep Archimedean generators               & $\times$                         & $\times$                         & $\times$                         & \cellcolor{oursblue}$\checkmark$ & \cellcolor{oursblue}$\checkmark$ & $\times$                         & N/A                              & $\checkmark$  \\
    Sampling                                  & \cellcolor{oursblue}$\checkmark$ & \cellcolor{oursblue}$\checkmark$ & \cellcolor{oursblue}$\checkmark$ & N/A                              & \cellcolor{oursblue}$\checkmark$ & \cellcolor{oursblue}$\checkmark$ & \cellcolor{oursblue}$\checkmark$ & $\checkmark$  \\
    Nesting depth (exact density)             & 2                                & \cellcolor{oursblue}any          & N/A                              & 2                                & N/A                              & N/A                              & N/A                              & any           \\
    Automatic density gradients               & $\times$                         & $\times$                         & $\times$                         & \cellcolor{oursblue}$\checkmark$ & \cellcolor{oursblue}$\checkmark$ & $\times$                         & $\times$                         & $\checkmark$  \\
    \textbf{Family-agnostic density}          & $\times$                         & $\times$                         & $\times$                         & $\times$                         & \cellcolor{oursblue}$\checkmark$ & N/A                              & $\times$                         & $\checkmark$  \\
    \textbf{Per-variable censoring}           & none                             & none                             & none                             & competing                        & bivar.                           & bivar.                           & none                             & per-variable  \\
    Tested $d$ (nested real-data)             & $35$                             & ${\sim}10$                       & N/A                              & $6$                              & $2$                              & $2$                              & $98$                             & $98$          \\
    \bottomrule
  \end{tabular}
\end{table}

\begin{itemize}[leftmargin=*]
      \item {\em Classical R-package ecosystem.}
            The R \texttt{copula}~\citep{hofert2011sampling,hofert2012estimators}
            and HAC~\citep{okhrin2014hac} packages compute generator derivatives
            via hand-coded closed forms for five families, restrict every edge
            to same-family pairs, and offer no analytic parameter gradients.
            The simulated maximum likelihood
            estimator~\citep{hofert2012estimators} relies on Laplace-transform
            frailty sampling, with Monte Carlo variance that grows with
            dimension; our exact-likelihood approach attains
            ${\sim}100\times$ speedup at matched accuracy (\Cref{app:smle}),
            and our deterministic-density comparison against R's
            \texttt{nacLL} reaches ${\sim}650\times$ at $d{=}35$ before
            R aborts (\Cref{sec:exp-scaling}).
      \item {\em Censoring-aware copula models.}
            CopulaCenR~\citep{sun2020copulacenr} fits censored bivariate
            and trivariate copulas, including interval-censoring via
            $2^d$-corner inclusion--exclusion of the survival copula;
            \citet{li2020copula,li2021evaluating} handle right-censoring
            using brute-force symbolic differentiation.
            Both restrict to low ambient dimension.
            HACSurv~\citep{liu2024hacsurv} fits hierarchical Archimedean
            copulas for competing risks where each observation reduces to one
            event time and the likelihood is a single first-order partial of
            the nested CDF. We handle the high-order mixed partials that
            arise with per-variable right-censoring in the survival sense,
            and discuss interval-censoring in \Cref{sec:discussion-validity}.
      \item {\em Deep and probabilistic ML copulas.}
            Recent neural-copula
            work~\citep{foomani2023copula,liu2025discrete} learns
            flexible generators for bivariate dependent censoring and
            discrete-diffusion joint sampling; we are complementary, targeting
            exact nested likelihoods at moderate-to-high $d$ with per-variable
            censoring.
            Vine copulas~\citep{aas2009pair,nagler2019rvinecopulib,stoeber2015simplified} and
            hierarchical Kendall copulas~\citep{brechmann2014hierarchical} are
            density-first: the censored likelihood requires a per-observation
            multivariate integral over the censored block, exponential in its size.
            METIC~\citep{chen2025metic} handles this at small $d$ via a
            stochastic-numerical scheme; no vine implementation reaches
            per-variable censoring at $d{=}53$ (\Cref{app:vine,app:hkc}).
      \item {\em Higher-order AD and differentiable probabilistic systems.}
            Our truncated-Taylor-series computation uses JAX's
            Taylor-mode AD primitive~\citep{bettencourt2019taylor}, also formalized in
            the Weil-algebraic framework
            of~\citet{betancourt2018geometric}: order-$d$ derivatives in a
            single forward pass on a truncated polynomial algebra
            (\Cref{sec:method-complexity}).
            \citet{lin2024autodiff} extends the JVP-rule protocol to
            functional derivatives, complementary to the Taylor-mode
            AD we use here.
            NumPyro~\citep{phan2019composable} exposes first-order but not higher-order
            gradients required for nested copulas;
            Gen-AC~\citep{ng2021generativearchimedean} learns Laplace transforms
            with Monte Carlo variance that grows with derivative order.
\end{itemize}

\paragraph{Roadmap.}
\Cref{sec:problem-setup,sec:method,sec:implementation} formulate the
problem, present the key algorithm, and describe the JAX
implementation.  \Cref{sec:experiments} validates correctness on
simulation, characterises wall-clock scaling against R's \texttt{nacLL}
with a ${\sim}650\times$ speedup at $d{=}35$ before R aborts, and runs
four real-data experiments: MIMIC-IV at $d{=}53$ with classical and
neural Archimedean generators under per-variable censoring; S\&P~500
at $d{=}98$---a nested likelihood beyond R's reach; diabetic
retinopathy across ten Archimedean families, five without any prior
censored implementation; and end-to-end reproduction of HACSurv's
Framingham competing-risks pipeline.
\Cref{sec:discussion-validity} addresses validity under mixed families
and amortized JIT compile cost across observations.

\section{Background and Notation}
\label{sec:problem-setup}

\begin{wrapfigure}{r}{0.36\textwidth}
  \vspace{-63pt}
  \centering
  \begin{tikzpicture}[
    >={Stealth[length=2pt]},
    cop/.style={draw, rounded corners=2pt, fill=blue!25,
        minimum width=0.72cm, minimum height=0.72cm, inner sep=0pt},
    leaf/.style={draw, circle, fill=white,
        minimum size=0.72cm, inner sep=0pt},
    leafc/.style={leaf, pattern=north east lines, pattern color=black!50},
    ann/.style={draw=none, font=\tiny, text=black!85, inner sep=0.5pt},
    grpann/.style={draw=none, font=\scriptsize, text=black!70},
    pdfcurve/.pic={
        \fill[blue!30]
        plot[smooth, samples=42, domain=-0.20:0.20]
        ({\x}, {0.22*exp(-90*\x*\x) - 0.10})
        -- (0.20, -0.10) -- (-0.20, -0.10) -- cycle;
        \draw[blue!80, line width=0.5pt, smooth, samples=42, domain=-0.20:0.20]
        plot ({\x}, {0.22*exp(-90*\x*\x) - 0.10});
        \draw[gray!75, line width=0.35pt] (-0.21, -0.10) -- (0.21, -0.10);
      },
    cdfcurve/.pic={
        \draw[gray!70, line width=0.35pt]
        (-0.16, 0.13) -- (0.16, 0.13);
        \draw[red!80, line width=0.7pt, smooth, samples=42, domain=-0.16:0.16]
        plot ({\x}, {0.26/(1+exp(-28*\x)) - 0.13});
        \draw[gray!75, line width=0.4pt] (-0.16, -0.13) -- (0.16, -0.13);
        \node[scale=0.5, font=\tiny, anchor=east, inner sep=0.4pt, text=black!75]
        at (-0.16, 0.13) {1};
        \node[scale=0.5, font=\tiny, anchor=east, inner sep=0.4pt, text=black!75]
        at (-0.16, -0.13) {0};
      },
    ]
    \node[cop] (root) at (0, 0) {};
    \node[cop] (A)    at (-1.95, -1.05) {};
    \node[cop] (B)    at ( 0.20, -1.05) {};

    \node[leaf]  (u1) at (-2.70, -2.10) {};
    \node[leaf]  (u2) at (-1.95, -2.10) {};
    \node[leafc] (u3) at (-1.20, -2.10) {};
    \node[leaf]  (u4) at (-0.175,-2.10) {};
    \node[leaf]  (u5) at ( 0.575,-2.10) {};
    \node[leaf]  (u6) at ( 1.37, -2.10) {};

    \foreach \n/\lab in {root/\psi_0, A/\psi_1, B/\psi_2} {
        \pic at ([yshift=4pt]\n.center) {cdfcurve};
        \node[ann] at ([yshift=-6pt]\n.center) {$\lab$};
      }

    \foreach \n/\lab in {u1/u_1, u2/u_2, u4/u_4, u5/u_5, u6/u_6} {
        \pic at ([yshift=4pt]\n.center) {pdfcurve};
        \node[ann] at ([yshift=-6pt]\n.center) {$\lab$};
      }
    \pic at ([yshift=4pt]u3.center) {pdfcurve};
    \node[ann] at ([yshift=-6pt]u3.center) {$u_3$};

    \draw[->, thin] (root) -- (A);
    \draw[->, thin] (root) -- (B);
    \draw[->, thin, rounded corners=3pt] (root) -- (1.37, -0.74) -- (u6.north);
    \draw[->, thin] (A) -- (u1);
    \draw[->, thin] (A) -- (u2);
    \draw[->, thin] (A) -- (u3);
    \draw[->, thin] (B) -- (u4);
    \draw[->, thin] (B) -- (u5);
    \draw[->, thin] ([yshift=8pt]root.north) -- (root.north);

    \begin{scope}[on background layer]
      \node[draw=orange!60, dashed, rounded corners=3pt,
        fit=(A)(u1)(u2)(u3),
        inner sep=1pt] (grp1) {};
      \node[draw=red!60, dashed, rounded corners=3pt,
        fit=(B)(u4)(u5),
        inner sep=1pt] (grp2) {};
    \end{scope}
    \node[grpann] at ([yshift=-3pt]grp1.south) {strong};
    \node[grpann] at ([yshift=-3pt]grp2.south) {strong};
  \end{tikzpicture}
  \caption{Nested Archimedean copula on six leaves.
    Nodes carry generators forming copula CDFs of their children;
    internal subtrees capture within-group dependence.
    Leaves are variables with marginal densities; the hatched leaf~$u_3$ is censored.}
  \label{fig:nac-tree}
  \vspace{-20pt}
\end{wrapfigure}
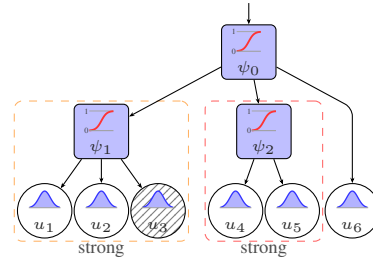

A {\em copula} $C\colon [0,1]^d \to [0,1]$ captures the dependence
among $d$ uniform random variables, independent of their marginal
distributions~\citep{sklar1959fonctions,embrechts2001modelling,nelsen2006introduction}.
An {\em Archimedean copula} encodes that dependence through a single
  {\em generator} $\psi\colon [0,\infty) \to [0,1]$---a continuous,
strictly decreasing function with $\psi(0){=}1$ that is
$d$-monotone, equivalent to convex for $d{=}2$ and ensured by complete
monotonicity for any $d$~\citep{mcneil2009multivariate}, producing the CDF:
\begin{equation}\label{eq:archcop}
  C(\mathbf{u}) = \psi\!\Bigl(\sum_{j=1}^{d} \psi^{-1}(u_j)\Bigr),
  \qquad \mathbf{u} \in [0,1]^d.
\end{equation}
{\em Nested Archimedean copulas}~\citep{mcneil2008sampling} relax the
flat exchangeability of \Cref{eq:archcop} by arranging generators in
a tree (\Cref{fig:nac-tree}).
Each node~$v$ carries its own generator~$\psi_v$ and the CDF
follows a recursive composition over its children~$c$:
\begin{equation}\label{eq:nested-cdf}
  C_v(\mathbf{u}_v)
  = \psi_v\!\Bigl(
  \sum_{c \in \mathrm{children}(v)}
  \psi_v^{-1}\!\bigl(C_c(\mathbf{u}_c)\bigr)
  \Bigr).
\end{equation}
\Cref{eq:nested-cdf} defines a valid copula whenever each composition
$\psi_{\mathrm{outer}}^{-1} \circ \psi_{\mathrm{inner}}$ is a {\em
    Bernstein function}---one with completely monotone derivative---a
sufficient condition due to~\citet{mcneil2008sampling};
\Cref{sec:discussion-validity} covers same-family closed-form ordering
and the mixed-family sign-alternation diagnostic.

\paragraph{Likelihood with per-variable censoring.}
Fitting requires the density, so we move from \Cref{eq:nested-cdf} to
its mixed partials and the censoring patterns they must accommodate.
Each observation~$i$ carries a {\em censoring mask}
$\delta^{(i)} \in \{0,1\}^d$ with $\delta^{(i)}_j = 1$ marking
variable~$j$ uncensored and $|\delta| := \sum_j \delta_j$ counting
uncensored variables.
Without loss of generality, we consider right-censoring in the survival
sense where marginals are $u_j = S_j(t_j) := \Pr(T_j > t_j)$ and the
copula~$C$ encodes $\Pr(T_1 > t_1, \ldots, T_d > t_d)$, with
right-censored observations $T_j > c_j$ entering as $u_j = S_j(c_j)$.
The likelihood is the order-$|\delta|$ mixed partial in the
uncensored variables:
\begin{equation}
  c_{\delta}(\mathbf{u})
  = \frac{\partial^{|\delta|} C}{\prod_{j: \delta_j = 1} \partial u_j}(\mathbf{u}),
  \qquad
  \ell(\boldsymbol{\theta})
  = \sum_{i=1}^{n} \log c_{\delta^{(i)}}(\mathbf{u}^{(i)};\,\boldsymbol{\theta}),
\end{equation}
with sample size $n$, observation $\mathbf{u}^{(i)} \in [0,1]^d$, and
$\boldsymbol{\theta}$ stacking all generator parameters.
Two prior routes evaluate $c_\delta$: hand-deriving a closed form per
$\delta$---infeasible at $2^d$ patterns---or nesting $|\delta|$
first-order AD calls at $O(2^{|\delta|})$ from intermediate trace
blow-up~\citep{betancourt2018geometric}; neither scales past
$d \approx 10$.
R's \texttt{copula} package, the dominant alternative, uses
hand-derived Stirling-number tables for five
families~\citep{hofert2011sampling,hofert2012density}, avoiding the
exponential cost but covering only single-family two-level nesting,
without censored likelihood or analytic gradients.
We assume covariate-conditional independent censoring throughout;
\Cref{sec:discussion-validity} returns to identifiability.

\paragraph{The \citet{hofert2012density} density.}
\citet{hofert2012density} expressed the nested density~$c_\delta$
as a tree-structured sum involving {\em partial Bell
    polynomials}~$B_{n,k}$ of higher-order derivatives of each
parent--child edge composition $h_{vc} = \psi_v^{-1} \circ \psi_c$
(writing $\psi^{(k)}$ for the $k$-th derivative throughout):
\begin{equation}\label{eq:hofert-density}
  c_\delta(\mathbf{u})
  = \biggl(\sum_{k=1}^{|\delta|} \beta_k^{(r)}
  \cdot \psi_r^{(k)}(t_r)\biggr)
  \cdot
  \!\!\prod_{\substack{\ell \in \mathrm{leaves} \\ \delta_\ell = 1}}\!\!
  \bigl(\psi_{\mathrm{par}(\ell)}^{-1}\bigr)'(u_\ell),
\end{equation}
with $r$ the root, $\mathrm{par}(\ell)$ the parent of leaf $\ell$,
$t_v := \sum_{c \in \mathrm{children}(v)}
  \psi_v^{-1}(C_c(\mathbf{u}_c))$ the inverse-generator argument at
node $v$, and weights~$\beta_k^{(r)}$ defined recursively in
\Cref{sec:method-algorithm}---\Cref{app:bell-polys} gives the $B_{n,k}$ formula.
Evaluating these Bell polynomials for user-defined generators
without hand-derived tables is the obstacle \Cref{sec:method}
resolves by powering Taylor coefficients.

\paragraph{Taylor-mode AD.}
Our algorithm needs univariate higher-order derivatives of each edge composition
$h = \psi_v^{-1} \circ \psi_c$.
  {\em Taylor-mode} AD~\citep{griewank2008evaluating}, or {\em jet
    propagation}, returns the factorial-scaled derivative vector
\(\textsc{Jet}(f, x, n) = \bigl(f(x), f'(x)/1!, \ldots, f^{(n)}(x)/n!\bigr)\)
in one forward pass at $O(n^2)$, versus $O(2^n)$ for $n$ nested
first-order AD calls.
Our implementation builds on a JIT-friendly variant of JAX's \texttt{jet}~\citep{radul2020you};
\Cref{sec:method} treats it as a black box.

\section{Method}
\label{sec:method}

We walk through a concrete example exhibiting every algorithmic
ingredient (\Cref{sec:method-worked}), state the bottom-up recursion
(\Cref{sec:method-algorithm}; full Bell-polynomial derivation in
\Cref{app:bell-polys,app:poly-power}), give the $O(d^3)$ complexity,
and treat the per-variable censoring mechanism.

\begin{wrapfigure}{r}{0.2\textwidth}
  \centering
  \vspace{-50pt}
  \begin{tikzpicture}[scale=0.55,
    every node/.style={draw, rounded corners, minimum width=0.4cm,
        minimum height=0.25cm, font=\scriptsize},
    leaf/.style={draw, circle, minimum size=0.28cm, inner sep=0pt,
        font=\scriptsize, fill=white},
    censored/.style={draw, circle, minimum size=0.28cm, inner sep=0pt,
        font=\scriptsize, pattern=north east lines, pattern color=black!50},
    cop/.style={fill=blue!25},
    level 1/.style={sibling distance=3cm, level distance=.5cm},
    level 2/.style={sibling distance=2cm, level distance=.6cm},
    edge from parent/.style={draw, -{Stealth[length=2.5pt]}},
    ]
    \node[cop] (root) {$\psi_0$}
    child { node[cop] (A) {$\psi_1$}
        child { node[leaf] (u1) {$u_1$} }
        child { node[leaf] (u2) {$u_2$} } }
    child { node[censored] (u3) {$u_3$} };
    \draw[-{Stealth[length=2.5pt]}, thin]
    ([yshift=6pt]root.north) -- (root.north);
  \end{tikzpicture}
  \label{fig:nac-tree-mini}
  \vspace{-10pt}
\end{wrapfigure}

\subsection{Worked example: $d{=}3$ nested copula with one censored leaf}
\label{sec:method-worked}
Consider the tree on the right: a root~$\psi_0$ with
sector $\psi_1$ over uncensored leaves~$u_1, u_2$ and a right-censored
sibling~$u_3$.  The censoring mask
$\delta = (1, 1, 0)$ gives an order-2 mixed partial
$\partial^2 C / \partial u_1 \partial u_2$ at $\mathbf{u}$.  Each leaf
carries an $\alpha$-vector: $\alpha^{(u_1)} = \alpha^{(u_2)} = [0,1]$
and $\alpha^{(u_3)} = [1]$, the Cauchy-product identity.  Four
operations propagate these vectors upward and yield the density.
\begin{enumerate}[leftmargin=*]
  \item {\em Cauchy product at the sector.}
        Combine the uncensored leaves:
        $\beta^{(\psi_1)} = \alpha^{(u_1)} \ast \alpha^{(u_2)} = [0,0,1]$,
        where $\beta_2^{(\psi_1)} = 1$ encodes the two differentiations
        $\psi_1$ contributes upstream.
  \item {\em Bell transform across the edge $\psi_0 \leftarrow \psi_1$.}
        Convert $\beta^{(\psi_1)}$ into a child-wise $\alpha$-vector at
        the root using higher-order derivatives of the edge composition
        $h = \psi_0^{-1} \circ \psi_1$ at
        $t_1 = \psi_1^{-1}(u_1) + \psi_1^{-1}(u_2)$:
        $\alpha_k^{(\psi_0, \psi_1)} = \sum_{j=k}^{2}
          \beta_j^{(\psi_1)} B_{j,k}(h'(t_1), h''(t_1), \ldots)$.
        Only $j = 2$ contributes, giving
        $\alpha_1^{(\psi_0, \psi_1)} = h''(t_1)$ and
        $\alpha_2^{(\psi_0, \psi_1)} = (h'(t_1))^2$.
  \item {\em Censored sibling.}
        The leaf $u_3$ contributes $\alpha^{(u_3)} = [1]$, which leaves
        $\alpha^{(\psi_0, \psi_1)}$ unshifted under the Cauchy product at
        the root:
        $\beta^{(\psi_0)} = \alpha^{(\psi_0, \psi_1)} \ast [1]
          = [0, h''(t_1), (h'(t_1))^2]$.
  \item {\em Density assembly.}
        The Hofert formula~\Cref{eq:hofert-density} gives
        \begin{equation}\label{eq:worked-density}
          c_\delta(\mathbf{u})
          = \bigl[\psi_0'(t_r)\,h''(t_1) + \psi_0''(t_r)\,(h'(t_1))^2\bigr]
          \cdot
          (\psi_1^{-1})'(u_1) \cdot (\psi_1^{-1})'(u_2),
        \end{equation}
        with $t_r = \psi_0^{-1}\!\bigl(C_1(u_1, u_2)\bigr) + \psi_0^{-1}(u_3)$
        and $C_1$ the inner copula generated by $\psi_1$.
        The censored leaf~$u_3$ enters $t_r$ via its inverse-generator
        term but drops out of the leaf product since $\delta_3 = 0$.
\end{enumerate}

\ifcompact
  \begin{wrapfigure}{R}{0.5\textwidth}
    \begin{minipage}{0.5\textwidth}\vspace{-30pt}
      \begin{algorithm}[H]
  \footnotesize
  \caption{Main algorithm.}
  \label{alg:bell-density}
  \algrenewcommand{\algorithmicindent}{0.8em}%
  \begin{algorithmic}[1]
    \Require ~Copula tree of valid nesting at root~$r$, observations~$\mathbf{u}$
    \Ensure ~Log-density $\log c(\mathbf{u})$
    \State Compute $t_v = \sum_c \psi_v^{-1}(C_c)$ at every node~$v$ (bottom-up)
    \For{each node~$v$, from leaves to root}
    \For{each child~$c$ of~$v$}
    \If{$c$ is an uncensored leaf}
    \State $\alpha^{(c)} \gets [0,\; 1]$
    \Comment{one derivative, coefficient~1}
    \ElsIf{$c$ is a censored leaf}
    \State $\alpha^{(c)} \gets [1]$
    \Comment{identity under convolution}
    \Else
    \State $p \gets \textsc{Jet}\bigl(\psi_v^{-1} \circ \psi_c,
      \; t_c,\; d_c\bigr)$\label{ln:edge-jet}
    \Comment{Taylor coeffs of $h_{vc}$, computed via implicit equation solve in \Cref{app:compose-taylor}}
    \State $q \gets p$
    \Comment{$q$ accumulates $p^k$; starts at $p^1$}
    \For{$k = 1,\ldots,d_c$}
    \State $\alpha^{(c)}_k \gets
      (1/k!)\,\langle \tilde\beta^{(c)},\; q \rangle$
    \Comment{reads off $B_{j,k}$}
    \State $q \gets q * p$
    \Comment{$q = p^{k+1}$}
    \EndFor
    \EndIf
    \EndFor
    \State $\beta^{(v)} \gets \alpha^{(c_1)} * \alpha^{(c_2)} * \cdots$
    \Comment{convolve children}
    \EndFor
    \State $q_1,\ldots,q_d \gets
      \textsc{Jet}(\psi_r, t_r, d)$
    \Comment{root Taylor coeffs}
    \State \Return $\log\Bigl(\bigl(\sum_k \beta_k^{(r)} \cdot k!\, q_k\bigr)
      \cdot \prod_{\ell:\,\delta_\ell = 1} (\psi_{\mathrm{par}(\ell)}^{-1})'(u_\ell)\Bigr)$
  \end{algorithmic}
\end{algorithm}

    \end{minipage}\vspace{-30pt}
  \end{wrapfigure}
\else
  \begin{figure}[t]\centering
    \begin{minipage}{0.92\textwidth}
      \begin{algorithm}[H]
  \footnotesize
  \caption{Main algorithm.}
  \label{alg:bell-density}
  \algrenewcommand{\algorithmicindent}{0.8em}%
  \begin{algorithmic}[1]
    \Require ~Copula tree of valid nesting at root~$r$, observations~$\mathbf{u}$
    \Ensure ~Log-density $\log c(\mathbf{u})$
    \State Compute $t_v = \sum_c \psi_v^{-1}(C_c)$ at every node~$v$ (bottom-up)
    \For{each node~$v$, from leaves to root}
    \For{each child~$c$ of~$v$}
    \If{$c$ is an uncensored leaf}
    \State $\alpha^{(c)} \gets [0,\; 1]$
    \Comment{one derivative, coefficient~1}
    \ElsIf{$c$ is a censored leaf}
    \State $\alpha^{(c)} \gets [1]$
    \Comment{identity under convolution}
    \Else
    \State $p \gets \textsc{Jet}\bigl(\psi_v^{-1} \circ \psi_c,
      \; t_c,\; d_c\bigr)$\label{ln:edge-jet}
    \Comment{Taylor coeffs of $h_{vc}$, computed via implicit equation solve in \Cref{app:compose-taylor}}
    \State $q \gets p$
    \Comment{$q$ accumulates $p^k$; starts at $p^1$}
    \For{$k = 1,\ldots,d_c$}
    \State $\alpha^{(c)}_k \gets
      (1/k!)\,\langle \tilde\beta^{(c)},\; q \rangle$
    \Comment{reads off $B_{j,k}$}
    \State $q \gets q * p$
    \Comment{$q = p^{k+1}$}
    \EndFor
    \EndIf
    \EndFor
    \State $\beta^{(v)} \gets \alpha^{(c_1)} * \alpha^{(c_2)} * \cdots$
    \Comment{convolve children}
    \EndFor
    \State $q_1,\ldots,q_d \gets
      \textsc{Jet}(\psi_r, t_r, d)$
    \Comment{root Taylor coeffs}
    \State \Return $\log\Bigl(\bigl(\sum_k \beta_k^{(r)} \cdot k!\, q_k\bigr)
      \cdot \prod_{\ell:\,\delta_\ell = 1} (\psi_{\mathrm{par}(\ell)}^{-1})'(u_\ell)\Bigr)$
  \end{algorithmic}
\end{algorithm}

    \end{minipage}
  \end{figure}
\fi
The {\em Bell polynomial values} in \Cref{eq:worked-density} ---
$h''(t_1)$ at $k{=}1$ and $(h'(t_1))^2$ at $k{=}2$---equal $n!/k!$
times the coefficients of $\varepsilon^n$ in $p(\varepsilon)^k$
(here $n{=}2$, with $2!/1!{=}2$ for $k{=}1$ and $2!/2!{=}1$ for
$k{=}2$), where $p(\varepsilon) = h'(t_1)\,\varepsilon + h''(t_1)\,\varepsilon^2/2 + \cdots$
(\Cref{app:poly-power}).
Taylor-mode AD applied to~$h$ at~$t_1$ returns these coefficients of~$p$
in one forward pass; raising $p$ to the $k$-th power by truncated
convolution yields all required Bell polynomial values without
combinatorial enumeration or per-family tables.
The Cauchy product, the Bell transform, and the polynomial powering of
Taylor coefficients are the entire algorithm.

\subsection{Algorithm}
\label{sec:method-algorithm}

The general procedure repeats the worked example bottom-up at every
internal node.
For each child~$c$ of node~$v$, where $m_v = |\mathrm{children}(v)|$,
the {\em $\alpha$-vector}
$\alpha^{(v,c)}$ encodes $c$'s contribution to the parent density,
and the {\em $\beta$-vector} $\beta^{(v)}$ collects the combined
contribution of children of~$v$; uncensored leaves initialise
$\alpha = [0,1]$ and censored leaves use $\alpha = [1]$, the Cauchy-product
identity.
Let $\delta_v$ be the count of uncensored leaves in the subtree at~$v$.
Two operations propagate these vectors upward, the {\em Bell transform}
(\Cref{eq:alpha-from-beta}) across each edge $v \leftarrow c$ and the
  {\em Cauchy product} (\Cref{eq:beta-from-alpha}) combining siblings:
\par\noindent
\begin{minipage}{0.53\textwidth}
  \vspace*{-\abovedisplayskip}
  \begin{multline}\label{eq:alpha-from-beta}
    \forall k                = 1,\ldots,|\delta_c| \colon \\
    \alpha_k^{(v,c)} = \sum_{j=k}^{|\delta_c|} \beta_j^{(c)} \;
    B_{j,k}\!\bigl(h'_{vc}(t_c), h''_{vc}(t_c), \ldots\bigr),
  \end{multline}
\end{minipage}\hfill
\begin{minipage}{0.46\textwidth}
  \vspace*{-\abovedisplayskip}
  \begin{multline}\label{eq:beta-from-alpha}
    \forall k = 0,\ldots,|\delta_v| \colon \\
    \beta_k^{(v)} = \!\!\sum^{\forall k_i \ge 0}_{k_1 + \cdots + k_{m_v} = k}\!\!
    \alpha_{k_1}^{(v,c_1)} \cdots \alpha_{k_{m_v}}^{(v,c_{m_v})} .
  \end{multline}
\end{minipage}
The recursion follows~\citet{hofert2012density};
our contribution is evaluating the Bell polynomials~$B_{j,k}$ in
\Cref{eq:alpha-from-beta} via polynomial powering of Taylor
coefficients.
With $\varepsilon$ a formal indeterminate and $p_{vc}(\varepsilon)
  = \sum_{m \ge 1} h_{vc}^{(m)}(t_c)\,\varepsilon^m / m!$
the exponential generating function of the edge derivatives, a
generating-function identity gives
$B_{n,k} = (n!/k!) \cdot \operatorname{Coeff}\!\bigl(p_{vc}(\varepsilon)^k;\; \varepsilon^n\bigr)$,
where $\operatorname{Coeff}(q;\,\varepsilon^n)$ extracts the
coefficient of $\varepsilon^n$ in the polynomial $q$.
Applying Taylor-mode AD to~$h_{vc}$ at~$t_c$ returns the coefficients
of~$p_{vc}$ directly; iterated truncated convolution $q \gets q \cdot
  p_{vc}$ produces $p_{vc}^k$.
Substituting back, \Cref{eq:alpha-from-beta} reduces to a single
inner product
$\alpha_k^{(v,c)} = (1/k!)\,\sum_{j=k}^{|\delta_c|} \tilde\beta^{(c)}_j\, (q_k)_j$
with $\tilde\beta_j := j!\,\beta_j$ the rescaled $\beta$-vector and
$(q_k)_j$ the $\varepsilon^j$ coefficient of~$p_{vc}^k$.

\Cref{alg:bell-density} summarises the resulting bottom-up traversal.
Every step is differentiable in the parameters, so first-order
AD of the log-likelihood enables gradient-based optimisers.
For numerical stability, the algorithm performs operations in the log-domain (\Cref{app:numerical-stability}).
Moreover, at \Cref{ln:edge-jet}, rather than feeding the explicit composition
$\psi_v^{-1} \circ \psi_c$ to \textsc{Jet}, the algorithm computes
$p_{vc}$ by solving the implicit equation
$\psi_v(h_{vc}(t)) = \psi_c(t)$ for its Taylor coefficients to
avoid the catastrophic cancellation that arises when $\psi_c(t)$
underflows before $\psi_v^{-1}$ can recover it
(\Cref{app:compose-taylor}).
In principle, the algorithm can also use arbitrary-precision arithmetic
to avoid these tricks at the cost of efficiency.

\paragraph{Complexity}
\label{sec:method-complexity}

\begin{wraptable}{r}{0.56\textwidth}\centering\scriptsize
  \caption{Asymptotic cost by tree shape, governed by the largest
    internal-child subtree leaf count $|\delta_v|$.  Any shape with
    $\max |\delta_v| \le O(\sqrt{d})$ at every level reaches the
    $O(d^2)$ root-jet cost.  $^{\dagger}$Chains do not appear in
    real copula data to our knowledge.  Derivation in
    \Cref{app:complexity}.}
  \label{tab:complexity-shapes}
  \tikzset{
  cop/.style={draw, rounded corners=0.5pt, fill=blue!25,
      minimum width=0.16cm, minimum height=0.10cm,
      inner sep=0.1pt, font=\tiny, scale=0.6},
  leaf/.style={draw, circle, fill=white, minimum size=0.07cm,
      inner sep=0pt, font=\tiny, scale=0.5},
  edge from parent/.style={draw, thin, black!85},
  arr/.style={draw, -{Stealth[length=1pt]}, thin, black!85},
  }
  \setlength{\tabcolsep}{2.5pt}
  \renewcommand{\arraystretch}{1.0}
  \begin{tabular}{@{}cccc@{}}
    \toprule
    Constant sector size
             & Square-root scaling
             & Two large sectors
             & Chain (depth $d$)$^{\dagger}$                         \\
    \midrule
    \begin{tikzpicture}[baseline=(r.base),
        level 1/.style={sibling distance=0.25cm, level distance=0.32cm},
        level 2/.style={sibling distance=0.10cm, level distance=0.28cm},
      ]
      \node[cop] (r) {$\psi$}
      child {node[cop] {$\psi$} child {node[leaf] {}} child {node[leaf] {}}}
      child {node[cop] {$\psi$} child {node[leaf] {}} child {node[leaf] {}}}
      child {node[cop] {$\psi$} child {node[leaf] {}} child {node[leaf] {}}}
      child {node[cop] {$\psi$} child {node[leaf] {}} child {node[leaf] {}}}
      child {node[cop] {$\psi$} child {node[leaf] {}} child {node[leaf] {}}}
      child {node[cop] {$\psi$} child {node[leaf] {}} child {node[leaf] {}}};
      \draw[arr] ([yshift=3pt]r.north) -- (r.north);
    \end{tikzpicture}
             &
    \begin{tikzpicture}[baseline=(r.base),
        level 1/.style={sibling distance=0.36cm, level distance=0.32cm},
        level 2/.style={sibling distance=0.12cm, level distance=0.28cm},
      ]
      \node[cop] (r) {$\psi$}
      child {node[cop] {$\psi$}
          child {node[leaf] {}} child {node[leaf] {}} child {node[leaf] {}}
        }
      child {node[cop] {$\psi$}
          child {node[leaf] {}} child {node[leaf] {}} child {node[leaf] {}}
        }
      child {node[cop] {$\psi$}
          child {node[leaf] {}} child {node[leaf] {}} child {node[leaf] {}}
        }
      child {node[cop] {$\psi$}
          child {node[leaf] {}} child {node[leaf] {}} child {node[leaf] {}}
        };
      \draw[arr] ([yshift=3pt]r.north) -- (r.north);
    \end{tikzpicture}
             &
    \begin{tikzpicture}[baseline=(r.base),
        level 1/.style={sibling distance=0.85cm, level distance=0.32cm},
        level 2/.style={sibling distance=0.12cm, level distance=0.28cm},
      ]
      \node[cop] (r) {$\psi$}
      child {node[cop] {$\psi$}
          child {node[leaf] {}} child {node[leaf] {}} child {node[leaf] {}}
          child {node[leaf] {}} child {node[leaf] {}} child {node[leaf] {}}
        }
      child {node[cop] {$\psi$}
          child {node[leaf] {}} child {node[leaf] {}} child {node[leaf] {}}
          child {node[leaf] {}} child {node[leaf] {}} child {node[leaf] {}}
        };
      \draw[arr] ([yshift=3pt]r.north) -- (r.north);
    \end{tikzpicture}
             &
    \begin{tikzpicture}[baseline={(current bounding box.center)},
        scale=1.35, transform shape]
      \node[cop] (a) {$\psi$};
      \draw[arr] ([xshift=-3pt]a.west) -- (a.west);
      \node[cop] (b) [right=1.5pt of a] {$\psi$};
      \node[font=\tiny] (dots) [right=2pt of b] {$\cdots$};
      \node[cop] (c) [right=2pt of dots] {$\psi$};
      \draw[arr] (a) -- (b);
      \draw[arr] (b.east) -- (dots.west);
      \draw[arr] (dots.east) -- (c.west);
      \node[leaf] (la) [below=3pt of a] {};
      \node[leaf] (lb) [below=3pt of b] {};
      \node[leaf] (lc) [below=3pt of c] {};
      \draw[arr] (a) -- (la);
      \draw[arr] (b) -- (lb);
      \draw[arr] (c) -- (lc);
    \end{tikzpicture}
    \\[6pt]
    $|\delta_v|{=}O(1)$
             & $|\delta_v|{\leq O(\sqrt{d})}$
             & $|\delta_v|{=}d/2$
             & $|\delta_v|{=}\Theta(d{-}\ell)$                       \\
    $O(d^2)$ & $O(d^2)$                        & $O(d^3)$ & $O(d^4)$ \\
    \bottomrule
  \end{tabular}
  \vspace*{-10pt}
\end{wraptable}

Let $T(\mathcal{T})$ denote the total runtime of
\Cref{alg:bell-density} on nesting tree~$\mathcal{T}$.  Each non-root
node~$v$ pays $O(|\delta_v|^3)$ for the polynomial powering its parent
applies to it, which dominates the $O(|\delta_v|^2)$ Cauchy product at
any internal~$v$, giving
\begin{equation}
  T(\mathcal{T}) \;=\; O\!\Bigl(d^2 + \textstyle\sum_{v \neq \mathrm{root}} |\delta_v|^3\Bigr),
\end{equation}
where $d^2$ comes from the order-$d$ jet at the root.
The largest subtree therefore dominates: total cost reaches the
$O(d^2)$ root-jet cost whenever $\sum_{v \neq r} |\delta_v|^3 = O(d^2)$, a
condition met for $\sqrt{d}${\em -decomposed} trees---those where every
non-root subtree satisfies $|\delta_v| = O(\sqrt{d})$ at constant tree
depth---since per-level disjointness gives
$\sum_{v \text{ at level}} |\delta_v|^3 = O(d^2)$.
Any subtree exceeding $\sqrt{d}$ raises the total by its $|\delta_v|^3$ contribution.
\Cref{tab:complexity-shapes} lists the common cases, and
\Cref{app:complexity} derives the bound rigorously.
Each edge's implicit-equation solve costs $O(|\delta_c|^3)$, matching the Bell powering it feeds, so it is absorbable into the total bound (\Cref{app:compose-taylor}).

In contrast, prior tools make weaker trade-offs: R's \texttt{copula} relies on hard-coded Stirling
tables for five families at two nesting levels, and R's symbolic differentiation
\texttt{D()}~\citep{li2020copula,li2021evaluating} incurs a combinatorial
complexity tractable only for $d \lesssim 6$.
\Cref{app:fft-discussion} discusses FFT and scan-based accelerations
for the polynomial-powering step.

\paragraph{Handling per-variable censoring}
\label{sec:method-censoring}

Censoring a leaf~$\ell$ sets $\alpha^{(\ell)} = [1]$ instead of $[0,1]$,
which leaves the Cauchy product unchanged on its level and reduces the
effective dimension~$|\delta_v|$ at every ancestor.
The reduction propagates automatically: shorter Cauchy product, lower
Bell-polynomial degree, and reduced \textsc{Jet} order; per-node cost
scales as $O(|\delta_v|^3)$, so heavy censoring runs faster
(\Cref{tab:censoring-runtime}).

\section{Implementation}
\label{sec:implementation}

\begin{figure}[hbt]
  \begin{minipage}[t]{0.37\textwidth}
    \begin{minted}{python}
      @copula
      class Clayton:
        theta: float
        def generator(self, t):
          return ((1 + t) **
                  (-1.0 / self.theta))

      @copula
      class AMH:
        theta: float
        def generator(self, t):
          e = jnp.exp(-t)
          return ((1 - self.theta) * e /
                  (1 - self.theta * e))
    \end{minted}
  \end{minipage}\hfill\begin{minipage}[t]{0.62\textwidth}
    \begin{minted}[firstnumber=13]{python}
      def nested_amh_clayton(p, u):
        return  AMH(p.theta_outer)(
                  Clayton(p.theta_inner)(
                    marginal(Weibull(1.5, 1.0),
                    obs=u[i, j],
                    censored=((i, j) == (1, 3)))
                  for j in range(5)) 
                for i in range(4))

      p0 = Params(theta_outer=0.5, theta_inner=2.0)
      obs = load_data()
      m = compile_model(nested_amh_clayton, template=p0)
      m.ll_fn(obs, m.flatten(p0))  # log density
      jax.grad(m.ll_fn, argnums=1)(obs, m.flatten(p0))  # exact grad
    \end{minted}
  \end{minipage}
  \vspace{-2pt}
  \captionof{listing}{A nested AMH/Clayton copula with 4 sectors of 5
    Weibull-marginal leaves, per-variable censoring, and end-to-end
    differentiable inference.}
  \label{lst:api}
  \vspace{-6pt}
\end{figure}

\textsc{acopula} is an open-source JAX~\citep{bradbury2018jax}
package.\footnote{\ifarxiv
    Available at \url{https://github.com/thisiscam/acopula}.
  \fi}
The user supplies a generator class; the framework derives the
inverse, all higher-order derivatives, the Bell-polynomial
coefficients, the density values, and the parameter gradients
automatically.
\Cref{lst:api} shows two custom families and a 4-sector nested model
with per-variable censoring fitted in fewer than twenty Python lines.

The \texttt{@copula} decorator registers parameters and derives the
generator inverse via symbolic
inversion~\citep{radul2020you}; when symbolic inversion fails,
\textsc{acopula} falls back to bisection with
implicit-function-theorem (IFT) gradients (\Cref{app:oryx-fallback}).
\textsc{acopula} compiles a single likelihood program that handles
every per-observation censoring mask~$\delta^{(i)}$ without
recompilation (\Cref{app:mask-aware-jit}).
As is standard in AD-based frameworks, users may override any of the
framework's automatic derivations with a hand-derived alternative ---
a custom generator inverse, a closed-form edge composition, or a
per-family Taylor-coefficient routine in place of \textsc{Jet} ---
for performance or numerical stability.

\section{Experiments}
\label{sec:experiments}

\begin{wraptable}{r}{0.36\textwidth}\centering\scriptsize
  \vspace{-30pt}
  \caption{Replicated parameter recovery on nested Clayton copulas
      ($500$ replications, $500$ samples per replication;
      $L$ denotes nesting depth, EmpSE the across-replication SE,
      HessSE the observed-Fisher SE).
      Bias stays under $2.7\%$ of the true value; the empirical SE matches the observed-Fisher
      SE. Results suggest standard Hessian-based confidence intervals on our
      exact-likelihood fits are reliable.}
\label{tab:recovery-highdim}
\setlength{\tabcolsep}{2pt}
\begin{tabular}{@{}c l r r r r@{}}
      \toprule
      $d$/L & Param                      & True & Mean  & EmpSE & HessSE \\
      \midrule
      10/2  & $\theta_{\mathrm{outer}}$  & 1.50 & 1.521 & 0.069 & 0.070  \\
            & $\theta_{\mathrm{inner}}$  & 3.00 & 3.023 & 0.054 & 0.053  \\
      \midrule
      20/2  & $\theta_{\mathrm{outer}}$  & 2.00 & 2.019 & 0.048 & 0.047  \\
            & $\theta_{\mathrm{inner}}$  & 4.00 & 4.023 & 0.048 & 0.047  \\
      \midrule
      50/3  & $\theta_{\mathrm{outer}}$  & 1.00 & 1.016 & 0.052 & 0.050  \\
            & $\theta_{\mathrm{middle}}$ & 2.00 & 2.016 & 0.028 & 0.028  \\
            & $\theta_{\mathrm{inner}}$  & 4.00 & 4.023 & 0.032 & 0.030  \\
      \bottomrule
\end{tabular}
\vspace{-30pt}
\end{wraptable}

We verify correctness on simulated data, characterise wall-clock
scaling against R's \texttt{nacLL}, then evaluate three real-data
benchmarks that each exercise a capability prior tools cannot deliver:
per-variable censoring at $d{=}53$ on MIMIC-IV, nested likelihood at
$d{=}98$ on S\&P~500, and family-agnostic censored MLE on retinopathy.
All runs share a single JAX code base in float64.
The scaling sweep, the recovery study, R comparison, and  S\&P~500 and retinopathy fit run on a single AMD Threadripper~3960X CPU
core in single-process mode for the apples-to-apples R comparison; the
MIMIC-IV experiments run on an NVIDIA RTX~3090~Ti GPU
(\Cref{tab:mimic-compile-cost}).

\subsection{Correctness validation}
\label{sec:exp-correctness}
We verify the implementation in three ways before turning to real
data.  The {\em parameter-recovery} experiment samples from a
nested Clayton model with known true parameters, refits by maximum
likelihood, and compares the estimate to the truth.
\Cref{tab:recovery-highdim} reports recovery with $R{=}500$ Monte
Carlo replications, $d \in \{10, 20, 50\}$, and tree depth up to
three: every parameter recovers to within $2.7\%$ bias, and the
empirical SE matches the observed-Fisher SE within $2.9\%$,
confirming that Hessian-based inference on the Bell-polynomial
likelihood is well-calibrated.
We verify density agreement with R's \texttt{nacLL} to ${<}10^{-14}$ and with
Mathematica's symbolic derivatives up to
$d{=}60$ (\Cref{app:r-comparison}).
AD gradients match central differences to
${<}10^{-8}$ in \Cref{app:gradient-check}; \Cref{app:recovery-extended}
extends parameter recovery to Frank and Gumbel.

\subsection{Scaling: head-to-head against R \texttt{nacLL}}
\label{sec:exp-scaling}

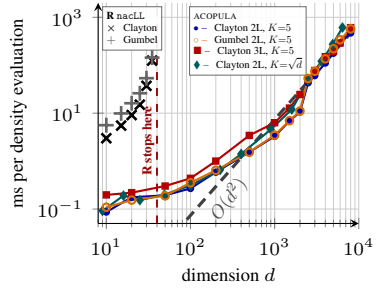
\begin{wrapfigure}{r}{0.378\textwidth}\vspace{-35pt}\centering
  \begin{tikzpicture}
  \begin{axis}[
      width=0.36\textwidth,
      height=4.5cm,
      xlabel={dimension $d$},
      ylabel={ms per density evaluation},
      xmode=log, ymode=log,
      xmin=8, xmax=10000,
      ymin=5e-2, ymax=2e3,
      log basis x={10}, log basis y={10},
      title style={yshift=-4pt, font=\scriptsize},
      xlabel style={font=\scriptsize, yshift=2pt},
      ylabel style={font=\scriptsize, yshift=-3pt},
      xticklabel style={font=\scriptsize},
      yticklabel style={font=\scriptsize},
      legend style={font=\fontsize{5}{6}\selectfont},
      legend cell align={left},
      grid=both,
      grid style={very thin, color=black!20},
      axis lines=left,
      clip=true,
    ]
    \addplot[black!75, dash pattern=on 4pt off 2.5pt, line width=1.2pt,
      domain=4:9000, samples=2, forget plot]
    {471.3920 / 8000^2 * x^2};
    \addplot[red!60!black, dashed, line width=0.7pt, forget plot]
    coordinates {(40,5e-2) (40,2e3)};
    \addplot[blue!70!black, mark=*, mark size=1.0pt, thick,
      forget plot]
    coordinates {
        (10,0.0867) (20,0.1761) (50,0.1933) (100,0.2691) (200,0.5875)
        (500,1.5389) (1000,3.2912) (1500,6.8475) (2000,10.5962)
        (2500,43.1103) (3000,60.7525) (4000,111.2474) (5000,178.1104)
        (6000,286.5924) (8000,471.3920)
      };
    \addplot[red!70!black, mark=square*, mark size=1.0pt, thick,
      forget plot]
    coordinates {
        (10,0.1974) (20,0.2195) (50,0.3020) (100,0.4373) (200,1.0009)
        (500,3.4329) (1000,6.3205) (1500,12.7621) (2000,24.5408)
        (2500,49.4341) (3000,77.5042) (4000,150.3705) (5000,194.9972)
        (6000,291.2742) (8000,607.8511)
      };
    \addplot[teal!70!black, mark=diamond*, mark size=1.4pt, thick,
      forget plot]
    coordinates {
        (9,0.0917) (16,0.1905) (25,0.1531) (49,0.2000)
        (100,0.3097) (225,0.6545) (400,1.4077) (900,4.5681)
        (1600,11.9699) (2500,48.1744) (3600,121.1474) (4900,274.7193)
        (6400,611.3822)
      };
    \addplot[orange!85!black, mark=o, mark size=1.4pt, thick,
      forget plot]
    coordinates {
        (10,0.1092) (20,0.1536) (50,0.1886) (100,0.3538) (200,0.6437)
        (500,1.5315) (1000,3.4816) (1500,6.9696) (2000,11.1022)
        (2500,52.9557) (3000,74.8007) (4000,136.2303) (5000,221.0412)
        (6000,340.1462) (8000,571.4411)
      };
    \addplot[only marks, black, mark=x, mark size=2.6pt,
      line width=1pt, forget plot]
    coordinates {
        (10,2.9994) (15,5.5079) (20,9.1342)
        (25,15.0436) (30,37.1981) (35,124.9910)
      };
    \addplot[only marks, black!60, mark=+, mark size=2.8pt,
      line width=1pt, forget plot]
    coordinates {
        (10,5.6151) (15,9.8816) (20,16.1118)
        (25,26.0307) (30,53.6612) (35,144.0508)
      };
    \node[scale=0.5, rotate=90, color=red!60!black, anchor=south west,
      font=\bfseries]
    at (axis cs:42, .4) {R stops here};
    \node[scale=0.7, color=black!70, rotate=53, anchor=north,
      font=\bfseries]
    at (axis cs:200, 0.258) {$O(d^2)$};

    \node[draw=black!55, fill=white, fill opacity=0.92, text opacity=1,
      inner sep=1.5pt, anchor=north west, align=left]
    at (rel axis cs:0.015, 0.985)
    {\scalebox{0.4}{\begin{tabular}{@{}l@{}}
          \textbf{R \texttt{nacLL}}                            \\[0.5pt]
          \textcolor{black}{$\boldsymbol{\times}$}\;\, Clayton \\
          \textcolor{black!60}{$\boldsymbol{+}$}\;\, Gumbel
        \end{tabular}}};

    \node[draw=black!55, fill=white, fill opacity=0.92, text opacity=1,
      inner sep=1.5pt, anchor=north east, align=left]
    at (rel axis cs:0.8, 0.985)
    {\scalebox{0.4}{\begin{tabular}{@{}l@{}}
          \textsc{acopula}                                                       \\[0.5pt]
          \textcolor{blue!70!black}{$\bullet$\;{--}}\;\, Clayton 2L, $K{=}5$     \\
          \textcolor{orange!85!black}{$\circ$\;{--}}\;\, Gumbel 2L, $K{=}5$      \\
          \textcolor{red!70!black}{$\blacksquare$\;{--}}\;\, Clayton 3L, $K{=}5$ \\
          \textcolor{teal!70!black}{$\blacklozenge$\;{--}}\;\, Clayton 2L, $K{=}\sqrt{d}$
        \end{tabular}}};
  \end{axis}
\end{tikzpicture}

  \caption{Wall-clock scaling.
    R only runs at 2-level $K{=}5$; aborts past $d{=}40$
    on Clayton/Gumbel.  By $d{=}35$ R is
    ${\sim}650\times$ slower.  \textsc{acopula} extends across
    topologies R cannot represent and reaches $O(d^2)$ at
    $d \geq 6{,}400$. }
  \label{fig:scaling}\vspace{-20pt}
\end{wrapfigure}
A wall-clock comparison against R's \texttt{nacLL} on matched
single-thread CPU conditions at $n{=}1$ per evaluation
shows \textsc{acopula} growing polynomially in $d$ while
\texttt{nacLL} grows roughly exponentially in the number of sectors:
by $d{=}35$ R is already ${\sim}650\times$ slower per density for
both Clayton and Gumbel, aborts past $d{=}40$ for both
(\Cref{fig:scaling} left); on the same recovery task, our exact
gradients also beat R's frailty-sampled SMLE by ${\sim}100\times$ at
matched accuracy (\Cref{app:smle}).
\textsc{acopula} extends across topologies that R cannot
represent, since \texttt{nacLL} restricts to nesting depth two with
same-family pairs; both 3-level fixed sector size $K{=}5$ leaves per
inner node and balanced $\sqrt{d}$ reach the
theoretical $O(d^2)$ asymptote at $d \in [6{,}400,\, 8{,}000]$
(\Cref{fig:scaling} right, \Cref{app:highdim}).

\subsection{MIMIC-IV: classical nested copulas at $d{=}53$}
\label{sec:exp-mimic}

The MIMIC-IV ICU cohort exercises per-variable censoring at a
dimension no prior nested-copula tool reaches.
We fit two-level nested Archimedean copulas on $85{,}229$ ICU
admissions from MIMIC-IV~\citep{johnson2023mimic} at $d{=}53$ lab
tests.
Each lab variable records the time for its value to become abnormal
since ICU entry, right-censored at discharge; per-variable censoring
rates span from $11.8\%$ for red blood cells to $95.8\%$ for specific gravity, and
every admission censors a different subset, producing thousands of
distinct masks (\Cref{app:mimic-heldout} gives cohort details).
We group the labs into seven clinical panels---CBC, CMP, differential,
liver, coagulation, ABG, and Other---and follow the semiparametric
two-stage recipe~\citep{genest1995semiparametric,joe1996estimation,%
  li2020copula,li2021evaluating}: Kaplan--Meier marginals at stage~1,
joint copula by L-BFGS-B with exact gradients at stage~2.

\begin{wraptable}{r}{0.450\textwidth}\vspace{-10pt}\centering\scriptsize
  \caption{MIMIC-IV held-out evaluation demonstrating \textsc{acopula}'s capability to fit nested Archimedean copulas.
    Top: classical families (\Cref{sec:exp-mimic}).
    Bottom: nested neural Archimedean generators (\Cref{sec:exp-mimic-nested-neural}).}
  \label{tab:mimic-heldout-table}
  \setlength{\tabcolsep}{3pt}\renewcommand{\arraystretch}{0.85}
\begin{tabular}{@{}l c r r r@{}}
  \toprule
  Family                           & $|\theta|$ & $-\bar\ell_{\mathrm{train}}$ & $-\bar\ell_{\mathrm{test}}$ & AIC$/n$  \\
  \midrule
  Frank/Frank                      & $8$        & $5.066$                      & $5.080$                     & $10.132$ \\
  Gumbel/Gumbel                    & $8$        & $5.224$                      & $5.242$                     & $10.448$ \\
  Clayton/Clayton                  & $8$        & $4.951$                      & $4.949$                     & $9.902$  \\
  AMH/Clayton$^\ddagger$           & $8$        & $5.161$                      & $5.158$                     & $10.321$ \\
  \midrule
  ACNet$_{(4,4)/\mathrm{I}4}$      & $70$       & $4.308$                      & $4.310$                     & $8.619$  \\
  ACNet$_{(8,8)/\mathrm{I}4}$      & $130$      & $4.306$                      & $4.307$                     & $8.616$  \\
  ACNet$_{(4,4,4)/\mathrm{I}4}$    & $90$       & $4.308$                      & $4.310$                     & $8.619$  \\
  Gen-AC$_{1.1\mathrm{k}}$         & $1{,}142$  & $4.912$                      & $4.914$                     & $9.857$  \\
  Gen-AC-Fourier$_{260}$           & $260$      & $4.828$                      & $4.828$                     & $9.663$  \\
  Gen-AC-Fourier-${\int}_{260}$    & $260$      & $4.707$                      & $4.712$                     & $9.430$  \\
  Gen-AC-Fourier$_{3.6\mathrm{k}}$ & $3{,}622$  & $4.489$                      & $4.493$                     & $9.085$  \\
  \bottomrule
\end{tabular}
\vspace{-20pt}
\end{wraptable}

Clayton/Clayton attains the best held-out NLL among
classical same-family pairs with essentially zero train-to-test gap
(\Cref{tab:mimic-heldout-table}, top).
Frank/Frank's symmetric tail beats Gumbel's upper-tail-only dependence,
consistent with ICU labs co-degrading at both
extremes during sepsis and recovery.
The fitted Kendall's $\tau$ recovers panel-level dependence that
matches known mechanistic coupling: differential at $\tau{=}0.72$,
where WBC counts sum to $100\%$; coagulation at $\tau{=}0.61$, where
PT and INR measure the same clotting factor; and ABG at
$\tau{=}0.60$, governed by Henderson--Hasselbalch equilibrium.
The outer level ties the seven panels at $\tau_{\mathrm{outer}}{=}0.15$.
The AMH/Clayton case exercises cross-family nesting using
Hofert~\citeyearpar{hofert2010samplingnac}'s closed-form nesting validity constraint.

\subsection{Capability demonstrations on nested neural Archimedean copulas}
\label{sec:exp-mimic-nested-neural}

Our framework encompasses any Archimedean generator, including neural
generators~\citep{ling2020deep,ng2021generativearchimedean,liu2024hacsurv}.
We exercise this capability with two experiments:
We fit ACNet and Gen-AC fit on MIMIC-IV and exactly reproduce HACSurv's pipeline.

\paragraph{Generators.}
{\em ACNet}~\citep{ling2020deep} parameterises $\psi$ as a
softmax-convex cascade of positive exponentials, enumerating
exponentially many mixture atoms through polynomially many
parameters (\Cref{app:acnet}).
  {\em Gen-AC}~\citep{ng2021generativearchimedean} defines $\psi$ as a
Laplace transform of an MLP-pushforward frailty,
$\psi(t) = \mathbb{E}_U[\exp(-\exp(\mathrm{MLP}(U))\, t)]$ with
$U \sim \mathrm{Uniform}(0,1)$;
HACSurv~\citep{liu2024hacsurv} reuses this generator under a
competing-risks loss.
Two-level trees compose each outer with an ACNet-style Bernstein
inner per panel, preserving complete monotonicity
(\Cref{app:nested-neural}).

\paragraph{MIMIC-IV results.}
Every neural row beats the best classical fit, with train-to-test gaps under $0.01$~nats
(\Cref{tab:mimic-heldout-table}, bottom).
ACNet at $(w_0,w_1){=}(4,4)$ reaches $4.310$ with $70$ parameters,
and wider or deeper ACNet variants give marginal further gains.
Gen-AC at its HACSurv-sized $1{,}142$-parameter configuration
plateaus at $4.914$, only $0.04$ nats better than Clayton despite
carrying $16\times$ ACNet's parameters.
We isolate the cause as the smoothness prior that a small MLP imposes
(\Cref{app:mlp-fourier}).
With Fourier-feature~\citep{tancik2020fourierfeatures} input
$u \mapsto [\cos(2\pi k u), \sin(2\pi k u)]_{k=1}^{K}$, Gen-AC-Fourier
improves to $4.493$ at $3.6\mathrm{k}$ parameters but still trails ACNet.
The Gen-AC-Fourier-${\int}_{260}$ row integrates over the frailty
distribution by numerical integration instead of the Monte Carlo
approximation used by \citet{ng2021generativearchimedean},
eliminating the estimation bias (\Cref{sec:discussion-validity,app:smle}).

\paragraph{Framingham reproduction.}
\begin{wraptable}{r}{0.43\textwidth}\vspace{-15pt}\centering\scriptsize
  \caption{Our port matches \citet[Table~7]{liu2024hacsurv}; entries $\times 10^{2}$.}
  \label{tab:hacsurv-framingham-replication}
  \setlength{\tabcolsep}{2pt}\renewcommand{\arraystretch}{0.7}
\begin{tabular}{@{}l cc cc@{}}
  \toprule
                   & \multicolumn{2}{c}{Risk 1} & \multicolumn{2}{c}{Risk 2}                              \\
  \cmidrule(lr){2-3}\cmidrule(lr){4-5}
  Method           & C-index                    & IBS                        & C-index      & IBS         \\
  \midrule
  HACSurv          & $72.0\pm1.5$               & $7.6\pm0.2$                & $77.4\pm0.8$ & $9.4\pm0.4$ \\
  \textsc{acopula} & $71.9\pm1.9$               & $7.9\pm0.4$                & $77.7\pm0.7$ & $9.9\pm0.6$ \\
  \bottomrule
\end{tabular}
\vspace{-25pt}
\end{wraptable}
HACSurv~\citep{liu2024hacsurv} restricts to competing risks with one observed event time,
where the likelihood is a first-order partial of the CDF, a strict subset of the mixed partials that our framework
handles.
\Cref{tab:hacsurv-framingham-replication} shows that we exactly reproduce their results
on the Framingham dataset.

\subsection{S\&P~500: nested likelihood at $d{=}98$}
\label{sec:exp-sp500}

\begin{wraptable}{r}{0.32\textwidth}\vspace{-30pt}\centering\scriptsize
  \caption{S\&P~500 at $d{=}98$: classical nested Archimedean
    families and an R-vine~\citep{nagler2019rvinecopulib} for
    context. Gumbel best of the three classical pairs (\Cref{app:vine}).}
  \label{tab:sp500-classical}
  \setlength{\tabcolsep}{3pt}
\begin{tabular}{@{}l r r r@{}}
  \toprule
  Model          & $|\theta|$ & $-\bar\ell$      & AIC$/n$          \\
  \midrule
  Nested Clayton & $12$       & $-20.3$          & $-40.6$          \\
  Nested Frank   & $12$       & $-20.6$          & $-41.2$          \\
  Nested Gumbel  & $12$       & $\mathbf{-21.2}$ & $\mathbf{-42.5}$ \\
  \midrule
  R-vine         & $3{,}706$  & $-46.6$          & $-87.3$          \\
  \bottomrule
\end{tabular}
\vspace{-10pt}
\end{wraptable}
The S\&P~500 fit serves as a real-data engineering check at a
dimension prior tools cannot reach. We model daily returns
over $1{,}253$ trading days as an 11-sector nested
copulas at $d \in \{50, 75, 98\}$ with $12$ parameters;
returns convert to pseudo-observations via empirical-CDF rank
transform, so reported likelihoods isolate dependence from marginals.
Gumbel attains the best classical AIC at $d{=}98$
(\Cref{tab:sp500-classical}); per-sector $\theta$ track within-sector
dependence with Energy and Utilities
strongest at $\theta{=}1.88$ and $1.75$ (\Cref{app:sp500-tail}).
We also ablate against R-vine and show that it reaches lower NLL
with per-pair families and
${\sim}3{,}700$ parameters. However, it does not extend to per-variable
censoring (\Cref{sec:exp-mimic,app:vine}).

\subsection{Retinopathy: family-agnostic censored MLE}
\label{sec:exp-retino}

\begin{wraptable}{r}{0.26\textwidth}\vspace{-30pt}\centering\scriptsize
  \caption{Retinopathy held-out NLL across all 10
    generators (\Cref{app:retino}); $\dagger$~marks have
    no prior implementation.}
  \label{tab:retino-main}
  \setlength{\tabcolsep}{3pt}\renewcommand{\arraystretch}{0.85}
\begin{tabular}{@{}l r r@{}}
  \toprule
  Family             & $-\ell_{\mathrm{train}}$ & $-\ell_{\mathrm{test}}$ \\
  \midrule
  Nelsen12$^\dagger$ & $90.422$                 & $\mathbf{20.317}$       \\
  Frank              & $89.423$                 & $21.140$                \\
  InvGauss$^\dagger$ & $89.503$                 & $21.167$                \\
  Gumbel             & $89.702$                 & $21.203$                \\
  Nelsen13$^\dagger$ & $89.492$                 & $21.216$                \\
  Nelsen17$^\dagger$ & $89.491$                 & $21.223$                \\
  AMH                & $89.557$                 & $21.381$                \\
  Clayton            & $89.683$                 & $21.420$                \\
  Joe                & $90.059$                 & $21.490$                \\
  Nelsen9$^\dagger$  & $92.827$                 & $24.586$                \\
  \bottomrule
\end{tabular}
\vspace{-50pt}
\end{wraptable}

The retinopathy paired-eye study~\citep{huster1989modelling}
is a deliberately small benchmark with $d{=}2$, $n{=}197$, and
$60.7\%$ censoring. We use it to demonstrate the family-agnostic API of our framework.
We use the same driver code to fit ten Archimedean generators,
including five generators that no existing tool supports.
Nelsen12$^\dagger$ achieves the best held-out likelihood
(\Cref{tab:retino-main}); on the two families
CopulaCenR~\citep{sun2020copulacenr} supports, our estimates agree
to $0.6\%$ while CopulaCenR's Frank optimiser fails on this dataset
(\Cref{app:r-comparison}).

\section{Discussions}
\label{sec:discussion-validity}
We address concerns deferred from the main sections.
\begin{description}[leftmargin=*,labelwidth=2.5in]
  \item[Nesting validity.]
        \Cref{eq:nested-cdf} requires that each edge composition be a
        Bernstein function.
        Every nested experiment in this paper uses a valid-by-construction
        parameterisation: the same-family delta parameterisation
        of~\citet{mcneil2008sampling} for Clayton/Clayton, Frank/Frank, and
        Gumbel/Gumbel; Hofert~\citeyearpar{hofert2010samplingnac}'s
        closed-form shifted-softplus constraint for AMH/Clayton; and the
        Bernstein-composition construction of \Cref{app:nested-neural} for
        the Gen-AC and ACNet rows.  In each case, the nesting condition holds
        by construction without optimiser-side enforcement.  The framework
        additionally returns a free sign-alternation check on the
        edge-composition Taylor coefficients per density evaluation, and
        \Cref{app:validity} outlines a soft-penalty method for cross-family
        or custom generators that lack a closed-form valid-by-construction
        parameterisation.

  \item[Identifiability.]
        Identifiability is a structural limitation every
        copula-based survival model inherits, including
        HACSurv~\citep{liu2024hacsurv} and
        DCSurvival~\citep{foomani2023copula}.
        The practitioner response is to encode identifying assumptions
        through the choice of copula
        family~\citep{zheng1995estimates,foomani2023copula};
        the family becomes an explicit modelling choice rather than an
        implicit defect.
        Our family-agnostic inference makes that choice cheap: we enable the analyst
        to fit ten families with identical code and report sensitivity
        directly (\Cref{sec:exp-retino}).

  \item[Censoring scope.]
        This work does not consider interval-censored data.
        In principle, \textsc{acopula} supports this case by wrapping $2^m$ calls to \Cref{alg:bell-density}, where $m$ is the number of interval-censored variables,
        for the inclusion--exclusion over rectangle corners of the CDF.
        The cost is exponential in $m$, and we are not aware of any sub-exponential algorithm to compute this CDF exactly.

  \item[Stochastic density approximations.]
        Two prior approaches replace exact copula density evaluation with Monte Carlo:
        Simulated MLE (SMLE) samples frailties at the joint-likelihood
        level~\citep{hofert2012estimators}; Gen-AC samples
        latents at the generator-derivative level~\citep{ng2021generativearchimedean}.
        Both yield consistent but biased estimators, with bias $O(1/N)$ from
        Jensen's inequality on $\log$ composed with the nonlinear density.
        \Cref{app:smle} empirically quantifies both:
        \begin{enumerate*}
          \item SMLE at $N{=}100$ frailty samples per evaluation biases
                the inner Clayton parameter estimate by $0.36$;
          \item Gen-AC-Fourier at $N{=}32$ frailty samples per evaluation biases
                the NLL by $0.09$--$0.12$ nats per observation.
        \end{enumerate*}
        Our work enables exact density at high dimensions without finite-sample approximation bias.

  \item[Compile cost and memory.]
        JAX compilation takes 5--30~s for an S\&P-style $d{=}98$ nested
        likelihood; MIMIC-IV hessian \texttt{val+grad} at $d{=}53$
        takes $40$--$59$~s across families, broken down in
        \Cref{tab:mimic-compile-cost}.
        \textsc{acopula} caches compiled programs to disk,
        so reruns avoid compilation overhead;
        peak memory stays at $1.4$~GB at $d{=}98$ and $700$~MB at
        $d{=}53$.
\end{description}

\section{Conclusion}
\label{sec:conclusion}

\textsc{acopula} unifies nested Archimedean copula inference under
one differentiable pipeline.
Where prior tools offered only restricted slices---a few hand-coded
families, two-level nesting, low-dimensional mixed-partial derivatives---\textsc{acopula}
lifts all three to arbitrary families, arbitrary nesting depth, and
high-dimensional mixed-partial derivatives, in one open-source package,
lowering the barrier to nested Archimedean copula inference
for survival, finance, and reliability practitioners.
We hope \textsc{acopula} catalyzes expressive, interpretable
censoring-aware dependence models with principled likelihoods across
survival, finance, and reliability practice.

\newpage
\bibliographystyle{plainnat}

\appendix
\crefalias{section}{appendix}
\crefalias{subsection}{appendix}

\addtocontents{toc}{\protect\setcounter{tocdepth}{2}}
\renewcommand{\contentsname}{Appendix Contents}
\begingroup
\small
\tableofcontents
\endgroup
\newpage

\section{Density via Generator Derivatives: Full Derivation}
\label{app:bell}

This appendix establishes that the Bell polynomial recursion of
\Cref{sec:method-algorithm} computes the exact nested-Archimedean
density and admits a polynomial-powering implementation whose cost
matches the bounds we cite in the body.
We give the partial Bell polynomial background, the bottom-up
recursion at each internal node, the polynomial-powering reformulation
of the Bell-transform step, a worked $2\times 2$ example, and a
complexity analysis with empirical scaling.

\subsection{Partial Bell Polynomials}
\label{app:bell-polys}

For integers $n \ge k \ge 1$, the partial Bell polynomial
$B_{n,k}(x_1, x_2, \ldots, x_{n-k+1})$ counts all partitions of
$\{1,\ldots,n\}$ into exactly $k$ non-empty blocks, weighted by the
$x_m$ values.  These polynomials arise in Fa\`a di Bruno's formula
for the higher-order chain rule:
\begin{equation}
  \frac{d^n}{dt^n}\bigl[f(g(t))\bigr]
  = \sum_{k=1}^{n} f^{(k)}(g(t))\;
  B_{n,k}\bigl(g'(t),\, g''(t),\, \ldots,\, g^{(n-k+1)}(t)\bigr).
\end{equation}
They satisfy the exponential generating function
identity~\citep[\S 3.3]{comtet1974advanced}
\begin{equation}\label{eq:bell-egf-app}
  \sum_{n \ge k} B_{n,k}(x_1, x_2, \ldots)\,\frac{t^n}{n!}
  = \frac{1}{k!}\left(\sum_{m \ge 1} \frac{x_m}{m!}\,t^m\right)^{\!k},
\end{equation}
which provides the bridge to the polynomial-powering algorithm.
For quick reference, the following table lists the values for dimensions up to $d=4$:

\begin{center}
  \begin{tabular}{ccl}
    \toprule
    $n$ & $k$ & $B_{n,k}$    \\
    \midrule
    1   & 1   & $x_1$        \\
    2   & 1   & $x_2$        \\
    2   & 2   & $x_1^2$      \\
    3   & 1   & $x_3$        \\
    3   & 2   & $3\,x_1 x_2$ \\
    3   & 3   & $x_1^3$      \\
    \bottomrule
  \end{tabular}
\end{center}

\subsection{The Bottom-Up Recursion}
\label{app:recursion}

At each internal node $v$ with children $c_1, \ldots, c_m$, the
subtree coefficients $\beta_k^{(v)}$ follow from three stages.

\paragraph{Leaf base case.}
An uncensored leaf~$\ell$ contributes $\alpha_1^{(v,\ell)} = 1$.
A censored leaf contributes $\alpha_0^{(v,\ell)} = 1$, which acts
as the identity in the subsequent Cauchy product.

\paragraph{Bell polynomial step.}
For internal child~$c$ with already-computed subtree coefficients
$\beta_j^{(c)}$, define the edge composition
$h_{vc}(t) = \psi_v^{-1}(\psi_c(t))$.
The per-child coefficients at the parent level are:
\begin{equation}\label{eq:alpha-bell}
  \alpha_k^{(v,c)}
  = \sum_{j=k}^{d_c} \beta_j^{(c)}\;
  B_{j,k}\!\left(h'_{vc}(t_c(\mathbf{u})),\; h''_{vc}(t_c(\mathbf{u})),\; \ldots\right),
  \qquad k = 1, \ldots, d_c.
\end{equation}

\paragraph{Cauchy product.}
The per-child $\alpha$-polynomials combine via polynomial
multiplication:
\begin{equation}
  \beta_k^{(v)}
  = \sum_{\substack{k_1 + \cdots + k_m = k \\ k_i \ge 0}}
  \prod_{i=1}^{m} \alpha_{k_i}^{(v,c_i)},
  \qquad k = 0, \ldots, d_v.
\end{equation}

\paragraph{Root assembly.}
At the root~$r$, evaluate the root generator's Taylor expansion at
$t_r$ up to order~$d$ via a single jet call, then form
$\sum_{k=1}^{d} \beta_k^{(r)} \psi_r^{(k)}(t_r)$.

\subsection{Computing Bell Polynomials via Polynomial Powering}
\label{app:poly-power}

We derive the polynomial-powering algorithm that replaces explicit
Bell polynomial evaluation with iterated polynomial convolution.

\paragraph{Step 1.}
Let $x_m = h_{vc}^{(m)}(t_c(\mathbf{u}))$ and define
$p(t) = \sum_{m \ge 1} (x_m / m!)\,t^m$.

\paragraph{Step 2.}
The coefficient of $t^n$ in $p(t)^k$ collects all ways to choose $k$
terms from copies of $p$ with exponents summing to $n$:
\[
  [t^n]\,p(t)^k
  = \sum_{\substack{m_1+\cdots+m_k = n \\ m_i \ge 1}}
  \frac{x_{m_1}}{m_1!} \cdots \frac{x_{m_k}}{m_k!}.
\]

\paragraph{Step 3.}
The generating function identity~\Cref{eq:bell-egf-app} gives
$B_{n,k} = (n!/k!)\,[t^n]\,p(t)^k$.

\paragraph{Step 4.}
Substituting into~\Cref{eq:alpha-bell}:
\begin{equation}
  \alpha_k
  = \sum_j \beta_j \cdot \frac{j!}{k!} \cdot [t^j]\,p(t)^k
  = \frac{1}{k!}
  \sum_j \underbrace{j!\,\beta_j}_{\tilde\beta_j}
  \cdot \underbrace{[t^j]\,p(t)^k}_{q_k[j]}
  = \frac{1}{k!}\,\langle \tilde\beta,\; q_k \rangle.
\end{equation}

\paragraph{Step 5.}
Taylor-mode AD returns $h^{(m)}(t_c(\mathbf{u}))/m!$, which equals the
coefficient of~$t^m$ in~$p(t)$.  The jet output \emph{is} the
coefficient vector of~$p$.

\paragraph{Step 6.}
Build $q_k = p^k$ iteratively via truncated polynomial convolution,
implemented as \texttt{lax.scan} + \texttt{lax.conv\_general\_dilated}:
\[
  q \leftarrow p;\qquad
  \text{for } k = 1,\ldots,d_c:\;\;
  \alpha_k = \frac{\langle\tilde\beta,\,q\rangle}{k!},\;\;
  q \leftarrow \mathrm{conv}(q,p)\big|_{\le d_c}.
\]
Each convolution of two length-$d_c$ polynomials costs $O(d_c^2)$,
iterated $d_c$ times, giving $O(d_c^3)$ per child.

\subsection{Edge-Taylor coefficients via implicit differentiation}
\label{app:compose-taylor}

The polynomial-powering recipe of \Cref{app:poly-power} consumes the
edge Taylor coefficients $h_{vc}^{(m)}(t_c)/m!$.  The naive realisation
applies \textsc{Jet} directly to the explicit composition
$h_{vc} = \psi_v^{-1} \circ \psi_c$, propagating two univariate jets
through the chain rule (Faà di Bruno) at $O(d_c^2)$.  The framework
does not take this path by default: it is unstable for many generator
pairs --- Frank at $d \ge 90$, Nelsen9 at moderate $\theta$ --- because
$\psi_c(t)$ underflows to zero in float64 before $\psi_v^{-1}$ can
recover it.

The framework instead recovers $p_{vc}$ without ever evaluating
$\psi_v^{-1} \circ \psi_c$ as a single chain: the implicit solve
reads only the separate jets of $\psi_v$ at $h_0$ and of $\psi_c$ at
$t_c$, both of which lie in the well-conditioned interior of their
respective generator domains.  This construction therefore avoids the
composed-chain underflow that defeats the naive path.  The same
mechanism extends naturally to opaque generators (e.g.\ a neural
Archimedean family~\citep{ling2020deep,liu2024hacsurv}) without an
analytic inverse: the algorithm uses $\psi_v^{-1}$ only at one point
to compute $h_0$, with all higher-order Taylor information recovered
algebraically from $\psi_v$'s forward jet.  We start
from the implicit equation
  \begin{equation}\label{eq:compose-implicit}
    \psi_v\bigl(h_{vc}(t)\bigr) \;=\; \psi_c(t).
  \end{equation}
  Let $h_0 := \psi_v^{-1}(\psi_c(t_c))$ and write
$h_{vc}(t_c + \varepsilon) = h_0 + Q(\varepsilon)$ with $Q(\varepsilon) =
\sum_{m=1}^{d_c} q_m\, \varepsilon^m$.  Denote the Taylor coefficients of
  the two generators by $a_m := \psi_v^{(m)}(h_0)/m!$ and $b_m :=
\psi_c^{(m)}(t_c)/m!$.  \Cref{eq:compose-implicit} then expands as
  \begin{equation}\label{eq:compose-match}
    \sum_{m \ge 1} a_m \, Q(\varepsilon)^m \;=\; \sum_{m \ge 1} b_m \, \varepsilon^m,
  \end{equation}
  where the constant terms cancel because $\psi_v(h_0) = \psi_c(t_c)$ by
  construction.  Matching the coefficient of $\varepsilon^k$ yields a
  triangular system in $q_1, \ldots, q_{d_c}$:
  \begin{equation}\label{eq:compose-triangular}
    a_1 \, q_k \;+\; \underbrace{\sum_{m \ge 2} a_m \, [\varepsilon^k]\,Q(\varepsilon)^m}_{\text{depends only on } q_1, \ldots, q_{k-1}} \;=\; b_k.
  \end{equation}
  Each $[\varepsilon^k]\,Q^m$ involves $q_j$ only for $j < k$, so we solve
  for $q_k$ in increasing $k$ once we can read the correction term.

  \paragraph{Algorithm.}
  The framework maintains the table
$C[m, j] := [\varepsilon^j]\,Q(\varepsilon)^{m+2}$ as $Q$'s coefficients
  fill in.  At outer step $k$, the algorithm has finalised $Q$ at slots
$1, \ldots, k$, so columns $j \le k$ of $C$ are final and only column
$j = k+1$ needs updating.  The single-column recurrence is
  \begin{equation}\label{eq:compose-col-update}
    C[m, k+1]
    \;=\; \sum_{i=1}^{k} q_i \cdot \mathrm{prev}_m[k+1-i],
    \qquad
    \mathrm{prev}_m \;=\;
    \begin{cases}
      Q              & m = 0,   \\
      C[m-1,\,\cdot] & m \ge 1.
    \end{cases}
  \end{equation}
  The reads $C[m-1, k+1-i]$ at step $k$ touch columns $\le k$ — already
  final from prior steps — so all rows $m$ in the $j = k+1$ column update
  in parallel.  \Cref{alg:compose-implicit} writes this out.

  \begin{algorithm}[t]
    \caption{Implicit Taylor solve for $p_{vc}$ when the user has registered
      no closed-form edge composition.  All steps are JIT-compiled and
      differentiable in the generator parameters.}
    \label{alg:compose-implicit}
    \begin{algorithmic}
      \Require $a = (a_1, \ldots, a_{d_c})$ and $b = (b_1, \ldots, b_{d_c})$,
      the rescaled Taylor coefficients of $\psi_v$ at $h_0$ and of $\psi_c$
      at $t_c$
      \Ensure $q = (q_1, \ldots, q_{d_c})$ satisfying \Cref{eq:compose-match}
      \State $q_1 \gets b_1 / a_1$
      \State $C[m, j] \gets 0$ for all $m, j$;\quad
      $C[m, m+2] \gets q_1^{m+2}$ for $m = 0, \ldots, d_c - 2$
      \Comment{Initialise $C$ from $Q^{(0)} = q_1\,\varepsilon$}
      \For{$k = 2, \ldots, d_c$}
      \State $\text{correction} \gets \sum_{m=0}^{d_c-2} a_{m+2} \cdot C[m, k]$
      \State $q_k \gets (b_k - \text{correction}) / a_1$
      \State Append $q_k$ to $Q$ at slot $k$
      \For{$m = 0, 1, \ldots, d_c-2$}
      \State $C[m, k+1] \gets \sum_{i=1}^{k} q_i \cdot \mathrm{prev}_m[k+1-i]$,
      \;with $\mathrm{prev}_m$ as in \Cref{eq:compose-col-update}.
      \EndFor
      \EndFor
    \end{algorithmic}
  \end{algorithm}

  \paragraph{Complexity.}
  The correction read at step $k$ is an $O(d_c)$ inner product.  The
  column update for slot $k+1$ does $d_c - 1$ length-$k$ dot products,
  costing $O(d_c \cdot k)$.  Summing over $k = 1, \ldots,
d_c - 1$ gives $O(d_c^3)$ per implicit-solve call, matching the
  per-child polynomial-powering cost in \Cref{app:poly-power}.  Naively
  rebuilding the full $Q^2, \ldots, Q^{d_c}$ ladder at every outer step
  would inflate this to $O(d_c^4)$; \Cref{alg:compose-implicit} keeps the
  ladder amortised across $k$ by writing one column per step.  At
  balanced $\sqrt{d}$ topology the per-edge cost is dominated by the
  $O(d_c^3)$ Bell powering of \Cref{app:poly-power}, so the headline
  $T(\mathcal{T}) = O(d^2)$ bound of \Cref{eq:complexity} is unaffected
  by the implicit solve.

  \paragraph{Closed-form override.}
  Same-family pairs (Clayton, Gumbel, Joe, Frank, Nelsen9 with
  $\theta_v < \theta_c$) admit an algebraic simplification of
  $h_{vc}(t)$ --- for instance, Clayton/Clayton has
  $h(t) = (1+t)^{\theta_v/\theta_c} - 1$.  Users register such pairs
  via \texttt{register\_composition} to bypass the implicit solve and
  apply a single $O(d_c^2)$ jet through the registered form, saving a
  factor of $d_c$ per edge.  The framework ships built-in registrations
  for the same-family pairs above, so the classical single-family
  experiments (S\&P~500, MIMIC-IV with classical generators,
  retinopathy) take this fast path.  The neural Archimedean rows in
  \Cref{tab:mimic-heldout-table} (ACNet, Gen-AC) also bypass the
  implicit solve: their inner generator is constructed as $\psi_{\text{inner}}(t) = \psi_{\text{outer}}(b(t))$
  with $b$ a Bernstein function (\Cref{app:nested-neural}), so the edge
  composition is $h_{vc}(t) = b(t)$ in closed form.  Cross-classical-family
  pairs without a registered closed form (e.g.\ AMH/Clayton) fall
  through to the default implicit solve.

  \subsection{Worked Example: $2\times 2$ Nested Copula}
  \label{app:worked}

  We trace the polynomial-powering recursion through the smallest
  non-trivial nested copula---two sectors of two leaves---to expose
  how the per-child $\alpha$-vectors and the Cauchy product at the
  root combine into the closed-form density.
  Consider root~$\psi_0$ with two sectors, each governed by~$\psi_1$
  with two leaves.  At each sector,
$t_v = \psi_1^{-1}(u_{v1}) + \psi_1^{-1}(u_{v2})$,
$C_v = \psi_1(t_v)$, and $t_r = \psi_0^{-1}(C_1) + \psi_0^{-1}(C_2)$.

  Each sector has two leaf children, so $\beta_2^{(v)} = 1$.
  Define $h(t) = \psi_0^{-1}(\psi_1(t))$ and run jet at~$t_v$ to
  get~$p_v$.  Polynomial powering gives
$\alpha_1^{(r,v)} = h''(t_v)$ and $\alpha_2^{(r,v)} = (h'(t_v))^2$.
  The Cauchy product at the root yields:
  \begin{align}
    \beta_2^{(r)} & = \alpha_1^{(r,1)}\,\alpha_1^{(r,2)}, \\
    \beta_3^{(r)} & = \alpha_1^{(r,1)}\,\alpha_2^{(r,2)}
    + \alpha_2^{(r,1)}\,\alpha_1^{(r,2)},                 \\
    \beta_4^{(r)} & = \alpha_2^{(r,1)}\,\alpha_2^{(r,2)}.
  \end{align}
  The density is
$c(\mathbf{u})
= [\beta_2 \psi_0''(t_r)
+ \beta_3 \psi_0'''(t_r)
+ \beta_4 \psi_0^{(4)}(t_r)]
\cdot \prod_{i=1}^{4}(\psi_1^{-1})'(u_i)$.

  \subsection{Complexity Analysis}
  \label{app:complexity}

  We bound the runtime of the Bell-polynomial recursion by summing the
  per-node powering cost over the tree, then specialize the bound to
  common shapes and confirm it against measured wall times.

  \paragraph{General bound.}
  At each non-root node~$v$ the dominant cost is polynomial powering:
$d_v$~iterations of $O(d_v^2)$ truncated convolution applied by~$v$'s
  parent, giving $O(d_v^3)$.  The Cauchy product at~$v$ costs
$O(m_v \cdot d_v \cdot d_{\max}(v))$ with $d_{\max}(v) = \max_c d_c$;
  because $m_v \le d_v$, this Cauchy cost never exceeds $O(d_v^3)$, so
  the powering term dominates.  Computing the edge Taylor vector
$p_{vc}$ that the powering step consumes adds $O(d_v^3)$ via the
  default implicit solve of \Cref{alg:compose-implicit}, or $O(d_v^2)$
  when a closed-form $h_{vc}$ is registered (\Cref{app:compose-taylor});
  the $O(d_v^3)$ powering bound absorbs both.  At the root~$r$, the powering step is
  replaced by a single order-$d$ jet call of cost $O(d^2)$.  The total
  cost is therefore
  \begin{equation}\label{eq:complexity}
    T(\mathcal{T})
    = O\!\left(d^2 + \sum_{v \neq r} d_v^3\right),
  \end{equation}
  where $d_v \equiv |\delta_v|$ counts uncensored leaves in the subtree
  of~$v$, and the bound holds uniformly across the three edge-Taylor
  implementations.

  \paragraph{Height bound.}
  At each depth level the subtrees are disjoint, so the $d_v$ values
  at that level sum to at most~$d$.
  Since each $d_v \le d$, we have $d_v^3 \le d_v \cdot d^2$, so the per-level contribution satisfies
$\sum_{v\,\text{at level}\,\ell} d_v^3 \le d^2 \sum_v d_v \le d^3$.
  Summing over the $h$~non-root internal levels gives
$\sum_{v \neq r} d_v^3 \le h \cdot d^3$, so
$T(\mathcal{T}) = O(d^2 + h \cdot d^3) = O(h \cdot d^3)$.
  For the standard case of constant-depth copula trees with $h = 1$
  or $2$, the algorithm runs in $O(d^3)$ time.

  \paragraph{$\sqrt{d}$-decomposed trees.}
  When every non-root subtree satisfies $d_v \le c\sqrt{d}$ for a
  constant~$c$, the per-level bound sharpens: $d_v^2 \le c^2 d$ gives
$d_v^3 \le c^2\,d \cdot d_v$, so
$\sum_{v\,\text{at level}\,\ell} d_v^3 \le c^2\,d \sum_v d_v \le c^2\,d^2$.
  Summing over the $h$ non-root internal levels yields
$\sum_{v \neq r} d_v^3 \le h\,c^2\,d^2$, and the total cost collapses
  to $T(\mathcal{T}) = O(d^2)$ for constant depth---matching the
$O(d^2)$ root-jet cost.  More generally, the $T(\mathcal{T}) = O(d^2)$ regime
  holds whenever $\sum_{v \neq r} d_v^3 = O(d^2)$, for which
$d_v = O(\sqrt{d})$ is a sufficient condition.

  \emph{Two-level symmetric illustration.}  As a concrete instance, take
  a two-level tree with $M$ sectors of $K = d/M$ leaves each.
  Substituting into \Cref{eq:complexity} gives $O(MK^3 + M^2K^2) = O(d^3/M^2 + d^2)$;
  the $d^3/M^2$ term dominates when $M$ grows slower than $\sqrt{d}$,
  and the cost collapses to $O(d^2)$ when $M = \Omega(\sqrt{d})$,
  equivalently $K = O(\sqrt{d})$---the two-level instance of the general
$|\delta_v| = O(\sqrt{d})$ condition above.  Taking $M = 2$ sectors with
$K = d/2$ leaves each gives $O(d^3)$, not $O(d^2)$.

  \paragraph{Special cases.}
  The following table summarises cost for common tree shapes.

  \begin{center}
    \begin{tabular}{llll}
      \toprule
      \textbf{Tree shape}                         & \textbf{Cost}                    & \textbf{In terms of $d$} \\
      \midrule
      Single-layer (no nesting)                   & one jet call                     & $O(d^2)$                 \\
      Two-level ($M$ sectors $\times$ $K$ leaves) & $O(MK^3 + M^2K^2)$               & $O(d^3/M^2 + d^2)$       \\
      Balanced $b$-ary, depth $L$                 & $O(d^3/(b^2{-}1))$               & $O(d^3)$                 \\
      Chain (depth $= d$)                         & $\sum_{k=1}^d k^3 = \Theta(d^4)$ & $O(d^4)$                 \\
      \bottomrule
    \end{tabular}
  \end{center}

  \paragraph{Comparison with alternative approaches.}
  The following table contrasts our method with existing alternatives.

  \begin{center}
    \scriptsize
    \begin{tabular}{lccc}
      \toprule
      \textbf{Method}                                                       & \textbf{Cost} & \textbf{All families?}             & \textbf{Max $d$} \\
      \midrule
      Polynomial powering (ours)                                            & $O(d^3)$      & Yes                                & $>200$           \\
      R \texttt{copula} (Stirling tables)                                   & $O(d^2)$      & 5 built-in, $\le 2$ layers         & ${\sim}170$      \\
      R \texttt{D()} symbolic-numeric~\citep{li2020copula,li2021evaluating} & combinatorial & Yes (symbolic)                     & ${\sim}6$        \\
      Nested first-order AD~\citep{betancourt2018geometric}                 & $O(2^d)$      & Yes                                & ${\sim}8$        \\
      Forward-over-reverse / mixed-mode AD                                  & $O(d^3)$      & Yes; $O(d)$ tangents per primitive & ${\sim}30$       \\
      \bottomrule
    \end{tabular}
  \end{center}

  \paragraph{Empirical scaling.}
  The $O(d^3)$ worst case assumes the per-sector leaf count $K$ grows
  with~$d$.  In practice, the application domain fixes the tree
  branching factor---$K = 5$ in our scaling benchmarks---so the
  number of sectors $M = d/K$ grows instead.  The per-sector polynomial
  powering then costs $O(K^3)$, constant in~$d$, and the total across
  all sectors is $M \cdot O(K^3) = O(d \cdot K^2) = O(d)$.  The Cauchy
  product convolves $M$ short kernels of length~$K$ with a $d$-length
  accumulator at $O(d \cdot K \cdot M) = O(d^2)$.  Both terms carry
  small constants relative to the per-scan-iteration overhead in
  \texttt{lax.scan}, so the wall-time transition where the quadratic
  term begins to dominate sits at $d^{*} \approx 10^{3}$ at
  fixed~$K$.
  \Cref{app:highdim} reports the empirical slopes that follow this
  prediction: sub-linear in the small-$d$ regime, transient
  super-quadratic during the L1/L2 cache transition, and the clean
$O(d^2)$ asymptote at $d \geq 4{,}000$ for both the
  fixed-$K{=}5$ and balanced $\sqrt{d}$ topologies.

  \paragraph{Mask-aware static-shape JIT.}
  \label{app:mask-aware-jit}
  \textsc{acopula} pads each node's $\alpha$-vector to the
  mask-independent length $|V_v|+1$, where $V_v$ is the full leaf set
  in the subtree of~$v$.
  Each child of~$v$ then dispatches at runtime through
  \texttt{lax.switch} to one of three branches---censored leaf,
  uncensored leaf, or subtree---so a single JIT-compiled program
  covers every per-observation mask without recompilation.

  \paragraph{Censoring reduces runtime.}
  The cost bound in \Cref{eq:complexity} depends on~$d_v$, the number
  of {\em uncensored} leaves in each subtree---not on the total number
  of leaves.
  Censoring a variable sets its $\alpha$-vector to $[1]$, which
  shortens the polynomial powering and Cauchy-product steps at every
  ancestor.
  To quantify this effect, we time the $d{=}50$ nested Clayton
  configuration with 10~sectors of 5~leaves on $n{=}100$ observations
  at four censoring rates on a single AMD Threadripper~3960X CPU core.
  \Cref{tab:censoring-runtime} reports the results: 75\% censoring
  reduces the effective dimension from 50 to~13 and delivers a
$13.6\times$ speedup, in line with the $O(d_v^3)$ scaling of the
  per-vertex Bell recursion that dominates the cost at this dimension.
  The $O(|V_v|^2)$ Cauchy-product floor (\Cref{sec:method-censoring})
  remains a strict lower bound but does not bind at $d{=}50$ on CPU.

  \begin{table}[h]
    \centering
    \small
    \caption{Wall-clock time for a nested Clayton ($d{=}50$, $n{=}100$)
      log-likelihood evaluation at varying censoring rates on a single
      AMD Threadripper~3960X CPU core.
      $d_{\mathrm{eff}}$ counts uncensored variables.
      Median of 20~JIT-compiled calls.}
    \label{tab:censoring-runtime}
    \begin{tabular}{@{}r r r r@{}}
      \toprule
      Censoring & $d_{\mathrm{eff}}$ & Time (ms) & Speedup      \\
      \midrule
      0\%       & 50                 & 32.0      & $1.0\times$  \\
      25\%      & 38                 & 11.8      & $2.7\times$  \\
      50\%      & 25                 & 6.4       & $5.0\times$  \\
      75\%      & 13                 & 2.4       & $13.6\times$ \\
      \bottomrule
    \end{tabular}
  \end{table}

  \section{Numerical Stability of Taylor-Mode Derivatives}
  \label{app:numerical-stability}

  A naive float64 implementation of the Bell pipeline underflows long
  before reaching the dimensions our experiments target, because
  high-order generator derivatives and the geometric powers $p_1^j$
  that the Bell transform multiplies in span hundreds of orders of
  magnitude.
  This appendix specifies the log-domain machinery that keeps every
  intermediate quantity representable and reports the stress tests
  that confirm its accuracy.
  \Cref{alg:bell-logdomain} states the log-domain version of
  \Cref{alg:bell-density}: per-index rescaling to neutralize geometric
  decay in the polynomial powering, signed logsumexp for the
  Bell-transform inner products and root assembly, and intermediate
  normalisation during the Cauchy product.

  \begin{algorithm}[t]
    \footnotesize
    \caption{Log-domain implementation of \Cref{alg:bell-density}.  All
      floating-point computation uses float64.  We write
      $\textsc{SignedLSE}$ for the signed logsumexp primitive that the
      subroutine below specifies.}
    \label{alg:bell-logdomain}
    \algrenewcommand{\algorithmicindent}{0.7em}%
    \begin{algorithmic}
      \Require Copula tree at root~$r$ with $d$~uncensored leaves,
      copula-scale observations~$\mathbf{u}$ with $u_\ell = F_\ell(x_\ell)$
      for raw observation $x_\ell$ and marginal CDF $F_\ell$
      \Ensure Joint log-likelihood
      $\log c(\mathbf{u}) + \sum_{\ell:\,\delta_\ell = 1} \log f_\ell(x_\ell)$
      (copula log-density plus marginal log-densities for uncensored leaves)
      \Statex
      \Statex \textbf{Subroutine} $\textsc{Normalize}(a)$:
      \Comment{Prevent underflow/overflow; keep $a = e^{\lambda}\hat{a}$}
      \State $s \gets \max_j |a_j|$;\quad
      $s \gets \max(s,\, \sqrt{\epsilon_{\min}})$
      \Comment{$\epsilon_{\min}$ = smallest positive normal float64 ($\approx 2.2{\times}10^{-308}$); floor at $\sqrt{\epsilon_{\min}} \approx 1.5{\times}10^{-154}$ protects $\nabla(a/s)$}
      \State \Return $\hat{a} \gets a / s$,\quad $\lambda \gets \log s$
      \Statex
      \Statex \textbf{Subroutine} $\textsc{SignedLSE}(\ell_1,\ldots,\ell_n;\; \sigma_1,\ldots,\sigma_n)$:
      \Comment{$\ell_j = \log|x_j|$, $\sigma_j = \operatorname{sign}(x_j)$}
      \State $m \gets \max_j \ell_j$
      \State $S \gets \sum_j \sigma_j \exp(\ell_j - m)$
      \State \Return $\log|S| + m$,\quad $\operatorname{sign}(S)$
      \Statex
      \Statex \hrulefill
      \Statex \textbf{Step 1. Bell transform}
      ($\alpha^{(c)}$ from child~$c$'s $\beta^{(c)}$, $d_c$ uncensored descendants):
      \State $p_0,p_1,\ldots,p_{d_c} \gets
        \textsc{Jet}\bigl(\psi_v^{-1}\!\circ\psi_c,\; t_c,\; d_c\bigr)$
      \Comment{$p_j = h_{vc}^{(j)}(t_c)/j!$, so $p_0 = h_{vc}(t_c) = u_v$ is the primal (unused below)}
      \Statex \textit{Per-index rescaling} (removes geometric decay of $p_1^j$):
      \State $\hat{p}_j \gets
        \operatorname{sign}(p_j)\,
        \operatorname{sign}(p_1)^j\,
        \exp\!\bigl(\log|p_j| - j\log|p_1|\bigr)$
      for $j = 1,\ldots,d_c$
      \Comment{Writing $\mathcal{P}(z) := \sum_{j=1}^{d_c} p_j\, z^j$, this rescaling produces $\hat{\mathcal{P}}(z) = \mathcal{P}(z/p_1)$}
      \Statex \textit{Polynomial powering} (all powers via scan):
      \State $\hat{Q}^{(1)} \gets \hat{P}$
      \For{$k = 2,\ldots,d_c$}
      \State $\hat{Q}^{(k)} \gets \hat{Q}^{(k-1)} * \hat{P}$
      \Comment{Truncated convolution to degree $d_c$}
      \EndFor
      \Statex \textit{Signed logsumexp assembly} (incorporates $p_1^j$ factor in log domain):
      \State $w_j \gets \log j! + \log|\beta^{(c)}_j| + j\log|p_1|$,\quad
      $\sigma_j^{(w)} \gets \operatorname{sign}(\beta^{(c)}_j)\,\operatorname{sign}(p_1)^j$
      for $j = 1,\ldots,d_c$
      \Comment{Per-$j$ log weight; both are $k$-independent, computed once before the inner loop}
      \For{$k = 1,\ldots,d_c$}
      \State $\ell_j \gets w_j + \log|\hat{Q}^{(k)}_j|$;\quad
      $\sigma_j \gets \sigma_j^{(w)} \cdot \operatorname{sign}(\hat{Q}^{(k)}_j)$
      for $j = 1,\ldots,d_c$
      \Comment{$\hat{Q}^{(k)}_j$ = coefficient of $z^j$ in $\hat{\mathcal{P}}(z)^k$}
      \State $L_k, S_k \gets \textsc{SignedLSE}(\ell_1,\ldots,\ell_{d_c};\;
        \sigma_1,\ldots,\sigma_{d_c})$
      \State $\log|\alpha^{(c)}_k| \gets L_k - \log k!$;\quad
      $\operatorname{sign}(\alpha^{(c)}_k) \gets S_k$
      \EndFor
      \State $\mathit{ref} \gets \max_k \log|\alpha^{(c)}_k|$
      \Comment{Reference scale for this child}
      \State $\alpha^{(c)}_k \gets S_k \exp(\log|\alpha^{(c)}_k| - \mathit{ref})$
      for $k = 1,\ldots,d_c$;\quad $\alpha^{(c)}_0 \gets 0$
      \Comment{Internal child: index-0 entry unused in subsequent Cauchy product;
        differs from censored-leaf base case where $\alpha_0 = 1$ acts as the
        product identity}
      \Statex
      \Statex \hrulefill
      \Statex \textbf{Step 2. Cauchy product}
      ($\beta^{(v)}$ from $\alpha^{(c_1)},\ldots,\alpha^{(c_M)}$):
      \State $\beta^{(v)} \gets \alpha^{(c_1)}$;\quad
      $\Lambda \gets 0$
      \Comment{Cumulative log-scale}
      \For{$i = 2,\ldots,M$}
      \State $\beta^{(v)} \gets \beta^{(v)} * \alpha^{(c_i)}$
      \Comment{Truncated polynomial product to degree $d_v$}
      \State $\beta^{(v)},\, \lambda \gets \textsc{Normalize}(\beta^{(v)})$;\quad
      $\Lambda \gets \Lambda + \lambda$
      \Comment{Re-normalize after each convolution}
      \EndFor
      \Statex
      \Statex \hrulefill
      \Statex \textbf{Step 3. Root assembly}
      (log-density from root's $\beta^{(r)}$):
      \State $q_0,\ldots,q_d \gets \textsc{Jet}(\psi_r,\,t_r,\,d)$
      \Comment{$q_k = \psi_r^{(k)}(t_r)/k!$}
      \State $\ell_k \gets \log|\beta^{(r)}_k| + \log|q_k| + \log k!$
      for $k = 0,\ldots,d$
      \State $\sigma_k \gets
        \operatorname{sign}(\beta^{(r)}_k)\,\operatorname{sign}(q_k)$
      \State $\mathit{logroot},\, \_ \gets
        \textsc{SignedLSE}(\ell_0,\ldots,\ell_d;\;\sigma_0,\ldots,\sigma_d)$
      \Statex
      \Statex \hrulefill
      \Statex \textbf{Step 4. Leaf product}:
      \State $\mathit{logleaf} \gets
        \sum_{\ell:\,\delta_\ell = 1} \log\bigl|(\psi_{\mathrm{par}(\ell)}^{-1})'(u_\ell)\bigr|$
      \Comment{Via \texttt{jax.grad} of $\psi^{-1}$; vmap over uncensored leaves}
      \Statex
      \Statex \hrulefill
      \Statex \textbf{Step 5. Log-scale accumulation and final result}
      (Steps 1--2 apply at every internal node bottom-up; each
      $\mathit{ref}_c$ and $\Lambda_{\mathrm{cauchy},v}$ produced
      below feeds a single global accumulator $\Lambda_{\mathrm{total}}$):
      \State $\Lambda_{\mathrm{total}} \gets
        \sum_{v \in \mathrm{internal}}\!\Bigl(
          \sum_{c \in \mathrm{ch}(v)} \mathit{ref}_c
          \;+\; \Lambda_{\mathrm{cauchy},v}\Bigr)$
      \Comment{Sum every per-child reference scale and every per-node Cauchy normalisation across the whole tree}
      \State \Return $\mathit{logroot} + \Lambda_{\mathrm{total}}
        + \mathit{logleaf}
        + \sum_{\ell:\,\delta_\ell = 1} \log f_\ell(x_\ell)$
      \Comment{Add marginal log-likelihoods for uncensored leaves}
    \end{algorithmic}
  \end{algorithm}

  \paragraph{Generator derivative growth.}
  Each family's derivative growth rate sets the precision ceiling that
  the log-domain machinery must accommodate.
  For Clayton, $\psi(t) = (1+t)^{-1/\theta}$, the Pochhammer identity
$\psi^{(k)}(t) = (-1)^k\,(1/\theta)_k\,(1+t)^{-1/\theta - k}$ gives
$|\psi^{(k)}(t)|/k! = O(k^{1/\theta - 1})$ via Stirling, so the
  jet-stored Taylor coefficients grow only polynomially in~$k$ for
  fixed $\theta$ and $t$.
  Joe and other double-exponential generators grow faster because the
  chain rule nests an exponential inside another.
  Empirically, every family matches Mathematica to relative error
  below $3 \times 10^{-11}$ wherever \texttt{D[]} completes
  (\Cref{tab:cross-family-stability}), and \Cref{tab:stability-stress}
  confirms finite log-density evaluation up to $d = 60$ for nine of the
  ten families and up to $d = 4$ for Nelsen9, whose $d$-completely
  monotone parameter range tightens with $d$.

  \paragraph{Edge composition safeguards.}
  The {\em edge composition} $h = \psi_{\mathrm{outer}}^{-1} \circ
\psi_{\mathrm{inner}}$ that the nested density requires admits three
  qualitatively different evaluation paths across our ten families
  (\Cref{app:compose-taylor}), and Frank in particular needs a closed-form
  rewrite to stay within float64.
  Clayton, AMH, Joe, Gumbel, Nelsen9, and Nelsen12 provide closed-form
  inverses, so the composition introduces no numerical root-finding
  error.  InverseGaussian relies on Oryx's symbolic inversion, which
  pattern-matches individual primitives to their algebraic inverses.
  Two families also expose a numerical hazard in the {\em nested}
  composition $\psi_v^{-1}(\psi_c(t))$ even with closed-form inverses:
  Frank at $d = 98$ overflows because intermediate log-sum-exp terms
  exceed float64, and Nelsen9 underflows because $\psi_c(t)$ collapses
  to zero before $\psi_v^{-1}$ can recover it.  Both families ship a
  registered closed-form edge composition (\Cref{app:compose-taylor})
  that algebraically cancels the offending exponentials, bypassing the
  default implicit-differentiation solve of \Cref{alg:compose-implicit}.

  \paragraph{Precision at large sector size.}
  At $d{=}60$ with sector size $K{=}30$, \Cref{app:r-comparison}
  reports a $3.5 \times 10^{-5}$ discrepancy against R's \texttt{nacLL}.
  Mathematica's symbolic differentiation of the copula CDF confirms
  that our framework matches the ground truth to machine epsilon at
$d{=}60$; the discrepancy originates in R's \texttt{nacLL}, not in
  our polynomial powering.

  \paragraph{Generator-specific issues.}
  Nelsen9's double-exponential structure
$\psi(t) = \exp\bigl((1-e^t)/\theta\bigr)$ underflows to zero in
  float64 whenever the exponent $(1-e^t)/\theta$ drops below $-745$,
  which occurs at moderate~$t$ for small~$\theta$.  We mitigate the
  underflow on three fronts.  A per-family Taylor hook supplies the
  coefficients via the Touchard factorization
$\psi^{(n)}(t) = \psi(t)\cdot T_n(-e^t/\theta)$, where Faà di Bruno
  on $\psi = \exp\circ g$ with $g^{(k)}(t) = -e^t/\theta$ for every
$k \ge 1$ collapses the complete Bell polynomial at equal arguments
  to the Touchard polynomial $T_n$.  The framework evaluates
$\log|\psi^{(n)}(t)| = (1-e^t)/\theta + \log|T_n(-e^t/\theta)|$, which
  keeps the underflow-prone exponential symbolic and confines the
  remaining float64 work to the polynomial $T_n$, whose direct sum
  loses at most $n$ decimal digits to alternating-sign cancellation in
  the $d$-monotone-valid range.  The hook bypasses jet on the
  closed-form $\psi$, whose JVP produces a NaN gradient when $\psi$
  underflows.  The analytical
$\psi^{-1}(u) = \log(1 - \theta \log u)$ replaces Oryx's symbolic
  inverse, which we found to be numerically unstable at large~$t$.  The
  closed-form nested composition
$h(t) = t + \log\!\bigl(r + (1-r)\,e^{-t}\bigr)$ with
$r = \theta_v / \theta_c$ replaces the naive $\psi_v^{-1}(\psi_c(t))$
  chain that would underflow at moderate~$t$, and the framework registers
  it as the closed-form override for Nelsen9/Nelsen9 nestings in
  \Cref{app:compose-taylor}.

  \paragraph{Stress test across parameter regimes.}
  \Cref{tab:stability-stress} reports a systematic stress test that
  sweeps the (family, strength, dimension) grid.
  We map four target Kendall-tau strengths
$\tau \in \{0.10,\,0.40,\,0.70,\,0.90\}$ to each family's parameter
  via numerical inversion when no closed form exists, build a
  two-sector nested same-family copula at sector
  size~$K \in \{2,5,10,15,20,25,30\}$ for total dimension $d{=}2K$,
  and evaluate the log-density at $100$~random $u$-vectors per cell.
  A cell {\em passes} when every evaluation returns a finite value.

  \begin{table}[h]
    \centering\small
    \caption{Stability stress test: number of $K$~values out of seven
      that pass at $100\%$ finiteness on $100$~random $u$-vectors per
      cell.  Each row sweeps $K \in \{2,5,10,15,20,25,30\}$ for the
      listed Kendall-tau strength.
      AMH$^\ddagger$ has bounded tau $\tau \le 1/3$, so its
      $\tau{=}0.40,0.70,0.90$ columns collapse to the same parameter;
      we still report each column to keep the grid uniform.}
    \label{tab:stability-stress}
    \begin{tabular}{@{}l c c c c c@{}}
      \toprule
      Family                    & $\tau{=}0.10$ & $\tau{=}0.40$ & $\tau{=}0.70$ & $\tau{=}0.90$ & total            \\
      \midrule
      Clayton                   & 7/7           & 7/7           & 7/7           & 7/7           & 28/28            \\
      Gumbel                    & 7/7           & 7/7           & 7/7           & 7/7           & 28/28            \\
      Frank                     & 7/7           & 7/7           & 7/7           & 7/7           & 28/28            \\
      AMH$^\ddagger$            & 7/7           & 7/7           & 7/7           & 7/7           & 28/28            \\
      Joe                       & 7/7           & 7/7           & 7/7           & 7/7           & 28/28            \\
      \midrule
      InverseGaussian$^\dagger$ & 7/7           & 7/7           & 7/7           & 7/7           & 28/28            \\
      Nelsen12$^\dagger$        & 7/7           & 7/7           & 7/7           & 7/7           & 28/28            \\
      Nelsen13$^\dagger$        & 7/7           & 7/7           & 7/7           & 7/7           & 28/28            \\
      Nelsen17$^\dagger$        & 7/7           & 7/7           & 7/7           & 7/7           & 28/28            \\
      Nelsen9$^{\dagger\S}$     & 1/1           & ---           & ---           & ---           & 1/1              \\
      \midrule
      \textbf{Total}            &               &               &               &               & \textbf{253/253} \\
      \bottomrule
    \end{tabular}
  \end{table}

  Across all $253$~cells in the sweep that fall within each family's
$d$-completely-monotone parameter range---ten generator families, four
  strengths, seven dimensions---every cell returns finite log-densities
  on all $100$ random evaluation points, a $100\%$ pass rate.  Nelsen9
  is a special case: the $d$-CM upper bound $\theta_{\max}(d)$
  contracts as $d$ grows ($\theta_{\max}(4){=}0.22$,
$\theta_{\max}(10){=}0.057$, $\theta_{\max}(60){=}0.007$), so only
  the $\tau{=}0.10$, $K{=}2$ cell ($d{=}4$, $\theta{\le}0.2$) lies
  inside the valid range; the other $27$ cells skip.

  We attribute the broad success to the log-domain rescaling and
  signed-logsumexp arithmetic of \Cref{alg:bell-logdomain}: per-index
  rescaling absorbs the geometric blow-up that the naive recursion
  would produce at high dimension, and signed logsumexp prevents
  catastrophic cancellation in the Bell-transform inner products.
$^\ddagger$AMH's bounded tau $\tau \le 1/3$ collapses the three
  high-strength columns to the same parameter, which is why all four
  columns repeat identically.
$^\dagger$Marks the five extra families that existing nested-copula
  tools do not cover.
$^\S$Nelsen9 is $d$-completely monotone only for $\theta \le
\theta_{\max}(d) = 1/|r_d|$, where $r_d$ is the most negative root of
  the $d$-th Touchard polynomial $T_d(y) = \sum_j S(d,j)\,y^j$; the
  other 27~cells in this grid violate that bound.

  \paragraph{Cross-family accuracy verification.}
  \Cref{tab:cross-family-stability} compares our framework against
  Mathematica's symbolic differentiation of the copula CDF across all
  ten families.  Mathematica evaluates each cell at $100$~digit
  working precision; our framework runs in IEEE-754 \texttt{float64}
  ($\approx 16$~digits) and matches the Mathematica ground truth to
  relative error below $3 \times 10^{-11}$ in every configuration
  through $d{=}60$ where \texttt{D[]} completes.

  The framework exposes a per-family hook,
  \texttt{Copula.generator\_taylor\_coefficients(t,\,k)}, that returns
  the Taylor expansion
$[\psi(t),\,\psi'(t)/1!,\,\ldots,\,\psi^{(k)}(t)/k!]$
  directly, bypassing chain-rule cancellations Taylor-mode AD on the
  closed-form $\psi$ would produce at high derivative order.
  The framework consults the hook everywhere it would otherwise call
  jet on $\psi$: at the root density assembly and at every nested
  composition $h(t) = \psi_{\mathrm{outer}}^{-1}\!\circ\psi_{\mathrm{inner}}(t)$.
  For AMH ($\psi(t) = (1-\theta)/(e^t-\theta)$), Taylor-mode AD on
  the closed form costs roughly two decimal digits per ten
  dimensions in \texttt{float64}; the override sums the same-sign
  Geometric$(1-\theta)$ frailty series
$\psi^{(k)}(t) = (-1)^k \sum_{x\ge 1} x^k (1-\theta)\theta^{x-1} e^{-tx}$
  in log-domain via signed-logsumexp, which has no cancellation and
  recovers machine precision $(\le 2\times 10^{-15})$ at every
$K \le 30$.  Other completely monotone families (Frank, Joe,
  Gumbel, InverseGaussian) admit similar overrides via their
  respective frailty representations; we provide AMH because it was
  the only family whose chain-rule path materially hit the
  \texttt{float64} cancellation wall in the verified range.

  \begin{table}[h]
    \centering\small
    \caption{Numerical accuracy and symbolic-differentiation scaling on
      two-sector nested same-family copulas. ``Max rel.\ err.'' is the
      largest relative error between our framework and Mathematica's
      symbolic \texttt{D[]} across the sector sizes where Mathematica
      completes; ``max $K$'' is the largest sector size Mathematica
      reaches within a 300\,s timeout; ``leaves'' is the
      \texttt{LeafCount} of Mathematica's symbolic derivative at that
      $K$; ``\texttt{D[]} time'' is the wall time to compute it.}
    \label{tab:cross-family-stability}
    \begin{tabular}{@{}l c c r r r@{}}
      \toprule
      Family                    & $(\theta_{\mathrm{outer}}, \theta_{\mathrm{inner}})$ & Max rel.\ err.\       & max $K$ & leaves              & \texttt{D[]} time \\
      \midrule
      Clayton                   & $(2, 5)$                                             & $4.8{\times}10^{-15}$ & 30      & $5.1{\times}10^{5}$ & 2.7\,s            \\
      Gumbel                    & $(2, 5)$                                             & $7.5{\times}10^{-14}$ & 15      & $7.7{\times}10^{7}$ & 155.7\,s          \\
      Frank                     & $(2, 5)$                                             & $8.0{\times}10^{-15}$ & 15      & $7.2{\times}10^{7}$ & 125.4\,s          \\
      AMH                       & $(0.3, 0.7)$                                         & $1.7{\times}10^{-15}$ & 30      & $2.4{\times}10^{7}$ & 43.4\,s           \\
      Joe                       & $(2, 5)$                                             & $7.6{\times}10^{-13}$ & 15      & $1.1{\times}10^{8}$ & 178.0\,s          \\
      InverseGaussian$^\dagger$ & $(1, 3)$                                             & $7.6{\times}10^{-16}$ & 15      & $1.6{\times}10^{8}$ & 178.5\,s          \\
      Nelsen9$^\dagger$         & $(1/200, 1/100)$                                     & $3.0{\times}10^{-11}$ & 30      & $1.9{\times}10^{7}$ & 80.0\,s           \\
      Nelsen12$^\dagger$        & $(2, 5)$                                             & $1.1{\times}10^{-14}$ & 5       & $7.6{\times}10^{7}$ & 4.5\,s            \\
      Nelsen13$^\dagger$        & $(2, 5)$                                             & $2.8{\times}10^{-15}$ & 15      & $1.1{\times}10^{8}$ & 181.2\,s          \\
      Nelsen17$^\dagger$        & $(1, 3)$                                             & $2.9{\times}10^{-15}$ & 15      & $7.3{\times}10^{7}$ & 158.6\,s          \\
      \bottomrule
    \end{tabular}
  \end{table}

  The scaling column reveals the core problem with symbolic
  differentiation: the expression size depends on the generator's
  algebraic complexity in ways that do not track with dimension alone.
  Clayton's power-law generator and AMH's rational generator remain
  tractable at $K{=}30$ with $5.1{\times}10^{5}$ and $2.4{\times}10^{7}$
  leaves respectively, but every other family blows up
  exponentially---Joe reaches $108$~million leaves at $K{=}15$,
  InverseGaussian $164$~million, and Nelsen12 $76$~million at $K{=}5$.
  Nelsen12 times out at $K{=}10$; Nelsen13 and Nelsen17 reach $K{=}15$
  before timing out at the same exponential rate.
  Our polynomial-powering algorithm avoids this family-dependent cost:
  it processes all ten families in $O(d^3)$ operations regardless of
  the generator's algebraic form, because the cost depends only on the
  polynomial degree~$K$, not on the symbolic complexity.

  \section{Log-Space Taylor-Mode Differentiation}
  \label{app:log-jet}

  The log-domain machinery of \Cref{app:numerical-stability} keeps the
  Bell pipeline's polynomial coefficients finite at high $d$, but it
  operates on \emph{outputs} of \texttt{jet\_array.jet} rather than
  inside the jet expansion itself.  For Frank and Joe generators on
  extreme-tail leaves, that boundary is exactly where the failure now
  appears: the forward jet returns finite-but-denormal Taylor
  coefficients, and reverse-mode AD then materialises $1/c$ for those
  denormal $c$, overflowing to $\infty$ and producing $\infty \cdot 0$
  in the next $\textsc{dot\_general}$ step.  We resolve the failure by
  representing every Taylor coefficient as a $(\text{sign}, \log|c|)$
  pair end-to-end; the {\em log-space jet}, opt-in via
  \texttt{compile\_model(\dots, log\_jet=True)}.

  \subsection{Why naive float64 fails for Frank-type generators}

  For an Archimedean generator that decays as
$\psi(t) \approx c\,e^{-t}$ at large $t$ (Frank, Joe), the $k$-th
  Taylor coefficient at the composed argument $t_r$ behaves like
$c\,e^{-t_r}/k!$ times a polynomial in $k$.  Once
$t_r \gtrsim 500$ — which for $d{=}75$ flat copies of $u{=}10^{-3}$
  gives $t_r = -\sum \log u_\ell \approx 518$ — the high-order
  coefficients drop into float64 denormal range
  ($|c_k| < 2.2 \times 10^{-308}$) by $k \gtrsim 60$.

  The forward computation tolerates the denormals.  Backward does not:
  the upstream rules in \texttt{jet\_array} (notably
  \texttt{\_log\_taylor} and the body of \texttt{\_div\_taylor\_rule})
  recursively divide by $c$ in their Faà di Bruno recurrences, and the
  Jacobian-vector product of $1/c$ at $c \sim 10^{-310}$ is
$\sim 10^{310}$, beyond \texttt{float64.max}.  The resulting
$\infty$ then meets a structural zero — for example, the constant
  divisor's higher Taylor terms in $u/\theta$ — and
$\textsc{dot\_general}$ returns NaN.  The minimal reproducer is
$d{=}75$ flat-Frank with all leaves at $u{=}10^{-3}$; raw forward
  log-density is $39.92$, raw gradient is $\mathrm{NaN}$.

  \subsection{Architecture: $(\text{sign}, \log|c|)$ throughout}

  We replace the raw float series in \texttt{jet\_array.jet} with a
  \texttt{LogSeries} pair $(\sigma, \lambda)$ whose entries satisfy
$c = \sigma \cdot e^{\lambda}$ for $\sigma \in \{-1, 0, +1\}$ and
$\lambda \in \mathbb{R}$.  Structural zero encodes as
$(\sigma, \lambda) = (0, -\infty)$, which acts as the additive
  identity~$\mathbf{0}$.
  Multiplication and division become sign products and log-magnitude
  addition or subtraction; addition uses signed logsumexp on the
  non-zero support, with $\mathbf{0} \oplus x = x$ by definition:
  \[
    (\sigma_a, \lambda_a) \oplus (\sigma_b, \lambda_b) \;=\;
    \bigl(\operatorname{sign}(S),\; \lambda_M + \log|S|\bigr),
    \quad
    \lambda_M = \max_{i \in A^*} \lambda_i,
    \quad
    S = \sum_{i \in A^*} \sigma_i\, e^{\lambda_i - \lambda_M},
  \]
  where $A^* = \{i \in \{a,b\} : \sigma_i \neq 0\}$ is the active
  support; for empty $A^*$ the result is $\mathbf{0}$, and a singleton
  reduces to its sole operand by inspection.  Inside each jet
  rule's scan body — division, log, log1p, exp, expm1, integer power,
  and multiplication — the same recurrence runs, but every arithmetic
  operation dispatches to its log-space sibling.  Coefficients with
  magnitude $10^{\pm 300}$ now occupy normal float64 territory in
$\lambda$ ($\lambda \approx \pm 690$), so no JVP rule ever forms
$1/c$ for a denormal $c$.

  \subsection{Crossing the bell boundary without leaving log-space}

  A naive {\em raw on entry, log-space inside, raw on exit} wrapper
  re-creates the failure: \texttt{log\_to\_raw} materialises
$\sigma \cdot e^{\lambda}$ at the boundary, and any genuinely-tiny
$\lambda$ becomes a denormal float that the next downstream
  \texttt{jnp.log(|c|)} call differentiates as $1/c \to \infty$.  We
  extend \texttt{jet\_array.jet} with \texttt{return\_log\_series=True}
  so the output stays as a \texttt{LogSeries} pair, and rewrite
  \texttt{bell.\_root\_assembly} to consume that pair directly: the
  existing signed-logsumexp reduction reads $\lambda$ as $\log|c|$ and
$\sigma$ as the sign, skipping the
$\texttt{jnp.log(jnp.abs}(\cdot))$ step that was the residual NaN
  site.  The bell pipeline's earlier stages already operate in log
  domain via \texttt{\_normalize\_poly} and the carried
  \texttt{total\_log\_scale}, so no further bell changes are needed.

  \subsection{Validation}

  \begin{itemize}
    \item \emph{Equivalence on benign inputs.} At $d \in \{10, 20, 50\}$
          with moderate $u$, \texttt{log\_jet=True} matches the raw path on
          forward log-density and gradient to relative error~$10^{-10}$.
    \item \emph{Stability where raw fails.}  At $d{=}75$ flat-Frank with
          all $u{=}10^{-3}$: raw gradient~$\mathrm{NaN}$, log-jet
          gradient~$29.70$ (finite).  At the nested 11-by-9 sp500-style
          pattern with five extreme leaves per sector: raw
          gradient~$\mathrm{NaN}$, log-jet gradient finite; forward
          log-density agrees to relative error~$10^{-7}$.
    \item \emph{Production fit.}  S\&P~500 d{=}98 Frank — the original
          failure that motivated this appendix — converges in $15$
          L-BFGS-B function evaluations to NLL~$=-25749.51$ with
          \texttt{log\_jet=True}, improving on Clayton ($-25346.78$) at the
          same dimension while remaining behind Gumbel ($-26532.01$).  Before
          the fix, the sp500 result table reported \texttt{converged=False}
          with $44$ evaluations and $\mathrm{NLL} = -25897.14$, a fictitious
          value the bell pipeline's old NaN-mask had silently swallowed (see
          \Cref{tab:sp500-classical}).
    \item \emph{Regression.}  The \texttt{acopula} test suite passes with
          \texttt{log\_jet=False} producing identical bytes; the additional
          \texttt{tests/test\_log\_jet.py} (six tests) and
          \texttt{tests/test\_log\_space.py} (twenty-two tests on the
          \texttt{LogSeries} primitives) pass under JIT.
  \end{itemize}

  \subsection{Cost and configurability}

  The log-space rules add one scalar field to every Taylor coefficient
  and one sign-tracking branch in each scan body, so memory doubles
  and per-iteration arithmetic roughly doubles versus the raw path.
  Compile time grows by $30$--$50\%$ on the d{=}98 Frank fit
  ($30$~seconds cold versus $18$~seconds for the raw d{=}75 case).
  Steady-state per-evaluation runtime grows by a similar factor.  The kwarg defaults to \texttt{False}, so existing callers
  pay nothing; users of high-$d$ Frank or Joe pass
  \texttt{log\_jet=True} once and the rest of the pipeline adapts.

  \section{Polynomial Multiplication Alternatives}
  \label{app:fft-discussion}

  We chose direct $O(K^2)$ convolution over the textbook $O(K \log K)$
  FFT path because FFT loses on both speed and precision in the regime
  copula applications occupy.
  Our implementation multiplies truncated polynomials with
  \texttt{lax.conv\_general\_dilated}, which executes a direct
  convolution.
  \Cref{tab:fft-vs-cauchy} confirms the choice empirically: direct
  convolution beats rfft-based multiplication for every sector size up
  to $K{=}192$ and only loses at $K{=}256$, far above the regime
  where copula applications operate.

  \begin{table}[h]
    \centering\small
    \caption{Wall-clock time per truncated polynomial-powering scan
      ($p^1, \ldots, p^K$) at sector size~$K$, JIT-compiled in JAX on a
      single CPU thread.
      Direct convolution wins by a factor of 2--5 for every $K \le 192$;
      the FFT path's relative error is unbounded past $K{=}64$, with the
      earlier $K{=}16$ spike (${\sim}0.02$) and $K{=}24$ spike
      (${\sim}0.56$) reflecting the alternating-sign cancellation that
      destabilises the convolution before the magnitude blow-up.}
    \label{tab:fft-vs-cauchy}
    \begin{tabular}{@{}r r r r@{}}
      \toprule
      $K$ & direct (ms) & FFT (ms) & direct speedup          \\
      \midrule
      4   & $0.007$     & $0.014$  & $2.01\times$            \\
      16  & $0.020$     & $0.067$  & $3.29\times$            \\
      32  & $0.030$     & $0.152$  & $5.08\times$            \\
      64  & $0.060$     & $0.331$  & $5.47\times$            \\
      128 & $0.366$     & $0.796$  & $2.18\times$            \\
      192 & $1.349$     & $1.783$  & $1.32\times$            \\
      256 & $5.443$     & $2.748$  & $0.50\times$ (FFT wins) \\
      \bottomrule
    \end{tabular}
  \end{table}

  Three observations support this choice.
  First, polynomial degrees in copula applications rarely exceed
$K = 50$, where FFT's constant overhead---complex-arithmetic
  promotion, bit-reversal permutations---wipes out its asymptotic
  savings.
  Second, the short-kernel optimisation in
  \texttt{\_convolve\_short\_kernel} already reduces the Cauchy-product
  step from $O(d_v^2)$ to $O(d_v \cdot d_{\max}(v))$, narrowing the gap
  further.
  Third---decisively for our regime---FFT-based polynomial multiplication
  satisfies only a {\em norm-wise} forward-error
  bound~\citep[Ch.~24]{higham2002accuracy},
$|\mathrm{fl}(a \ast b)_k - (a \ast b)_k| \le c\,\epsilon\log K\,
\|a\|_2\|b\|_2$,
  so every output entry inherits an absolute error proportional to the
    {\em largest} coefficient in the inputs.
  Direct convolution, by contrast, preserves {\em component-wise}
  precision: each output entry's error scales with the coefficients that
  actually sum into it.
  Bell polynomial coefficients within a single $\beta$-vector span many
  orders of magnitude---factorial growth from $\beta_1$ to
$\beta_{d_v}$ gives dynamic range above $10^{60}$ already at $d_v = 50$,
  far exceeding the $\sim\!10^{16}$ resolution of float64---so the small
  leading coefficients that the nested-density formula multiplies against
$\psi_r^{(k)}$ at low~$k$ lose all digits to FFT round-off.
  The repeated power scan compounds this loss at every step, so
  FFT-based powering retains no usable precision well below $K{=}100$
  in float64 unless the implementation first rescales the polynomial
  into a narrow dynamic range---a rescaling that our log-domain
  pipeline already performs for the direct path at zero asymptotic cost.
  Should an application drive $K$ above 200, swapping in an FFT path
  requires changing only two internal helpers, with no public-API
  impact.

  \paragraph{Alternative Bell-polynomial algorithms.}
  Factorisation-based algorithms compute partial Bell polynomials via
  recurrence relations or matrix factorisations, achieving the same
$O(d^3)$ asymptotic complexity with different constants.
  Our polynomial-powering formulation sidesteps these alternatives
  because Taylor-mode AD already produces jet coefficients in exactly
  the truncated-power-series representation that polynomial powering
  consumes, so no format conversion is needed.
  Graph-level AD optimisations such as cross-country
  elimination~\citep{bettencourt2019taylor} can reduce compilation
  overhead but remain orthogonal to the density-level contribution of
  this work.

  \section{Symbolic Inversion and Numeric Fallback}
  \label{app:oryx-fallback}

  Two paths supply the inverse generator $\psi^{-1}$ that the density
  and sampler require: a symbolic derivation via Oryx for closed-form
  generators, and a numeric bisection fallback for everything else.
  Oryx~\citep{radul2020you} derives $\psi^{-1}$ by tracing the
  generator through JAX's intermediate representation and
  pattern-matching each primitive against a table of known algebraic
  inverses---for example, $\exp \leftrightarrow \log$,
$x \mapsto x^a \leftrightarrow x \mapsto x^{1/a}$, and
$x \mapsto x + c \leftrightarrow x \mapsto x - c$.
  The procedure succeeds whenever the generator is a composition of
  individually invertible operations with no data-dependent control flow.
  All ten built-in families satisfy this condition.

  Oryx's pattern matcher cannot invert functions that contain
  data-dependent branching---\texttt{lax.cond} or
  \texttt{lax.switch}---because the branch the tracer follows at trace
  time may differ from the branch the runtime selects at evaluation
  time.
  Functions that call numerical subroutines such as adaptive
  quadrature or iterative solvers also fail, because those routines
  expand into opaque \texttt{while\_loop} primitives.
  The matcher raises an error at trace time in either case, never
  silently returning an incorrect result.

  The framework probes each generator at initialisation by evaluating the
  Oryx-derived inverse on a concrete test value.  When the probe succeeds,
  the symbolic inverse handles all subsequent calls.  When it fails,
  the framework switches transparently to a numeric path that solves
$\psi(s) = t$ for~$s$ via \texttt{optimistix.Bisection} with tolerance
$10^{-12}$.  Because generators are strictly decreasing on $[0,\infty)$,
  the bisection bracket always exists.
  Gradients propagate through the root-finding step via the implicit
  function theorem: \texttt{optimistix.ImplicitAdjoint} differentiates
  through the solution by solving
$\frac{\partial\psi}{\partial s}\big|_{s^{\star}} \, \delta s = \delta t$.
  Our test suite verifies round-trip accuracy---$|\psi(\psi^{-1}(t)) - t|
< 10^{-10}$---and gradient correctness against the analytic derivative.

  \section{Rosenblatt Sampling via Generator Derivatives}
  \label{app:rosenblatt}

  The same Bell-polynomial machinery that delivers the density also
  samples from a nested Archimedean copula for free.
  The Bell-polynomial recursion of \Cref{app:bell} supplies, as a
  byproduct, the high-order generator-composition derivatives that
  the textbook {\em conditional-distribution method}---also called
  Rosenblatt sampling---requires.
  The same sampler draws the synthetic data that validates the
  Frank and Gumbel parameter-recovery results in
  \Cref{app:recovery-extended}, and samples the
  InverseGaussian$^\dagger$, Nelsen9$^\dagger$, Nelsen12$^\dagger$,
  Nelsen13$^\dagger$, and Nelsen17$^\dagger$ families used in the
  retinopathy experiment (\Cref{sec:exp-retino}) with no per-family
  implementation.

  \paragraph{The conditional-distribution method.}
  Sampling from an Archimedean copula
$C(u_1,\ldots,u_d) = \psi\bigl(\psi^{-1}(u_1)+\cdots+\psi^{-1}(u_d)\bigr)$
  proceeds variable by variable.
  Let $t_j = \sum_{k=1}^{j}\psi^{-1}(u_k)$ denote the running sum of
  inverse-generator transforms.
  The conditional distribution of $u_{j}$ given $u_1,\ldots,u_{j-1}$
  has the closed form
  \begin{equation}
    C_{j\mid 1..j-1}(u_j) \;=\;
    \frac{\psi^{(j-1)}(t_j)}{\psi^{(j-1)}(t_{j-1})},
    \label{eq:rosenblatt-cond}
  \end{equation}
  which depends only on the $(j{-}1)$-th derivative of the generator
  evaluated at two arguments.
  The sampler draws $V_j \sim \mathrm{Uniform}(0,1)$ independently for
  each variable, then inverts \Cref{eq:rosenblatt-cond} for $u_j$
  satisfying $C_{j\mid 1..j-1}(u_j) = V_j$ by bisection on $u_j$.
  The nested case extends the same recursion through the tree by
  evaluating the appropriate edge-composition derivative at each
  visited internal node.
  \Cref{alg:rosenblatt} gives the flat-copula sampler.
  The nested case generalises by replacing the single-generator
  derivative $\psi^{(j-1)}(\cdot)$ with the full Bell-polynomial mixed
  partial of \Cref{alg:bell-density}: each conditional CDF
$C_{j\mid 1..j-1}(u_j)$ is a ratio of two partial CDFs that
  differentiate variables $1,\ldots,j{-}1$ and fix variables
$j,\ldots,d$ to specific values, with $u_k=1$ marginalising the
  unsampled tail.
  The censoring-mask machinery that already drives likelihood
  evaluation supplies exactly this primitive, so the nested sampler
  reuses the density code path twice per variable
  (\Cref{alg:rosenblatt-nested}).

  \begin{algorithm}[H]
    \footnotesize
    \caption{Rosenblatt sampler for an Archimedean copula with generator
      $\psi$ in dimension~$d$.}
    \label{alg:rosenblatt}
    \algrenewcommand{\algorithmicindent}{0.8em}%
    \begin{algorithmic}
      \Require Generator $\psi$, dimension $d$, RNG
      \Ensure Sample $u \in (0,1)^d$
      \State Draw $V_1, \ldots, V_d \sim \mathrm{Uniform}(0,1)$ iid
      \State $u_1 \gets V_1$;\quad $t_1 \gets \psi^{-1}(u_1)$
      \For{$j = 2, \ldots, d$}
      \State Define $G_j(u) \;:=\;
        \psi^{(j-1)}\!\bigl(t_{j-1} + \psi^{-1}(u)\bigr) \,/\,
        \psi^{(j-1)}(t_{j-1})$
      \Comment{$\psi^{(j-1)}$ via Taylor jet}
      \State $u_j \gets$ \Call{Bisect}{$G_j(u) = V_j$,\; $u \in (0,1)$}
      \State $t_j \gets t_{j-1} + \psi^{-1}(u_j)$
      \EndFor
      \State \Return $(u_1, \ldots, u_d)$
    \end{algorithmic}
  \end{algorithm}

  \begin{algorithm}[H]
    \footnotesize
    \caption{Rosenblatt sampler for a nested Archimedean copula on
      tree~$T$.  Each inner call invokes the Bell-polynomial mixed
      partial of \Cref{alg:bell-density} with a mask that differentiates
      only the already-sampled variables.  An efficient implementation
      memoizes, at every internal node $v$, the Bell-polynomial state of
      \Cref{alg:bell-density}: the running sum $t_v$, the count $d_v$
      of leaves already sampled in $v$'s subtree, and the list of
      completed children's $\alpha^{(c)}$ vectors; each per-leaf step
      then traces only from the leaf's parent up to the root.}
    \label{alg:rosenblatt-nested}
    \algrenewcommand{\algorithmicindent}{0.8em}%
    \begin{algorithmic}
      \Require Nested copula tree~$T$, leaf order $u_1, \ldots, u_d$, RNG
      \Ensure Sample $u \in (0,1)^d$
      \State Draw $V_1, \ldots, V_d \sim \mathrm{Uniform}(0,1)$ iid
      \For{$j = 1, \ldots, d$}
      \State $\delta_j \gets (\underbrace{1,\ldots,1}_{j-1},\; 0, \ldots, 0)$
      \Comment{differentiate $u_1,\ldots,u_{j-1}$ only}
      \State Define $D_j(u) \gets$
      \Call{BellPartial}{$T,\; (u_1, \ldots, u_{j-1},\, u,\, 1, \ldots, 1),\; \delta_j$}
      \Comment{partial CDF}
      \State $u_j \gets$ \Call{Bisect}{$D_j(u)\,/\,D_j(1) = V_j$,\; $u \in (0,1)$}
      \EndFor
      \State \Return $(u_1, \ldots, u_d)$
    \end{algorithmic}
  \end{algorithm}

  \paragraph{Connection to prior work.}
  \citet{hofert2013preprint} identified the high-order
  generator-composition derivatives in \Cref{eq:rosenblatt-cond} as
  the principal obstacle to applying this method beyond small
  dimensions and developed Marshall--Olkin frailty sampling as the
  practical alternative.
  That alternative requires a known Laplace transform per family,
  which blocks cross-family nesting and rules out any generator
  without a closed-form mixing distribution---for example, the
  Gen-AC generator of~\citet{ng2021generativearchimedean} adopted by
  \citet{liu2024hacsurv}.
  The Bell-polynomial recursion of \Cref{app:bell} computes exactly
  these derivatives in closed form for any user-supplied generator
  with the same complexity profile as the density evaluation, so
  sampling and likelihood share a single code path.

  \section{Validity of Nested Constructions}
  \label{app:validity}

  A nested Archimedean copula defines a valid distribution only when
  every parent--child generator pair satisfies a sufficient nesting condition.
  \citet{mcneil2008sampling} required the composition
$h = \psi_{\mathrm{outer}}^{-1} \circ \psi_{\mathrm{inner}}$ to be a
    {\em Bernstein function} (i.e., its derivative completely monotone):
$(-1)^{k-1} h^{(k)}(t) \geq 0$ for all $k \geq 1$ and $t \geq 0$.
  \citet{mcneil2009multivariate} subsequently characterised $d$-dimensional
  Archimedean copulas through the Williamson $d$-transform, weakening the
  complete-monotonicity requirement on the generator itself to
$d$-monotonicity and setting the foundation for the results below.
  \citet{rezapour2015construction} showed that a weaker condition suffices:
$h$ need only be {\em $d_c$-monotone}, that is,
$(-1)^{m-1} h^{(m)}(t) \geq 0$ for $m = 1, \ldots, d_c$ and all $t \geq 0$,
  where $d_c$ counts uncensored leaves under the child node.
  This finite condition matches the information our algorithm already computes:
  the \textsc{Jet} call at each edge returns exactly $d_c$ Taylor coefficients of~$h$.

  For the classical families, closed-form sufficient conditions
  exist: Clayton and Gumbel both require
$\theta_{\mathrm{outer}} \leq \theta_{\mathrm{inner}}$, Frank requires
  the same ordering, and Joe requires
$1 \leq \theta_{\mathrm{outer}} \leq \theta_{\mathrm{inner}}$.
  These orderings guarantee complete monotonicity and therefore $d_c$-monotonicity.

  Our framework does \emph{not} enforce these constraints automatically.
  For same-family pairs, a delta parameterisation
$\theta_{\mathrm{inner}} = \theta_{\mathrm{outer}} + \operatorname{softplus}(\delta)$
  enforces the ordering by construction; all same-family experiments use
  this approach (\Cref{tab:nesting}).

  \begin{table}[h]
    \centering\small
    \caption{Nesting conditions for generator pairs in our experiments.}
    \label{tab:nesting}
    \begin{tabular}{@{}lll l l@{}}
      \toprule
      Experiment      & Outer                               & Inner                                         & Condition                                                      & Enforcement      \\
      \midrule
      S\&P / MIMIC-IV & Clayton                             & Clayton                                       & $\theta_{\mathrm{outer}} \le \theta_{\mathrm{inner}}$          & delta param.     \\
      S\&P / MIMIC-IV & Frank                               & Frank                                         & $\theta_{\mathrm{outer}} \le \theta_{\mathrm{inner}}$          & delta param.     \\
      S\&P / MIMIC-IV & Gumbel                              & Gumbel                                        & $\theta_{\mathrm{outer}} \le \theta_{\mathrm{inner}}$          & delta param.     \\
      MIMIC-IV        & AMH                                 & Clayton                                       & $\theta_{\mathrm{inner}} \geq 1$~\citep{hofert2010samplingnac} & shifted softplus \\
      Retinopathy     & \multicolumn{2}{l}{all 10 families} & \multicolumn{2}{l}{single-layer: nesting N/A}                                                                                     \\
      \bottomrule
    \end{tabular}
  \end{table}
  Same-family rows enforce $\theta_{\mathrm{outer}} \le \theta_{\mathrm{inner}}$
  via the delta parameterisation.
  The AMH/Clayton row uses
  Hofert~\citeyearpar{hofert2010samplingnac}'s closed-form constraint
$\theta_{\mathrm{Clayton}} \geq 1$, enforced by shifted-softplus
$\theta_{\mathrm{Clayton}} = 1 + \mathrm{softplus}(\delta)$; this is
  the single documented closed-form cross-family Archimedean pair with a
  McNeil-2008 Bernstein-function nesting proof.
  InverseGaussian, Nelsen13, and Nelsen17 appear only in the retinopathy experiment, which uses single-layer bivariate copulas; nesting conditions do not apply.

  \paragraph{Cross-family and custom generators.}
  The closed-form constraints above cover the families used in our
  experiments, but a user nesting two different families---or supplying
  a custom generator---needs a more general route.
  When no closed-form parameter constraints exist, enforcing the nesting
  condition during maximum likelihood estimation becomes a
    {\em semi-infinite program} (SIP).
  The $d_e$-monotone condition must hold at every $t \geq 0$, producing
  infinitely many inequality constraints over the finite-dimensional
  parameter vector $\boldsymbol{\theta} \in \mathbb{R}^p$ stacking all
  per-edge generator parameters.
  Let $\mathcal{E}$ denote the set of parent--child edges in the copula
  tree, $d_e$ the number of uncensored leaves under the child of
  edge~$e$, and
$h_e(t;\,\boldsymbol{\theta}) =
\psi_{\mathrm{outer}}^{-1}\!\bigl(\psi_{\mathrm{inner}}(t)\bigr)$
  the edge composition, with $h_e \equiv h_{vc}$ when $e = (v, c)$.
  The constrained MLE is
  \begin{equation}
    \label{eq:sip}
    \begin{aligned}
      \max_{\boldsymbol{\theta}}\quad
       & \ell(\boldsymbol{\theta};\, \mathbf{X})                     \\
      \text{s.t.}\quad
       & (-1)^{m-1}\, h_e^{(m)}(t;\, \boldsymbol{\theta}) \;\geq\; 0
      \quad \forall\, e \in \mathcal{E},\;\;
      m = 1,\ldots, d_e,\;\;
      t \geq 0.
    \end{aligned}
  \end{equation}

  \paragraph{Grid discretisation.}
  A standard approach to a SIP replaces the continuum of constraints
  with a finite grid $\{t_1, \ldots, t_G\} \subset [0, t_{\max}]$
  of $G$ user-chosen points:
  \begin{equation}
    \label{eq:sip-grid}
    \begin{aligned}
      \max_{\boldsymbol{\theta}}\quad
       & \ell(\boldsymbol{\theta};\, \mathbf{X})                       \\
      \text{s.t.}\quad
       & (-1)^{m-1}\, h_e^{(m)}(t_g;\, \boldsymbol{\theta}) \;\geq\; 0
      \quad \forall\, e \in \mathcal{E},\;\;
      m = 1,\ldots, d_e,\;\;
      g = 1,\ldots, G.
    \end{aligned}
  \end{equation}
  Checking each grid point at one edge requires a single \textsc{Jet}
  call that returns the $d_e$~Taylor coefficients of~$h_e$; the signs
  of those coefficients encode the $d_e$-monotone condition.
  Concretely, the \textsc{Jet} call returns the Taylor coefficient vector
$p_m = h_e^{(m+1)}(t) / (m{+}1)!$ for $m = 0, \ldots, d_e{-}1$.
  Because factorial scaling preserves sign, the $d_e$-monotone condition
$(-1)^{m-1} h_e^{(m)}(t) \geq 0$ for $m = 1,\ldots,d_e$ reduces to
  checking that $(-1)^m p_m \geq 0$ for each $m$---an alternating-sign
  test on the Taylor coefficients.
  Each density evaluation already performs one such call per edge at
$t = t_e$, so $N$~training observations contribute $N$~grid points
  per edge at no extra cost; a denser grid requires
$G - N$ additional \textsc{Jet} calls.

  \paragraph{Computing the penalty in principle.}
  The \textsc{acopula} framework can compute the nesting penalty as a
  byproduct of the density evaluation.
  At each edge~$e$, the \textsc{Jet} call that produces the Taylor
  coefficients $p_0, p_1, \ldots, p_{d_e}$ for the Bell transform also
  provides the sign information for the diagnostic.
  The penalty takes the form
  \begin{equation}
    P(\boldsymbol{\theta};\, \mathbf{x})
    = \sum_{e \in \mathcal{E}}\;
    \sum_{m=0}^{d_e - 1}
    \bigl[\min\!\bigl(0,\; (-1)^m\, p_m \bigr)\bigr]^2,
  \end{equation}
  evaluated at $t = t_e(\mathbf{x})$, the sector sum for the current
  observation~$\mathbf{x}$.
  When $P = 0$, all Taylor coefficients satisfy the alternating-sign
  condition at every evaluation point encountered during training.
  A penalised optimiser would minimise
$-\ell(\boldsymbol{\theta}) + \rho\, P(\boldsymbol{\theta})$ with
  penalty weight $\rho \ge 0$, but all experiments in this paper avoid
  this route: every nested fit uses one of the three valid-by-construction
  parameterisations summarised in \Cref{tab:nesting}---same-family delta,
  AMH/Clayton shifted-softplus, or the Gen-AC/ACNet Bernstein
  composition (\Cref{app:nested-neural})---which makes the penalty
  identically zero by construction.
  We describe the penalty approach here as a conceptual extension for
  users who might want to experiment with cross-family or custom
  generators that lack closed-form constraints; we leave empirical
  validation of the penalised route to future work.

  \paragraph{Limitations.}
  The grid relaxation~\Cref{eq:sip-grid} is necessary but not
  sufficient for the full SIP~\Cref{eq:sip}: violations can hide
  between grid points.
  Certifying the $d_e$-monotone condition on all of $[0,\infty)$ for
  arbitrary generators remains an open problem.

  \subsection{Validity of the nested Gen-AC/ACNet construction}
  \label{app:nested-neural}
  The nested construction used by the Gen-AC and ACNet rows of
  \Cref{sec:exp-mimic-nested-neural} and by the HACSurv port of
  \Cref{app:hacsurv} pairs a shared outer generator
$\psi_{\mathrm{out}}(s)$---a mixture of exponentials
$L_{\mathrm{out}}^{-1} \sum_{l=1}^{L_{\mathrm{out}}} \exp(-M_l s)$ for
  Gen-AC (at $L_{\mathrm{out}}$ MC nodes) and the softmax-convex cascade
  of \Cref{app:acnet} for ACNet---with per-panel inner
$\psi_{\mathrm{in},k}(t) = \psi_{\mathrm{out}}(b_k(t))$
  and inner Bernstein function
$b_k(t) = e^{\mu_k}\, t + e^{\beta_k}\bigl(1 - L_{\mathrm{in}}^{-1}
\sum_{l=1}^{L_{\mathrm{in}}} e^{-N_{k,l} t}\bigr)$,
  where $L_{\mathrm{out}}, L_{\mathrm{in}} \in \mathbb{N}$ are the
  outer and inner mixture sizes,
$M_l, N_{k,l} \ge 0$ are exponential rates,
$k$ indexes the per-panel inner generator,
  and $\mu_k, \beta_k \in \mathbb{R}$ are scalar log-shift parameters.
  The construction is valid for any
$(N_k, \mu_k, \beta_k) \in \mathbb{R}_{\ge 0}^{L_{\mathrm{in}}}
\times \mathbb{R}^2$, and the optimiser need not enforce any
  constraint.
  The argument runs in three steps.
  First, $b_k$ is a Bernstein function: $b_k(0) = 0$, $b_k > 0$, and
$b_k'$ is completely monotone (CM) as a sum of CM components.
  Second, the composition of a CM function with a Bernstein function
  is itself
  CM~\citep{mcneil2008sampling,rezapour2015construction}, so
$\psi_{\mathrm{in},k}$ is a valid generator.
  Third, $\psi_{\mathrm{out}}^{-1} \circ \psi_{\mathrm{in},k} = b_k$ is
  automatically Bernstein, so the parent--child pair satisfies the
  nesting condition.

  \section{Gen-AC MLP-pushforward generator with Fourier features}
  \label{app:mlp-fourier}

  The Gen-AC rows of \Cref{tab:mimic-heldout-table} implement the
  MLP-pushforward generator of~\citet{ng2021generativearchimedean} with
  frailty $V = \exp(\mathrm{MLP}(U))$, $U \sim \mathrm{Uniform}(0,1)$,
  so the Laplace transform is
$\psi(t) = \mathbb{E}_U[\exp(-V t)]$.
  HACSurv~\citep{liu2024hacsurv} uses this construction directly with a
  three-layer MLP of hidden width $10$ and leaky-ReLU activations; the
  row labelled Gen-AC$_{1.1\mathrm{k}}$ reproduces that architecture
  with one outer MLP and a per-panel inner MLP, composed with the
  Bernstein drift of \Cref{app:nested-neural}, for
$7 \cdot 2 + 8 \cdot 141 = 1{,}142$ parameters.
  We evaluate $\mathbb{E}_U$ at $N_z{=}32$ fixed equispaced nodes on
$(0,1)$; equispaced nodes instead of per-step Monte-Carlo resamples
  remove the Jensen bias and shift held-out NLL by under $0.02$ nats.

  The raw-$u$ input of HACSurv's MLP plateaus at held-out NLL
$4.914$---barely better than Clayton's $4.949$---because a small MLP
  of Lipschitz constant $\approx 1$ cannot place neighbouring
$\log M_i$ at widely separated rates: the 32 atoms collapse onto
  a narrow band.
  We lift that prior by replacing the scalar input with Fourier
  features~\citep{tancik2020fourierfeatures} at $K$ frequencies:
  \[
    u \;\longmapsto\;
    \bigl[\cos(2\pi k u),\, \sin(2\pi k u)\bigr]_{k=1}^{K}
    \; \in \; \mathbb{R}^{2K},
  \]
  passed into an otherwise identical three-layer MLP of hidden width
$h$.
  The Fourier basis lets the network output sharply different
$\log M_i$ at neighbouring $u_i$ without requiring a large Lipschitz
  constant, recovering the flexibility that direct trainable
$\log M \in \mathbb{R}^{32}$ would have.
  \Cref{tab:mimic-heldout-table} reports two configurations: a
  parameter-matched variant Gen-AC-Fourier$_{260}$ at $(K{=}2, h{=}3)$,
  and the full model Gen-AC-Fourier$_{3.6\mathrm{k}}$ at
$(K{=}16, h{=}10)$.
  At matched capacity Gen-AC remains less efficient than ACNet in this
  cohort---at $260$ parameters Gen-AC reaches $4.828$ versus ACNet's
$4.310$ at $70$ parameters---and even the full Fourier model trails
  the best ACNet by ${\sim}0.17$ nats once its parameter budget is
  released.

  \section{Deep ACNet Generator}
  \label{app:acnet}

  The ACNet rows of \Cref{tab:mimic-heldout-table} use the depth-$L$
  softmax-convex-combination generator of~\citet{ling2020deep} as the
  outer $\psi_{\mathrm{out}}$ and the Bernstein-composition inner
$\psi_{\mathrm{in},k}(t) = \psi_{\mathrm{out}}(b_k(t))$ from
  \Cref{app:nested-neural}.
  Given widths $(w_0, \ldots, w_{L-1})$ and a query $t \in [0, \infty)$,
  the generator is the terminal state of an $L$-layer cascade of
  positive exponential shifts interleaved with simplex-constrained
  convex combinations, defined recursively as
  \[
    \begin{aligned}
      F^{(0)}_i                               & \;=\; e^{-s_{0,i} t},                                               \\
      F^{(\ell)}_i                            & \;=\; e^{-s_{\ell,i} t} \cdot \bigl(A_{\ell-1} F^{(\ell-1)}\bigr)_i
      \text{ for } \ell = 1, \ldots, L-1,                                                                           \\
      \psi_{\mathrm{out}}^{\mathrm{ACNet}}(t) & \;=\; A_{L-1} F^{(L-1)},
    \end{aligned}
  \]
  where $s_\ell = e^{\tilde s_\ell} \in \mathbb{R}^{w_\ell}$ are
  per-layer positive exponential rates and $\tilde s_\ell$ their
  unconstrained log parameters, $A_\ell \in \mathbb{R}^{w_{\ell+1} \times w_\ell}$
  for $\ell < L-1$ and $A_{L-1} \in \mathbb{R}^{1 \times w_{L-1}}$ are
  the row-wise softmaxes of free matrices
$\tilde A_\ell$ so that every row of $A_\ell$ lies on the simplex.
  The first layer's convex combination acts on the scalar $1$, so the
  recursion absorbs it into the initialisation $F^{(0)}$.
  Each per-layer factor $e^{-s_{\ell,i} t}$ is a completely monotone
  function of $t$, every row of $A_\ell$ is a convex combination, and
  closure of complete monotonicity under convex combinations and under
  products of completely monotone functions (applied at every step of
  the recursion) yields a scalar-valued completely monotone function,
  so $\psi_{\mathrm{out}}^{\mathrm{ACNet}}$ is a valid Archimedean
  generator.
  The parameterisation imposes no constraint, and the optimiser updates
$\tilde s_\ell, \tilde A_\ell$ freely.

  The inner generator follows the same Bernstein-composition recipe as
  \Cref{app:nested-neural},
$\psi_{\mathrm{in},k}(t) = \psi_{\mathrm{out}}^{\mathrm{ACNet}}(b_k(t))$
  with $b_k(t) = e^{\mu_k}\, t + e^{\beta_k}(1 - L_{\mathrm{in}}^{-1} \sum_l e^{-N_{k,l} t})$;
  the same Schoenberg-theorem argument shows
$\psi_{\mathrm{in},k}$ is completely monotone and the
  parent--child composition $\psi_{\mathrm{out}}^{-1} \circ \psi_{\mathrm{in},k} = b_k$
  is Bernstein.
  The compact-parameter count in \Cref{tab:mimic-heldout-table}'s
$|\theta|$ column sums the per-layer unconstrained
$\tilde s_\ell \in \mathbb{R}^{w_\ell}$ and
$\tilde A_\ell \in \mathbb{R}^{w_{\ell+1} \times w_\ell}$ of the
  shared outer, the scalar log-shift pair $(\mu_k, \beta_k)$ per panel,
  and the per-panel inner rates
$N_k \in \mathbb{R}^{L_{\mathrm{in}}}$.

  \section{Detailed R Comparison}
  \label{app:r-comparison}

  Our Bell polynomial log-density agrees with R's
  \texttt{nacLL}~\citep{hofert2012density} to machine precision on every
  configuration where R completes, establishing numerical correctness of
  the framework's forward evaluation.
  We compare the two implementations on 12~nested Clayton--Clayton
  configurations spanning $d = 4$ to $60$ total dimensions with
  identical random input matrices, $\theta_{\mathrm{outer}} = 2$, $\theta_{\mathrm{inner}} = 5$, and
$n = 5$ observations per configuration.
  \Cref{tab:density-correctness} reports the results.

  \begin{table}[h]
    \centering
    \small
    \caption{Density correctness of \textsc{acopula} versus R's
      \texttt{nacLL} on nested Clayton--Clayton copulas.
      Here, $s$ denotes the number of sectors, $K$ the leaves per
      sector, and $d = sK$ the total dimension.
      The relative error is $|\ell_{\mathrm{ours}} - \ell_R| / |\ell_R|$.
      Arbitrary-precision verification attributes the $d{=}60$
      discrepancy to R's \texttt{nacLL}, not our framework.}
    \label{tab:density-correctness}
    \begin{tabular}{@{}r r r r r@{}}
      \toprule
      $s$ & $K$ & $d$ & Rel.\ error          & Status           \\
      \midrule
      2   & 2   & 4   & $< 10^{-14}$         & \checkmark       \\
      2   & 5   & 10  & $< 10^{-14}$         & \checkmark       \\
      2   & 10  & 20  & $< 10^{-14}$         & \checkmark       \\
      2   & 20  & 40  & $< 10^{-14}$         & \checkmark       \\
      2   & 30  & 60  & $3.5 \times 10^{-5}$ & R precision loss \\
      3   & 5   & 15  & $< 10^{-14}$         & \checkmark       \\
      4   & 5   & 20  & $< 10^{-14}$         & \checkmark       \\
      5   & 5   & 25  & $< 10^{-14}$         & \checkmark       \\
      5   & 10  & 50  & $< 10^{-14}$         & \checkmark       \\
      6   & 5   & 30  & $< 10^{-14}$         & \checkmark       \\
      7   & 5   & 35  & $< 10^{-14}$         & \checkmark       \\
      8   & 5   & 40  & $< 10^{-14}$         & \checkmark       \\
      \bottomrule
    \end{tabular}
  \end{table}

  Eleven of 12~configurations agree to relative error below $10^{-14}$,
  matching float64 roundoff.
  We attribute the single $d{=}60$ discrepancy to precision loss in R's
  \texttt{nacLL}, not our framework: Mathematica's symbolic
  differentiation of the copula CDF via \texttt{D[C, u1, \ldots, ud]}
  provides an algorithm-agnostic ground truth that agrees with our
  framework to machine epsilon at every $K$ from~2 to~30, while R's
  \texttt{nacLL} diverges at $K \ge 31$ (see also
  \Cref{app:numerical-stability} for the cross-family Mathematica
  check).
  The scripts \texttt{mathematica\_ground\_truth.wls} and
  \texttt{verify\_against\_mathematica.py} reproduce this comparison.

  \paragraph{Controlled runtime comparison.}
  Under matched single-thread settings, \textsc{acopula} runs orders of
  magnitude faster than R's \texttt{nacLL} at moderate dimension and
  completes families that \texttt{nacLL} cannot evaluate at all.
  \Cref{tab:runtime-comparison} reports wall-clock time for the
  log-likelihood of $n{=}100$ observations under an apples-to-apples
  single-thread configuration that pins thread counts and the BLAS
  backend across both runtimes.
  Both R and Python run with
  \texttt{OMP\_NUM\_THREADS}=\texttt{OPENBLAS\_NUM\_THREADS}=\texttt{MKL\_NUM\_THREADS}{=}1,
  read the same float64 input matrices from disk, and time identical
  nested same-family configurations with $\theta_{\mathrm{outer}}{=}2$ and
$\theta_{\mathrm{inner}}{=}5$ via \texttt{microbenchmark} on the R side and
  \texttt{time.perf\_counter} after a JIT warmup on the Python side.
  We compare three families that R's \texttt{nacLL} purports to support:
  Clayton, Frank, and Gumbel.

  \begin{table}[h]
    \centering
    \small
    \caption{Controlled single-thread runtime of \textsc{acopula},
      JIT-compiled on CPU, versus R \texttt{nacLL} on nested same-family
      copulas with $n{=}100$ observations, $\theta_{\mathrm{outer}}{=}2$, and
      $\theta_{\mathrm{inner}}{=}5$.
      The ``Compile'' column reports the one-time JIT cost; N/A indicates
      that R's \texttt{nacLL} aborts on the configuration.}
    \label{tab:runtime-comparison}
    \begin{tabular}{@{}l r r r r r@{}}
      \toprule
      Family  & $d$ & \textsc{acopula} (ms) & Compile (ms) & R \texttt{nacLL} (ms) & Speedup       \\
      \midrule
      Clayton & 4   & 0.73                  & 3{,}240      & 1.95                  & $2.7\times$    \\
      Clayton & 10  & 1.69                  & 3{,}766      & 6.04                  & $3.6\times$    \\
      Clayton & 15  & 1.58                  & 3{,}766      & 19.61                 & $12.4\times$   \\
      Clayton & 20  & 2.19                  & 4{,}768      & 81.70                 & $37.3\times$   \\
      Clayton & 25  & 2.39                  & 5{,}103      & 448.65                & $187.4\times$  \\
      Clayton & 30  & 2.57                  & 5{,}854      & 3{,}137.66            & $1218.7\times$ \\
      \midrule
      Frank   & 4   & 1.25                  & 3{,}339      & N/A                   & ---            \\
      Frank   & 10  & 1.99                  & 3{,}676      & N/A                   & ---            \\
      Frank   & 15  & 2.01                  & 4{,}364      & N/A                   & ---            \\
      Frank   & 20  & 2.76                  & 5{,}219      & N/A                   & ---            \\
      Frank   & 25  & 3.18                  & 5{,}622      & N/A                   & ---            \\
      Frank   & 30  & 3.47                  & 6{,}199      & N/A                   & ---            \\
      \midrule
      Gumbel  & 4   & 0.94                  & 2{,}033      & 3.71                  & $3.9\times$    \\
      Gumbel  & 10  & 1.80                  & 2{,}370      & 11.86                 & $6.6\times$    \\
      Gumbel  & 15  & 1.55                  & 2{,}954      & 29.12                 & $18.8\times$   \\
      Gumbel  & 20  & 2.25                  & 3{,}480      & 96.97                 & $43.1\times$   \\
      Gumbel  & 25  & 2.75                  & 3{,}916      & 492.29                & $179.1\times$  \\
      Gumbel  & 30  & 4.21                  & 4{,}292      & 3{,}106.20            & $737.4\times$  \\
      \bottomrule
    \end{tabular}
  \end{table}

  Three findings emerge from \Cref{tab:runtime-comparison}.
  First, our framework matches R's log-likelihood values to relative
  error below $1.7 \times 10^{-15}$ in every cell where R completes,
  ruling out a precision artefact in the timing gap.
  Second, the speedup grows from ${\sim}4\times$ at $d{=}4$ to over
$700\times$ at $d{=}30$ for Gumbel: R's \texttt{nacLL} scales
  exponentially in the number of sectors via partition enumeration,
  whereas our polynomial powering scales as $O(d^3)$, so the gap widens
  monotonically with~$d$.
  Third, R's \texttt{nacLL} aborts on every Frank configuration with
  the error \mbox{\texttt{a.coeff:\ Frank's family is currently not
      supported}}; its documented support for nesting across the five
  classical families therefore fails for Frank, AMH, and Joe, because
  \texttt{nacLL} hard-codes the partial-Bell coefficients only for
  Clayton and Gumbel.
  \Cref{tab:comparison-detailed} aggregates the structural differences
  that drive these gaps.

  \begin{table}[h]
    \centering
    \small
    \caption{Comparison of \textsc{acopula} with R's \texttt{copula}/\texttt{nacopula} packages.}
    \label{tab:comparison-detailed}
    \begin{tabular}{@{}lp{4.2cm}p{4.8cm}@{}}
      \toprule
       & \textbf{R copula/nacopula}                             & \textbf{\textsc{acopula} (ours)} \\
      \midrule
      Derivatives
       & Hand-derived closed forms; must implement $\psi^{(k)}$
       & Automatic via jet; only $\psi$ needed                                                     \\[3pt]
      New family
       & $\psi$ + all derivatives in C/R
       & $\psi$ in 3 lines of Python                                                               \\[3pt]
      Param.\ gradients
       & Analytical for built-in families; numerical otherwise
       & Exact via \texttt{jax.grad} (end-to-end)                                                  \\[3pt]
      Compilation
       & Interpreted R + C extensions
       & XLA JIT compilation                                                                       \\[3pt]
      Censoring
       & Custom per-family code
       & Automatic: \texttt{censored=True}                                                         \\[3pt]
      Bell polynomials
       & Per-family Stirling-number tables
       & Polynomial powering of jet output                                                         \\[3pt]
      Nested density
       & 2-level only; Clayton/Gumbel only
       & Arbitrary depth and all families                                                          \\
      \bottomrule
    \end{tabular}
  \end{table}

  Two practical caveats sharpen the comparison.
  R's closed-form evaluations are faster per call than our jet-based
$k$-step scan, but the Stirling and Eulerian numbers they rely on
  overflow 64-bit floats at moderate orders.
  Our JIT-compiled pipeline pays a one-time compilation cost of
${\sim}50$\,ms and thereafter evaluates in $< 0.25$\,ms per
  observation.
  The \texttt{Copula} base class also exposes an optional
  \texttt{log\_generator\_kth\_derivative(t, k)} hook for users who
  have a closed-form $\psi^{(k)}$, recovering R's per-call speed
  while retaining full framework compatibility.

  \paragraph{Polynomial scaling in~$s$.}
  Our Cauchy-product algorithm replaces R's exponential partition
  enumeration with polynomial multiplication, so cost in the number of
  sectors~$s$ grows polynomially rather than exponentially.
  To quantify the difference, we time the JIT-compiled log-likelihood for
  nested Clayton copulas while varying $s$ from~2 to~50 at sector size
$K = 2$, for which $d = 2s$.
  Execution time grows from 0.4\,ms at $s{=}2$ to 72\,ms at $s{=}50$,
  matching approximately $O(s^{1.7})$ scaling.
  Repeating with $K = 5$ up to $s{=}30$, with $d = 5s$, yields a similar
  polynomial exponent.
  By contrast, R's partition count exceeds $10^{15}$ at $s = 50$,
  rendering exact enumeration infeasible.
  Beyond raw scaling, the choice between exact and Monte Carlo
$\psi^{(k)}$ evaluation involves further design tradeoffs that we
  discuss next.

  \paragraph{Extended Gen-AC comparison.}
  Our Taylor-mode computation of $\psi^{(k)}$ and the Monte Carlo
  estimation in Gen-AC~\citep{ng2021generativearchimedean} occupy complementary points in
  the exactness--scalability tradeoff.
  Taylor-mode AD computes each $\psi^{(k)}$ exactly in $O(k)$ time per
  coefficient, yielding machine-precision likelihoods that support
  gradient-based parameter estimation, nested copula structures, and
  per-variable censoring without approximation error.
  Gen-AC, in contrast, sidesteps generator differentiation by
  representing $\psi^{(k)}$ as a Monte Carlo average over learned latent
  variables; this representation scales to very high $k$ with bounded
  per-sample cost but introduces sampling variance and restricts the
  generator to the learned Laplace class.
  For the moderate dimensions typical of actuarial and survival-analysis
  applications, $d \leq 100$, the exact approach dominates: it produces
  unbiased gradients and handles arbitrary user-defined generators.
  Gen-AC suits regimes where only samples---not likelihoods---matter, or
  where $d$ is so large that even $O(k)$ exact derivative cost grows
  prohibitive.

  \paragraph{Bivariate censored comparison with CopulaCenR.}
  \textsc{acopula} reproduces CopulaCenR's parameter estimates on
  right-censored bivariate data and succeeds where CopulaCenR's
  finite-difference optimiser fails.
  We compare against CopulaCenR~\citep{sun2020copulacenr}, an R package
  that fits bivariate copulas to right-censored data via joint MLE with
  parametric marginals.
  On the retinopathy dataset, with $197$~patients at $60.7\%$ censoring,
  CopulaCenR's Clayton fit yields $\hat\eta = 0.895$ versus our
  two-stage Inference Functions for Margins
  (IFM)~\citep{joe1996estimation,genest1995semiparametric} estimate $\hat\theta = 0.897$,
  a $0.3\%$ difference; its Gumbel fit yields $\hat\eta = 1.256$ versus
  our $\hat\theta = 1.249$, a $0.6\%$ difference.
  The CopulaCenR estimate $\hat\eta$ uses the same Clayton/Gumbel
  parameterisation as our $\hat\theta$, so the two are directly
  comparable.
  CopulaCenR's Frank optimiser fails on this dataset with a
  finite-difference overflow, while our AD-based gradient computation
  converges to $\hat\theta = 2.255$.
  The close parameter agreement on two families confirms that the
  copula likelihoods agree in the bivariate censored setting, despite
  two methodological differences: joint MLE versus IFM, and shared
  versus per-margin Weibull marginals.

  \section{AD Gradient Verification}
  \label{app:gradient-check}

  JAX's reverse-mode AD produces parameter gradients that match central
  finite differences to relative error below $10^{-8}$ across families,
  dimensions, and censoring, validating the gradient path that drives
  parameter estimation.
  We compare the AD gradient $g_{\mathrm{AD}} = \nabla_\theta \ell$
  (where $\ell$ is the total log-likelihood across the configuration's
  training sample) that \texttt{jax.grad} returns against the central
  finite-difference approximation
$g_{\mathrm{FD},i}
= \bigl[\ell(\theta + \varepsilon e_i) - \ell(\theta - \varepsilon e_i)\bigr]
/ (2\varepsilon)$,
  with $\varepsilon = 10^{-5}$ and $e_i \in \mathbb{R}^p$ the $i$-th
  standard basis vector for parameter dimension $p$ (the \texttt{Params}
  column in \Cref{tab:gradient-check}), across four nested copula
  configurations that exercise different families, dimensions, and
  censoring.
  \Cref{tab:gradient-check} reports the maximum relative error
$\max_i |g_{\mathrm{AD},i} - g_{\mathrm{FD},i}|
/ \max(|g_{\mathrm{AD},i}|, 10^{-10})$.

  \begin{table}[h]
    \centering\small
    \caption{AD gradient vs.\ central finite differences
    ($\varepsilon{=}10^{-5}$).
    All configurations use 2-level nesting with the stated sector
    structure $(M \times K)$ denoting $M$ sectors of $K$ leaves each.}
    \label{tab:gradient-check}
    \begin{tabular}{@{}l c c r@{}}
      \toprule
      Configuration                      & $d$ & Params & Max rel.\ error       \\
      \midrule
      Clayton (2 $\times$ 5)             & 10  & 2      & $5.7 \times 10^{-9}$  \\
      Frank (2 $\times$ 5)               & 10  & 2      & $2.0 \times 10^{-9}$  \\
      Gumbel (2 $\times$ 5)              & 10  & 2      & $8.9 \times 10^{-11}$ \\
      Clayton (4 $\times$ 5), 55\% cens. & 20  & 2      & $5.9 \times 10^{-10}$ \\
      \bottomrule
    \end{tabular}
  \end{table}

  All four configurations achieve relative error below $10^{-8}$,
  confirming that JAX's reverse-mode AD produces correct gradients
  through the full Bell-polynomial pipeline---including the log-domain
  rescaling, signed logsumexp, and polynomial powering steps that
  \Cref{app:numerical-stability} describes.
  The censored configuration, with 55\% of variables censored, shows
  no degradation relative to the fully observed cases, which validates
  the gradient path through the dynamic censoring mechanism.

  \section{High-Dimensional Parameter Recovery}
  \label{app:highdim-recovery}

  The MLE recovers the true parameters with bias under $2.7\%$ and
  sampling variability that the observed-Fisher SE calibrates to within
$2.9\%$ on average, across $N \in \{200, 500, 1000\}$ and 500
  replications per cell.
  \Cref{tab:recovery-highdim} in the main text reports replicated
  recovery at $N{=}500$; \Cref{tab:recovery-reps-grid} extends that
  result to the full $N$ grid and reports bias, empirical SE,
  observed-Fisher SE, RMSE, and the L-BFGS-B convergence rate per cell.
  The empirical SE shrinks at the expected $1/\sqrt{N}$ rate as $N$
  grows: the EmpSE ratio between $N{=}200$ and $N{=}1000$ ranges from
$2.03$ to $2.33$ across the seven parameters, all close to the
  theoretical $\sqrt{5} \approx 2.24$.
  The observed-Fisher SE matches the empirical SE within $2.9\%$ on
  average and within $8.3\%$ at worst, confirming that the Hessian-based
  asymptotic approximation calibrates the actual sampling variability
  of the MLE on the Bell-polynomial likelihood.
  The bias remains uniformly small, under $2.7\%$, and stable across
$N$, indicating that the residual bias arises from finite-sample MLE
  bias rather than from convergence failures.

  \begin{table}[htbp]
    \centering\small
    \caption{Replicated recovery on nested Clayton copulas with
    $R{=}500$ Monte Carlo replications per cell.
    Each replication draws a fresh sample of size $N$ and refits the MLE.
    EmpSE is the across-replication standard deviation of $\hat\theta$;
    RMSE is its root-mean-squared error against the true value;
    Bias is the mean of $\hat\theta - \theta_{\mathrm{true}}$.
    HessSE is $\sqrt{\mathrm{diag}((N\hat H)^{-1})}$ averaged across
    replications, where $\hat H = -N^{-1}\nabla^2\ell(\hat\theta)$
    is the per-observation observed information; every replication
    produces a positive-definite Hessian.
    Conv\% is the L-BFGS-B convergence rate.}
    \label{tab:recovery-reps-grid}
    \setlength{\tabcolsep}{4pt}
    \begin{tabular}{@{}l r l r r r r r r@{}}
      \toprule
      Config & $N$       & Param                      & True & Mean  & EmpSE & HessSE & RMSE  & Conv\% \\
      \midrule
      2-level
             & 200       & $\theta_{\mathrm{outer}}$  & 1.50 & 1.517 & 0.112 & 0.111  & 0.113 & 100    \\
      $d{=}10$
             &           & $\theta_{\mathrm{inner}}$  & 3.00 & 3.023 & 0.084 & 0.084  & 0.087 & 100    \\
             & 500       & $\theta_{\mathrm{outer}}$  & 1.50 & 1.521 & 0.069 & 0.070  & 0.072 & 100    \\
             &           & $\theta_{\mathrm{inner}}$  & 3.00 & 3.023 & 0.054 & 0.054  & 0.059 & 100    \\
             & $1{,}000$ & $\theta_{\mathrm{outer}}$  & 1.50 & 1.519 & 0.049 & 0.050  & 0.053 & 100    \\
             &           & $\theta_{\mathrm{inner}}$  & 3.00 & 3.022 & 0.039 & 0.038  & 0.044 & 100    \\
      \midrule
      2-level
             & 200       & $\theta_{\mathrm{outer}}$  & 2.00 & 2.019 & 0.071 & 0.074  & 0.074 & 100    \\
      $d{=}20$
             &           & $\theta_{\mathrm{inner}}$  & 4.00 & 4.026 & 0.077 & 0.074  & 0.081 & 100    \\
             & 500       & $\theta_{\mathrm{outer}}$  & 2.00 & 2.019 & 0.048 & 0.047  & 0.052 & 100    \\
             &           & $\theta_{\mathrm{inner}}$  & 4.00 & 4.023 & 0.048 & 0.047  & 0.054 & 100    \\
             & $1{,}000$ & $\theta_{\mathrm{outer}}$  & 2.00 & 2.019 & 0.035 & 0.033  & 0.040 & 100    \\
             &           & $\theta_{\mathrm{inner}}$  & 4.00 & 4.027 & 0.033 & 0.033  & 0.043 & 100    \\
      \midrule
      3-level
             & 200       & $\theta_{\mathrm{outer}}$  & 1.00 & 1.019 & 0.079 & 0.079  & 0.081 & 99.8   \\
      $d{=}50$
             &           & $\theta_{\mathrm{middle}}$ & 2.00 & 2.019 & 0.044 & 0.045  & 0.048 & 99.8   \\
             &           & $\theta_{\mathrm{inner}}$  & 4.00 & 4.024 & 0.050 & 0.047  & 0.056 & 99.8   \\
             & 500       & $\theta_{\mathrm{outer}}$  & 1.00 & 1.016 & 0.052 & 0.050  & 0.054 & 99.4   \\
             &           & $\theta_{\mathrm{middle}}$ & 2.00 & 2.016 & 0.029 & 0.028  & 0.033 & 99.4   \\
             &           & $\theta_{\mathrm{inner}}$  & 4.00 & 4.023 & 0.032 & 0.030  & 0.039 & 99.4   \\
             & $1{,}000$ & $\theta_{\mathrm{outer}}$  & 1.00 & 1.018 & 0.035 & 0.035  & 0.039 & 99.4   \\
             &           & $\theta_{\mathrm{middle}}$ & 2.00 & 2.019 & 0.021 & 0.020  & 0.028 & 99.4   \\
             &           & $\theta_{\mathrm{inner}}$  & 4.00 & 4.023 & 0.022 & 0.021  & 0.032 & 99.4   \\
      \bottomrule
    \end{tabular}
  \end{table}

  \subsection{Frank and Gumbel Recovery}
  \label{app:recovery-extended}

  The recovery result extends to Frank and Gumbel generators, confirming
  that the Bell-polynomial likelihood and its AD gradients remain
  correct regardless of the generator choice.
  \Cref{tab:recovery-highdim} uses Clayton exclusively because Clayton
  admits closed-form Marshall--Olkin sampling; \Cref{tab:recovery-extended}
  repeats the experiment with Frank and Gumbel generators and samples
  via the Rosenblatt transform.
  L-BFGS-B recovers all parameters to within 5\% at $d{=}10$ and within
  2.1\% at $d{=}20$ for both families.
  The Frank $d{=}10$ root parameter shows 13\% error because only two
  sector-level aggregates inform the root; at $d{=}20$ with four sectors
  this error drops below 1\%.

  \begin{table}[htbp]
    \centering\small
    \caption{Parameter recovery on nested Frank and Gumbel copulas
      ($N{=}500$, 2-level nesting, L-BFGS-B optimiser).
      The experiment samples data via the Rosenblatt transform,
      then recovers parameters using exact AD gradients.}
    \label{tab:recovery-extended}
    \setlength{\tabcolsep}{4pt}
    \begin{tabular}{@{}l l r l r r r@{}}
      \toprule
      Family & $d$ & Sectors & Param                     & True & MLE  & Err\% \\
      \midrule
      Frank  & 10  & 2       & $\theta_{\mathrm{outer}}$ & 2.00 & 1.74 & 13.0  \\
             &     &         & $\theta_{\mathrm{inner}}$ & 5.00 & 4.95 & 1.1   \\
      \midrule
      Gumbel & 10  & 2       & $\theta_{\mathrm{outer}}$ & 2.00 & 1.90 & 4.9   \\
             &     &         & $\theta_{\mathrm{inner}}$ & 5.00 & 5.02 & 0.3   \\
      \midrule
      Frank  & 20  & 4       & $\theta_{\mathrm{outer}}$ & 2.00 & 1.98 & 0.9   \\
             &     &         & $\theta_{\mathrm{inner}}$ & 4.00 & 3.94 & 1.5   \\
      \midrule
      Gumbel & 20  & 4       & $\theta_{\mathrm{outer}}$ & 2.00 & 2.04 & 2.1   \\
             &     &         & $\theta_{\mathrm{inner}}$ & 4.00 & 4.04 & 1.0   \\
      \bottomrule
    \end{tabular}
  \end{table}

  \section{Censoring Validation}
  \label{app:censoring}

  The per-variable censoring mechanism reproduces the closed-form
  mixed partial derivatives of the bivariate and trivariate Clayton and
  Gumbel CDFs to machine precision across every censoring pattern.
  For the Clayton family with generator $\psi(t) = (1+t)^{-1/\theta}$,
  the $k$-th order mixed partial of the $d$-variate CDF admits the
  closed form
  \[
    \frac{\partial^k C}{\partial u_1 \cdots \partial u_k}
    = \prod_{j=0}^{k-1}(1 + j\theta)\;\prod_{i=1}^k u_i^{-(\theta+1)}
    \;\Bigl(\textstyle\sum_{i=1}^d u_i^{-\theta} - (d-1)\Bigr)^{-(1/\theta+k)}.
  \]
  We compare the output of our Bell-polynomial censoring implementation,
  which sets $\alpha^{(0)} = [1]$ for censored leaves, against this
  closed form for $d \in \{2,3\}$ with every possible censoring pattern
  running over $k=0,\ldots,d$ observed variables, 5--6 evaluation
  points per pattern, and $\theta \in \{0.5, 1, 2, 5\}$.
  We perform analogous checks for the Gumbel family with
$\psi(t)=e^{-t^{1/\theta}}$ and $\theta\in\{1.5,2,3,5\}$, where nested
  \texttt{jax.grad} applied to the known closed-form CDF yields the
  reference values.

  \begin{table}[htbp]
    \centering
    \caption{Censoring validation: framework output versus closed-form
      expressions.
      The test sweep covers every censoring pattern from $k=0$ to $k=d$
      observed variables.}
    \label{tab:censoring-validation}
    \begin{tabular}{llccc}
      \toprule
      Family  & Dimension & Tests & Pass rate & Max $|\Delta\log f|$ \\
      \midrule
      Clayton & $d=2$     & 48    & 48/48     & $3.6\times 10^{-15}$ \\
      Clayton & $d=3$     & 80    & 80/80     & $7.1\times 10^{-15}$ \\
      Gumbel  & $d=2$     & 48    & 48/48     & $2.7\times 10^{-15}$ \\
      Gumbel  & $d=3$     & 80    & 80/80     & $2.8\times 10^{-15}$ \\
      \bottomrule
    \end{tabular}
  \end{table}

  All 256 tests pass with a maximum absolute error of $7.1\times 10^{-15}$
  in log-likelihood, confirming that the per-variable censoring
  mechanism correctly reduces to the appropriate mixed partial derivative
  of the copula CDF for every censoring configuration in the sweep.

  \section{Supported Archimedean Families}
  \label{app:families}

  \textsc{acopula} ships with ten Archimedean families---the five
  classical generators plus five additional generators drawn from
  Nelsen's catalogue~\citep{nelsen2006introduction} that existing
  software cannot evaluate---and admits arbitrary user-defined
  generators through a single decorator.
  \Cref{tab:families} lists the included generators; the classical
  families above the rule are standard in the literature, and the five
  additional families below appear in our experiments.
  Users can add further families by supplying only the generator
  function via the \texttt{@copula} decorator, after which \textsc{acopula}
  derives all derivatives, inverses, and sampling procedures
  automatically.

  \begin{table}[h]
    \centering
    \caption{One-parameter Archimedean copula families included in
      \textsc{acopula}.  Classical families appear above the rule;
      additional families appear below.  Sampling uses numerical Laplace
      inversion for all families.}
    \label{tab:families}
    \begin{tabular}{@{}llll@{}}
      \toprule
      \textbf{Family} & \textbf{Parameter}                               & \textbf{Generator $\psi_\theta(t)$}
                      & \textbf{Mixing dist.\ $F_{\psi}$}                                                      \\
      \midrule
      Clayton
                      & $\theta > 0$
                      & $(1+t)^{-1/\theta}$
                      & $\operatorname{Gamma}(1/\theta, 1)$                                                    \\
      Frank
                      & $\theta > 0$
                      & $-\frac{1}{\theta}\log[1-(1-e^{-\theta})e^{-t}]$
                      & Log-series$(1-e^{-\theta})$                                                            \\
      Gumbel
                      & $\theta \geq 1$
                      & $e^{-t^{1/\theta}}$
                      & $S(1/\theta, 1)$ stable                                                                \\
      Joe
                      & $\theta \geq 1$
                      & $1-(1-e^{-t})^{1/\theta}$
                      & Sibuya$(1/\theta)$                                                                     \\
      Ali--Mikhail--Haq
                      & $0 \leq \theta < 1$
                      & $\frac{1-\theta}{e^t - \theta}$
                      & Geometric$(1-\theta)$                                                                  \\
      \midrule
      InverseGaussian
                      & $\theta > 0$
                      & $e^{(1-\sqrt{1+2\theta t})\,/\,\theta}$
                      & Inv-Gauss$(1,\, 1/\theta)$                                                             \\
      Nelsen9
                      & $\theta > 0$
                      & $e^{(1-e^t)\,/\,\theta}$
                      & Poisson$(1/\theta)$                                                                    \\
      Nelsen12 (Burr)
                      & $\theta \geq 1$
                      & $(1+t^{1/\theta})^{-1}$
                      & Mittag-Leffler$(1/\theta)$                                                             \\
      Nelsen13
                      & $\theta \geq 1$
                      & $e^{1-(1+t)^{1/\theta}}$
                      & ---                                                                                    \\
      Nelsen17
                      & $\theta > 0$
                      & $\bigl[1+(2^{-\theta}{-}1)e^{-t}\bigr]^{-1/\theta} - 1$
                      & ---                                                                                    \\
      \bottomrule
    \end{tabular}
  \end{table}

  \section{High-Dimensional Nested Scaling}
  \label{app:highdim}

  This appendix establishes that the $O(d^2)$ complexity bound of
  \Cref{sec:method-complexity} holds in practice across three orders of
  magnitude in $d$, and that the prefactor stays small enough for
  single-observation densities to evaluate in milliseconds at $d{=}1{,}000$.
  We measure single-observation log-likelihood wall-clock for nested
  Clayton copulas at $d \in [10, 8{,}000]$ on a single AMD
  Threadripper~3960X core under float64.
  We scan three topologies: {\em fixed $K{=}5$} two-level (root
  over $M{=}d/5$ Clayton sectors of five leaves) and three-level
  (mid layer of five sectors per group); and {\em balanced $\sqrt{d}$}
  ($M{=}K{=}\lceil\sqrt{d}\,\rceil$).
  All times are medians over five post-warmup calls of a single
  JIT-compiled XLA program; \Cref{fig:scaling} plots the curves and
  \Cref{tab:highdim-app} lists selected dimensions.

  \paragraph{Two regimes.}
  The empirical scaling exhibits two regimes separated by $d{\approx}1{,}000$:
  a sub-quadratic overhead-dominated regime at the dimensions our
  experiments use, and the predicted $O(d^2)$ asymptote past
$d{\approx}4{,}000$ (\Cref{fig:scaling}).
  In the {\em real-data regime} that covers our experiments---retinopathy
  at $d{=}2$, MIMIC-IV at $d{=}53$, and S\&P~500 at $d{=}98$---the
  empirical exponent is sub-linear, $\alpha \approx 0.55$ on
$d \in [10, 100]$.
  The Bell-polynomial work scales as $c \cdot d^2$ with
$c \approx 10^{-5}$\,ms, so per-call XLA dispatch and the per-sector
  \texttt{lax.scan} that walks the tree dominate the small quadratic
  constant.
  Across the measured regime $d \in [100, 8{,}000]$ a global OLS
  log-log fit gives $\alpha \approx 2.03$ at 2-level and
$\alpha \approx 1.92$ at 3-level, both consistent with the
  theoretical $O(d^2)$ bound of \Cref{sec:method-complexity}.
  The local slopes in \Cref{tab:highdim-app} show a more nuanced
  picture: the 2-level slope reaches $\alpha_\Delta{=}2.06$ at
$d{=}8{,}000$, settled at the asymptote, while the 3-level slope
  at $d{=}8{,}000$ is $\alpha_\Delta{=}2.89$, super-quadratic.  This
  last value reflects L2 cache pressure as the per-sector polynomial
  powering exceeds cache capacity, not an algorithmic shift; the
  pure-Cauchy microbenchmark of \Cref{app:complexity} isolates the
  cache transition from the algorithmic complexity.
  The balanced $\sqrt{d}$ topology hits the cache wall earlier because
  each sector contains $K{=}\lceil\sqrt{d}\,\rceil$ leaves---about $80$
  at $d{=}6{,}400$---inflating its $d \geq 1{,}000$ slope to
$\alpha \approx 3.24$.

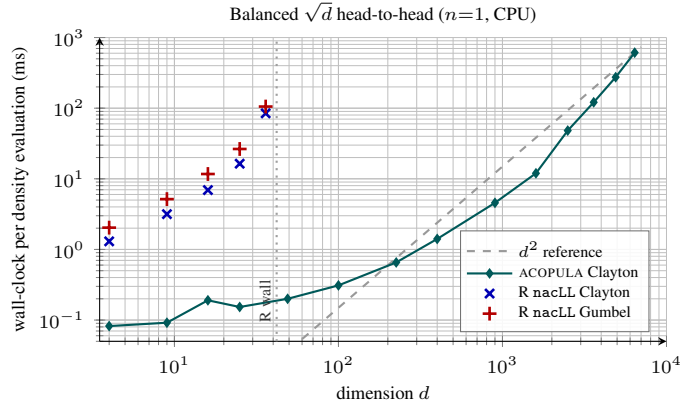
\begin{figure}[h]
  \centering
  \begin{tikzpicture}
    \begin{axis}[
        width=0.65\textwidth,
        height=5.6cm,
        xlabel={dimension $d$},
        ylabel={wall-clock per density evaluation (ms)},
        xmode=log, ymode=log,
        xmin=3.5, xmax=10000,
        ymin=5e-2, ymax=1e3,
        log basis x={10}, log basis y={10},
        title={\scriptsize Balanced $\sqrt{d}$ head-to-head ($n{=}1$, CPU)},
        title style={yshift=-4pt, font=\scriptsize},
        xlabel style={font=\scriptsize, yshift=2pt},
        ylabel style={font=\scriptsize, yshift=-3pt},
        xticklabel style={font=\scriptsize},
        yticklabel style={font=\scriptsize},
        legend style={
          font=\tiny,
          at={(0.97,0.03)}, anchor=south east,
          draw=black!55, fill=white, fill opacity=0.9, text opacity=1,
          row sep=-2pt, inner sep=2pt,
        },
        legend cell align={left},
        grid=both,
        grid style={very thin, color=black!25},
        axis lines=left,
        clip=true,
      ]
      \addplot[gray!80, dashed, thick, domain=4:7000, samples=2,
        forget plot]
        {611.3822 / 6400^2 * x^2};
      \addlegendimage{gray!80, dashed, thick}
      \addlegendentry{$d^2$ reference}
      \addplot[gray!75, dotted, thick, forget plot]
        coordinates {(42,5e-2) (42,1e3)};
      \addplot[teal!70!black, mark=diamond*, mark size=1.4pt, thick]
        coordinates {
          (4,0.0822) (9,0.0917) (16,0.1905) (25,0.1531) (49,0.2000)
          (100,0.3097) (225,0.6545) (400,1.4077) (900,4.5681)
          (1600,11.9699) (2500,48.1744) (3600,121.1474) (4900,274.7193)
          (6400,611.3822)
        };
      \addlegendentry{\textsc{acopula} Clayton}
      \addplot[only marks, blue!70!black, mark=x, mark size=2.4pt,
        line width=1pt]
        coordinates {
          (4,1.3008) (9,3.1608) (16,6.9263) (25,16.3785) (36,84.4727)
        };
      \addlegendentry{R \texttt{nacLL} Clayton}
      \addplot[only marks, red!70!black, mark=+, mark size=2.6pt,
        line width=1pt]
        coordinates {
          (4,2.0347) (9,5.1579) (16,11.6934) (25,26.4382) (36,105.4474)
        };
      \addlegendentry{R \texttt{nacLL} Gumbel}
      \node[font=\tiny, rotate=90, color=black!80, anchor=south west]
        at (axis cs:44, 7e-2) {R wall};
    \end{axis}
  \end{tikzpicture}
  \caption{Head-to-head against R \texttt{nacLL} under the balanced
    $M{=}K{=}\lceil\sqrt{d}\,\rceil$ topology, $n{=}1$ single-evaluation
    timings, single-thread CPU, float64.  R aborts at $d{\geq}40$ for
    Clayton/Gumbel and on Frank for every $d$.  \textsc{acopula}
    extends to $d{=}6{,}400$ with the same $O(d^2)$ asymptote as the
    fixed-$K{=}5$ topology (\Cref{fig:scaling}); the slope steepening
    in $d \in [1{,}600,\, 6{,}400]$ is a hardware cache transition as
    each sector size $K{=}\lceil\sqrt{d}\,\rceil$ approaches $80$ leaves
    (\Cref{app:complexity}).}
  \label{fig:scaling-full}
\end{figure}

  \paragraph{Head-to-head against R \texttt{nacLL}: balanced
    $\sqrt{d}$ topology.}
  The R wall at $d{\approx}40$ that \Cref{fig:scaling} reports under the
  fixed-$K{=}5$ topology persists under the harder balanced
$M{=}K{=}\lceil\sqrt{d}\,\rceil$ topology, where the per-sector arity
  itself grows with $d$ (\Cref{fig:scaling-full}).
  The dominant per-density cost in R \texttt{nacLL} is the partition
  enumeration in \texttt{b.coeff} from the \texttt{copula} source file
  \texttt{R/Rsource/dnac.R}, lines 95--145: for the Hofert--Pham
  formula, the inner loop is $O(n \cdot d_0 \cdot \prod_v K_v)$, where
$d_0$ is the root arity and $K_v$ are the per-sector sizes; with
  balanced sectors of size $K$ and $M{=}d/K$ sectors this becomes
$O(K^M)$ per density, exponential in the number of sectors.
  The classical $O(d^2)$ Stirling-table closed form of
  \citet{hofert2012estimators} applies only to flat single-level
  Archimedean copulas; for nested copulas no Stirling shortcut exists
  once $M \geq 2$.
  Timing R at $n{=}1$ requires a two-line patch to \texttt{nacLL} that
  adds \texttt{drop=FALSE} at \texttt{dnac.R}~line~189; upstream
  \texttt{nacLL} only handles $n{\geq}2$.

\begin{table}[h]
    \centering\small
    \caption{Single-observation wall-clock at selected dimensions on
      AMD Threadripper~3960X (single thread, float64).  $\alpha_\Delta$
      is the local log-log slope from the previous row, illustrating the
      transition from overhead-dominated ($\alpha_\Delta \ll 2$) at small
      $d$ to the $O(d^2)$ asymptote ($\alpha_\Delta \approx 2$) at large
      $d$; intermediate slopes also reflect L1/L2 cache transitions, not
      algorithmic shifts.}
    \label{tab:highdim-app}
    \begin{tabular}{@{}r r r r r r r@{}}
      \toprule
              & \multicolumn{3}{c}{2-level (fixed $K{=}5$)}
              & \multicolumn{3}{c}{3-level (fixed $K{=}5$)}                                                             \\
      \cmidrule(lr){2-4}\cmidrule(lr){5-7}
      $d$     & JIT (s)                                     & Exec (ms) & $\alpha_\Delta$
              & JIT (s)                                     & Exec (ms) & $\alpha_\Delta$                               \\
      \midrule
      10      & $0.78$                                      & $0.10$    & ---             & $1.31$ & $0.12$    & ---    \\
      100     & $0.78$                                      & $0.28$    & $0.45$          & $1.12$ & $0.35$    & $0.47$ \\
      1{,}000 & $1.59$                                      & $16.16$   & $1.76$          & $1.68$ & $7.38$    & $1.32$ \\
      4{,}000 & $3.47$                                      & $150.55$  & $1.61$          & $2.61$ & $195.81$  & $2.36$ \\
      8{,}000 & $6.40$                                      & $629.91$  & $2.06$          & $5.53$ & $1454.98$ & $2.89$ \\
      \bottomrule
    \end{tabular}
  \end{table}

  \section{Memory and Compilation Benchmarks}
  \label{app:memory}

  The compilation cost a user pays once at fit time grows linearly in
$d$, the gradient program stays at the standard reverse-mode
${\sim}3\times$ multiple of the forward program, and the entire fused
  HLO fits in a few thousand instructions---small enough that the JIT
  cache amortises across the inner-loop evaluations of any optimiser.
  \Cref{tab:memory-bench} reports compilation overhead, execution time,
  and XLA program size for the Bell polynomial likelihood and its
  gradient at increasing dimensions $d$.
  All measurements use a two-level nested Clayton copula at sector size
$K{=}5$, on an AMD Threadripper~3960X CPU (3.8\,GHz, 24 cores / 48
  threads) with JAX~0.8.3 in float64 mode.
  We count HLO instructions in the single fused program XLA emits via
  \texttt{jax.jit(f).lower(x).compile().as\_text()}.

  \begin{table}[h]
    \centering
    \caption{Compilation and execution benchmarks for the Bell polynomial
      likelihood and gradient on CPU (AMD Threadripper~3960X, $n{=}1$
      observation).
      ``JIT'' denotes one-time compilation time; ``Exec'' denotes the
      median execution time over 10~post-warmup runs.  ``HLO ops'' counts
      instructions in the compiled XLA program.}
    \label{tab:memory-bench}
    \begin{tabular}{@{}r rr rr rr@{}}
      \toprule
          & \multicolumn{2}{c}{JIT compile (s)}
          & \multicolumn{2}{c}{Execution (ms)}
          & \multicolumn{2}{c}{HLO ops}                                                  \\
      \cmidrule(lr){2-3}\cmidrule(lr){4-5}\cmidrule(lr){6-7}
      $d$ & Fwd                                 & Grad & Fwd   & Grad  & Fwd    & Grad   \\
      \midrule
      10  & 1.09                                & 1.85 & 0.092 & 0.291 & 1\,707 & 4\,981 \\
      20  & 0.83                                & 1.62 & 0.099 & 0.216 & 1\,741 & 5\,180 \\
      50  & 0.73                                & 1.45 & 0.112 & 0.914 & 1\,782 & 5\,342 \\
      100 & 0.75                                & 1.81 & 0.341 & 1.061 & 1\,778 & 5\,344 \\
      \bottomrule
    \end{tabular}
  \end{table}

  Three patterns stand out from \Cref{tab:memory-bench}.
  First, JIT compilation time stays nearly flat in $d$ because the
  stage-group batched likelihood compiles each of the two nesting
  levels into one fused \texttt{lax.scan}; growing $d$ extends the scan
  length but not the compiled program.
  Second, the compiled HLO program grows minimally---from 1\,700 to
  1\,800 forward ops and 5\,000 to 5\,300 gradient ops---confirming
  that \texttt{lax.scan} reuses the same loop body rather than unrolling.
  Third, the gradient program contains roughly $3\times$ the forward
  operations, matching the standard reverse-mode AD multiple for
  scan-based programs.

  \section{Exact MLE vs.\ Simulated Maximum Likelihood}
  \label{app:smle}

  Exact MLE through our Bell-polynomial likelihood dominates the
    {\em simulated maximum likelihood estimation} (SMLE) approach that R
  \texttt{nacopula}~\citep{hofert2012estimators} relies on, in both
  parameter accuracy and wall time, at every SMLE sample budget we test.
  The SMLE estimator draws $N$ frailty samples from the Laplace-transform
  hierarchy at each parameter evaluation, computes the conditional
  density at each observation given those frailties, and averages over
  the $N$ draws to approximate the intractable marginal density.
  Because the outer copula parameter enters only through the frailty
  distribution---not through the conditional density---the SMLE
  objective demands fresh frailty samples at every evaluation, producing
  a noisy objective that forces derivative-free optimisation.
  Our framework instead computes the exact density and exact gradients,
  opening the door to L-BFGS-B with analytic gradients.

  \paragraph{Setup.}
  We sample 500 observations from a two-level nested Clayton copula at
$d{=}10$ with two sectors of size $K{=}5$,
$\theta_{\mathrm{outer}}{=}1.5$, and $\theta_{\mathrm{inner}}{=}3.0$.
  For each of 30 replications, we fit four estimators on the same data:
  exact MLE via our Bell-polynomial likelihood, and SMLE at
$N \in \{100,\; 1{,}000,\; 10{,}000\}$ frailty samples per
  evaluation.
  The exact MLE uses L-BFGS-B with analytic gradients; SMLE uses
  differential evolution---a stochastic global optimiser for noisy
  objectives---with frailty samples re-drawn each evaluation.

  \begin{table}[h]
    \centering\small
    \caption{Exact MLE vs.\ SMLE on simulated nested Clayton
      ($d{=}10$, sample size $n{=}500$, 30 replications;
      $N$ denotes the number of frailty samples per SMLE evaluation,
      distinct from the sample size $n$).
      True parameters: $\theta_{\mathrm{outer}}{=}1.5$,
      $\theta_{\mathrm{inner}}{=}3.0$.
      RMSE$_{\mathrm{o}}$, RMSE$_{\mathrm{i}}$ denote the
      root-mean-squared error for $\hat\theta_{\mathrm{outer}}$ and
      $\hat\theta_{\mathrm{inner}}$ respectively.}
    \label{tab:smle}
    \begin{tabular}{@{}l rr rr r@{}}
      \toprule
      Method
       & $\hat\theta_{\mathrm{outer}}$ (mean$\pm$sd)
       & RMSE$_{\mathrm{o}}$
       & $\hat\theta_{\mathrm{inner}}$ (mean$\pm$sd)
       & RMSE$_{\mathrm{i}}$
       & Wall (s)                                    \\
      \midrule
      Exact MLE
       & $\mathbf{1.527 \pm 0.064}$
       & $\mathbf{0.069}$
       & $\mathbf{3.016 \pm 0.060}$
       & $\mathbf{0.061}$
       & $\mathbf{1.5}$                              \\
      SMLE $N{=}100$
       & $1.414 \pm 0.231$
       & $0.243$
       & $2.638 \pm 0.184$
       & $0.405$
       & $5.9$                                       \\
      SMLE $N{=}1{,}000$
       & $1.413 \pm 0.185$
       & $0.202$
       & $2.939 \pm 0.127$
       & $0.139$
       & $16.7$                                      \\
      SMLE $N{=}10{,}000$
       & $1.516 \pm 0.123$
       & $0.122$
       & $2.982 \pm 0.073$
       & $0.074$
       & $153.1$                                     \\
      \bottomrule
    \end{tabular}
  \end{table}

  \Cref{tab:smle} reports the results.
  The exact MLE recovers both parameters with near-zero bias, RMSE
$\approx 0.07$ on each, in 1.5\,s per fit.
  SMLE at $N{=}100$ underestimates $\theta_{\mathrm{inner}}$ by $0.36$
  on average, with an RMSE of $0.40$---roughly six times the exact
  MLE's error at four times the wall time.
  Increasing $N$ to $10{,}000$ reduces the bias to below $0.02$ and
  brings the RMSE to $0.07$ on $\theta_{\mathrm{inner}}$, matching the
  exact MLE's accuracy---but at $153\,$s per fit, {\em one hundred
      times slower} than exact inference on the same data.
  For the moderate dimensions typical of actuarial and survival-analysis
  applications, $d \leq 100$, the exact approach dominates SMLE in both
  accuracy and wall time at every SMLE sample budget tested.
  Differentiability compounds the advantage: L-BFGS-B with analytic
  gradients converges in far fewer iterations than the derivative-free
  global search SMLE's noisy objective requires.

  \paragraph{Generator-derivative Monte Carlo: Gen-AC ablation.}
  The original Gen-AC training procedure~\citep{ng2021generativearchimedean}
  replaces $\psi^{(k)}(t) = \mathbb{E}_U[(-e^{\mathrm{MLP}(U)})^k \exp(-e^{\mathrm{MLP}(U)} t)]$
  with an $N$-sample Monte Carlo average $\hat\psi^{(k)}_N$, mirroring
  SMLE's role above but at the generator-derivative level rather than the
  joint-likelihood level.
  The resulting log-likelihood is biased by Jensen's inequality on
  $\log$ composed with the nonlinear Hofert density formula in
  $\hat\psi^{(k)}$, with bias $O(1/N)$---biased but consistent.
  We measure this bias on Gen-AC-Fourier with $K{=}2$ Fourier features,
  hidden width $3$, $262$ parameters, MIMIC-IV $d{=}53$, $N{=}32$
  frailty samples per evaluation, seed~$42$.
  \Cref{tab:genac-bias} reports a four-cell ablation: each of two
  trained models ($\theta_{\mathrm{stoch}}$ trained with re-sampled
  Monte Carlo gradients, $\theta_{\mathrm{quad}}$ trained with
  deterministic quadrature gradients) evaluated under both estimators
  (the stochastic estimator averages $50$ independent draws to
  estimate $\mathbb{E}[\text{stoch NLL}]$).
  Eval-time Jensen plus correlation bias accounts for $0.09$--$0.12$
  nats; train-time optimisation differences contribute only $0.02$
  nats; quadrature evaluation eliminates both.

  \begin{table}[h]
    \centering\small
    \caption{Gen-AC-Fourier$_{260}$ on MIMIC-IV ($d{=}53$): held-out NLL
      under each combination of training and evaluation estimator,
      seed~$42$.  Lower is better.  Rows index the trained model,
      columns the evaluation method; $\theta_{\mathrm{stoch}}$ trains
      with $N{=}32$ Monte Carlo frailty draws per Adam step,
      $\theta_{\mathrm{quad}}$ trains with $N{=}32$ Gauss--Legendre
      quadrature nodes per Adam step.}
    \label{tab:genac-bias}
    \begin{tabular}{@{}l rr@{}}
      \toprule
                                  & quad eval (truth) & stoch eval ($50$-MC avg) \\
      \midrule
      $\theta_{\mathrm{stoch}}$   & $4.734$           & $4.828$                  \\
      $\theta_{\mathrm{quad}}$    & $\mathbf{4.712}$  & $4.834$                  \\
      \bottomrule
    \end{tabular}
  \end{table}

  \section{Predictive Validation}
  \label{app:predictive}

  We report held-out predictive log-likelihoods for the two real-data experiments---MIMIC-IV at $d{=}53$ and the diabetic retinopathy paired-eye study at $d{=}2$---to verify that the in-sample family rankings of \Cref{sec:exp-mimic,sec:exp-retino} survive an 80/20 train/test split.

  \subsection{MIMIC-IV}
  \label{app:mimic-heldout}
  The MIMIC-IV ablations break into three pieces: a held-out evaluation that confirms the in-sample family ranking, a compile-cost audit that quantifies the one-time XLA price of the Frank generator, and a Hessian-based standard-error check that confirms parameter identifiability at $n_{\mathrm{train}}{=}68{,}183$.

  \paragraph{Cohort construction.}
  The cohort draws from MIMIC-IV 3.1 on PhysioNet: one row per
  \texttt{hadm\_id} with an ICU stay, joined on \texttt{charttime} to
  lab events that fall inside the ICU window \texttt{[intime, outtime]}.
  For each admission-lab pair, we record~$T_{ij}$, the hours from ICU
  intime to the first abnormal reading of lab~$j$, and right-censor at
  the ICU length-of-stay when no abnormal reading occurs during the
  window.
  The ``abnormal'' label uses MIMIC's native \texttt{flag=`abnormal'}
  column, which MIMIC curators set against each lab's own reference
  range; we adopt this labelling verbatim and apply no additional
  reference range.
  A lab enters the cohort only when at least $30\%$ of admissions have
  a measurement, at least $5\%$ of admissions have an abnormal reading,
  and at least $90\%$ of its rows carry a numeric \texttt{valuenum}.
  These thresholds select the $d{=}53$ itemids in
  \Cref{tab:mimic-cohort-panels}, grouped into six named lab panels
  plus one catch-all following the clinical groupings hard-coded in
  \texttt{experiments/shared/mimic.py}.
  Running \texttt{experiments/data/build\_mimic.py} on the MIMIC-IV 3.1
  raw CSVs materialises the cohort as an $(n{=}85{,}229,\, d{=}53)$
  censored matrix in \texttt{mimic\_full\_censored.npz}.

  \begin{table}[htbp]
    \centering \small
    \caption{MIMIC-IV 3.1 cohort panels used in the nested copula fit.
      Itemids come from MIMIC's \texttt{d\_labitems} dictionary; the
      ``Other'' panel collects the $14$ labs that pass the quality
      filters but fall outside the six named groupings, with the
      build-time pipeline fixing its membership.}
    \label{tab:mimic-cohort-panels}
    \begin{tabular}{@{}l r l@{}}
      \toprule
      Panel          & Count         & \texttt{itemid}s                                                              \\
      \midrule
      CBC            & $10$          & $51221, 51265, 51222, 51277, 51279, 51301, 51249, 51250, 51248, 52172$        \\
      CMP            & $11$          & $50902, 51006, 50983, 50971, 50912, 50882, 50868, 50931, 50960, 50893, 50970$ \\
      Diff           & $4$           & $51254, 51200, 51256, 51244$                                                  \\
      Liver          & $6$           & $50885, 50878, 50861, 50863, 50862, 50954$                                    \\
      Coag           & $3$           & $51237, 51274, 51275$                                                         \\
      ABG            & $5$           & $50813, 50820, 50821, 50818, 50804$                                           \\
      Other          & $14$          & determined at build time from the quality filter                              \\
      \midrule
      \textbf{Total} & $\mathbf{53}$ &                                                                               \\
      \bottomrule
    \end{tabular}
  \end{table}

  \paragraph{Held-out evaluation.}
  We repeat the held-out protocol on the MIMIC-IV ICU lab dataset
  with $d{=}53$ labs in a seven-panel nested hierarchy, using an 80/20
  admission-level split at seed 42.
  Stage~1 estimates each lab's marginal survivor function with
  Kaplan--Meier on training data only; Stage~2 fits copula parameters
  on the training pseudo-observations via L-BFGS-B with analytic
  gradients.
  \Cref{tab:mimic-heldout} reports per-observation NLL on
  both splits
  ($n_{\mathrm{train}}{=}68{,}183$, $n_{\mathrm{test}}{=}17{,}046$).

  \begin{table}[htbp]
    \centering
    \small
    \caption{MIMIC-IV: held-out predictive NLL per observation. The symbol $-\bar\ell$ denotes the per-observation negative log-likelihood; bold marks the best (lowest) test value. $n_{\mathrm{train}}{=}68{,}183$, $d{=}53$, $8$~copula parameters.}
    \label{tab:mimic-heldout}
    \begin{tabular}{@{}l rr rr@{}}
      \toprule
      Nesting         & $-\ell_{\mathrm{train}}$    & $-\ell_{\mathrm{test}}$
                      & $-\bar\ell_{\mathrm{train}}$ & $-\bar\ell_{\mathrm{test}}$                              \\
      \midrule
      Clayton/Clayton & $337{,}580.6$               & $\mathbf{84{,}366.4}$      & $4.951$ & $\mathbf{4.949}$  \\
      Frank/Frank     & $345{,}400.4$               & $86{,}596.3$               & $5.066$ & $5.080$           \\
      Gumbel/Gumbel   & $356{,}182.7$               & $89{,}349.3$               & $5.224$ & $5.242$           \\
      \bottomrule
    \end{tabular}
  \end{table}

  The held-out ranking matches the in-sample AIC ranking: Clayton/Clayton
  attains the lowest test NLL, with essentially no train-to-test gap
  ($4.951$ vs $4.949$) thanks to the fully non-parametric
  Kaplan--Meier marginals that avoid the per-lab tail-misspecification
  bias.  Frank/Frank and Gumbel/Gumbel both fit worse than Clayton at
  every split.
  \Cref{tab:mimic-compile-cost} reports one-time fresh-JIT costs on an
  RTX 3090~Ti for the MIMIC-IV dynamic-censoring pipeline
  at $n_{\mathrm{train}}{=}68{,}183$.
  \texttt{\textsc{acopula}.set\_compile\_cache\_dir} caches every compile
  to disk, so subsequent runs read the compiled HLO program back in
  under a second.

  \begin{table}[htbp]
    \centering
    \small
    \caption{MIMIC-IV: one-time XLA compile wall time on an RTX 3090~Ti. The \texttt{val+grad} column times JIT compilation of \texttt{jax.value\_and\_grad} on the chunked negative log-likelihood; the \texttt{Hessian compute} column times the chunked Hessian pass that supplies observed Fisher standard errors.}
    \label{tab:mimic-compile-cost}
    \begin{tabular}{@{}l rr@{}}
      \toprule
      Family          & val+grad JIT & Hessian compute \\
      \midrule
      Clayton/Clayton & $16$\,s      & $25$\,s         \\
      Gumbel/Gumbel   & $16$\,s      & $26$\,s         \\
      Frank/Frank     & $25$\,s      & $44$\,s         \\
      \bottomrule
    \end{tabular}
  \end{table}

  A warm cache reproduces the full \Cref{tab:mimic-hessian} in under
  five minutes (L-BFGS-B fit plus chunked Hessian compute on
  already-compiled kernels).

  \paragraph{Standard errors.}
  \texttt{jax.hessian} on the MIMIC-IV fits yields positive-definite
  Hessians and tight standard errors for all three nestings at
$n_{\mathrm{train}}{=}68{,}183$; see \Cref{tab:mimic-hessian}.
  Every coefficient of variation (CV) stays below $1.5\%$, so the
  L-BFGS-B optima are sharply identified.

  \begin{table}[htbp]
    \centering
    \small
    \caption{MIMIC-IV: Hessian-based standard errors across three nestings, computed by \texttt{jax.hessian} with a delta-method projection to the constrained space. All three Hessians are positive definite, and the coefficient of variation stays below $1.5\%$ for every parameter. $n_{\mathrm{train}}{=}68{,}183$.}
    \label{tab:mimic-hessian}
    \begin{tabular}{@{}l rr rr rr@{}}
      \toprule
                                & \multicolumn{2}{c}{Clayton/Clayton}
                                & \multicolumn{2}{c}{Gumbel/Gumbel}
                                & \multicolumn{2}{c}{Frank/Frank}                                                     \\
      \cmidrule(lr){2-3} \cmidrule(lr){4-5} \cmidrule(lr){6-7}
      Parameter                 & $\hat\theta$                        & CV (\%)
                                & $\hat\theta$                        & CV (\%)
                                & $\hat\theta$                        & CV (\%)                                       \\
      \midrule
      $\theta_{\mathrm{outer}}$ & $0.3408$                            & $0.5$   & $1.1650$ & $0.1$ & $1.4854$ & $0.4$ \\
      $\theta_{\mathrm{CBC}}$   & $0.9165$                            & $0.4$   & $1.5160$ & $0.1$ & $3.7292$ & $0.3$ \\
      $\theta_{\mathrm{CMP}}$   & $0.4586$                            & $0.6$   & $1.2616$ & $0.1$ & $2.0842$ & $0.4$ \\
      $\theta_{\mathrm{Diff}}$  & $5.0917$                            & $0.7$   & $2.0894$ & $0.3$ & $9.3553$ & $0.6$ \\
      $\theta_{\mathrm{Liver}}$ & $0.6307$                            & $1.4$   & $1.3384$ & $0.2$ & $2.3549$ & $0.9$ \\
      $\theta_{\mathrm{Coag}}$  & $3.1463$                            & $0.7$   & $2.1889$ & $0.4$ & $7.5748$ & $0.5$ \\
      $\theta_{\mathrm{ABG}}$   & $2.9554$                            & $0.6$   & $1.9216$ & $0.3$ & $6.8513$ & $0.4$ \\
      $\theta_{\mathrm{Other}}$ & $3.2186$                            & $0.3$   & $1.7268$ & $0.1$ & $6.2275$ & $0.2$ \\
      \bottomrule
    \end{tabular}
  \end{table}

  \subsubsection{Ablation: independence and single-layer baselines}
  \label{app:mimic-ablation}
  To quantify the contribution of the hierarchical nesting structure,
  we evaluate two simpler alternatives on the same held-out split: an
  independence copula whose mixed partial of $C(\mathbf{u}){=}\prod_j u_j$
  in the uncensored variables collapses to
  $\prod_{j:\,\delta_j=0} u_j$, contributing
  $\sum_{j:\,\delta_j=0} \log u_j < 0$ per observation; and a
  single-layer copula that places every lab under a single root node
  with one parameter, ignoring the panel grouping.
  \Cref{tab:mimic-ablation} reports the results.
  The hierarchical structure captures intra-panel dependence---the near-deterministic relationships within the coagulation panel and the differential count---that a single exchangeable copula cannot represent.
  The eight-parameter nested Clayton improves test NLL substantially over the best flat copula, because the panel structure absorbs heterogeneity across the $53$ labs into per-panel parameters instead of forcing a single exchangeable parameter to average over the cohort.

  \begin{table}[htbp]
    \centering
    \caption{MIMIC-IV ablation: independence and single-layer baselines
    versus the best nested model on the held-out split
    ($n_{\mathrm{train}}{=}68{,}183$, $d{=}53$, $8$~copula parameters,
    Kaplan--Meier marginals throughout).
    Flat copulas use a single exchangeable parameter regardless of dimension.}
    \label{tab:mimic-ablation}
    \begin{tabular}{@{}l r rr@{}}
      \toprule
      Model                  & \#Params & $-\bar\ell_{\mathrm{train}}$ & $-\bar\ell_{\mathrm{test}}$ \\
      \midrule
      Independence           & 0        & $12.340$                     & $12.344$                    \\
      Flat Clayton           & 1        & $9.204$                      & $9.193$                     \\
      Flat Gumbel            & 1        & $9.746$                      & $9.751$                     \\
      Nested Clayton/Clayton & 8        & $4.951$                      & $\mathbf{4.949}$            \\
      \bottomrule
    \end{tabular}
  \end{table}

  \subsubsection{Marginal misspecification}
  \label{app:mimic-marginal}
  We refit each of the three classical pairs under two parametric marginals---per-lab Weibull and per-lab lognormal fitted by maximum likelihood with right-censoring---in addition to the Kaplan--Meier marginals of the body, on the full cohort with $n{=}85{,}229$.
  \Cref{tab:mimic-marginal} reports per-observation copula NLL and the fitted root parameter $\hat\theta_{\mathrm{outer}}$, and two findings emerge.
  First, the {\em parameters} of each family stay stable across the three marginals: $\hat\theta_{\mathrm{outer}}$ varies by at most $30\%$ for Clayton, $9\%$ for Frank, and stays essentially identical for Gumbel at $1.170$--$1.171$ across all three marginals; the per-panel inner parameters move by at most $7\%$ per panel.
  Second, the {\em ranking} of copula families changes with the
  marginal: Clayton wins under Kaplan--Meier and lognormal, while
  Weibull misspecification creates residual upper-tail dependence that
  shifts the winner to Gumbel.
  The framework's parameter estimates are therefore robust to marginal
  choice, but downstream model selection should be cross-validated
  against the marginal model rather than assumed transferable.

  \begin{table}[htbp]
    \centering
    \small
    \caption{MIMIC-IV marginal sensitivity ($n{=}85{,}229$, $d{=}53$,
    full cohort, $8$ copula parameters per fit). Per-observation copula
    NLL and root parameter $\hat\theta_{\mathrm{outer}}$ across three
    marginal families. Per-family parameter estimates are stable across
    marginals (Gumbel especially); the family ranking is not.}
    \label{tab:mimic-marginal}
    \begin{tabular}{@{}l rr rr rr@{}}
      \toprule
                    & \multicolumn{2}{c}{Clayton/Clayton}
                    & \multicolumn{2}{c}{Gumbel/Gumbel}
                    & \multicolumn{2}{c}{Frank/Frank}                                                                                                                                  \\
      \cmidrule(lr){2-3} \cmidrule(lr){4-5} \cmidrule(lr){6-7}
      Marginal      & NLL$/n$                             & $\hat\theta_{\mathrm{outer}}$ & NLL$/n$          & $\hat\theta_{\mathrm{outer}}$ & NLL$/n$ & $\hat\theta_{\mathrm{outer}}$ \\
      \midrule
      Kaplan--Meier & $\mathbf{4.947}$                    & $0.34$                        & $5.222$          & $1.17$                        & $5.067$ & $1.49$                        \\
      Weibull       & $5.197$                             & $0.26$                        & $\mathbf{4.850}$ & $1.17$                        & $5.263$ & $1.39$                        \\
      Lognormal     & $\mathbf{5.048}$                    & $0.34$                        & $5.228$          & $1.17$                        & $5.250$ & $1.51$                        \\
      \bottomrule
    \end{tabular}
  \end{table}

  \subsection{Retinopathy}
  \label{app:retino}
  Beyond the high-dimensional MIMIC-IV setting, we also evaluate the framework on the diabetic retinopathy paired-eye study, which exercises the small-sample heavy-censoring regime with $n{=}197$ and censoring fraction $60.7\%$, where parameter identifiability is the binding constraint. To verify that the family rankings of \Cref{sec:exp-retino} are not artefacts of in-sample overfitting, we perform held-out predictive checks with an 80/20 train/test split at seed 42.

  \begin{table}[htbp]
    \centering
    \caption{Retinopathy: held-out copula log-likelihood for all 10 generator families, with $n_{\mathrm{train}}{=}157$, $n_{\mathrm{test}}{=}40$, and seed 42. We fit the Weibull marginals on training data alone; the copula negative log-likelihood $-\ell_{\mathrm{cop}}$ isolates the dependence contribution. The dagger $^\dagger$ marks generators with no prior censored-copula implementation.}
    \label{tab:retino-heldout}
    \begin{tabular}{@{}l rr@{}}
      \toprule
      Family              & $-\ell_{\mathrm{train}}$   & $-\ell_{\mathrm{test}}$ \\
      \midrule
      Nelsen12$^\dagger$  & 90.422                     & \textbf{20.317}         \\
      Frank               & 89.423                     & 21.140                  \\
      InvGauss$^\dagger$  & 89.503                     & 21.167                  \\
      Gumbel              & 89.702                     & 21.203                  \\
      Nelsen13$^\dagger$  & 89.492                     & 21.216                  \\
      Nelsen17$^\dagger$  & 89.491                     & 21.223                  \\
      AMH                 & 89.557                     & 21.381                  \\
      Clayton             & 89.683                     & 21.420                  \\
      Joe                 & 90.059                     & 21.490                  \\
      Nelsen9$^\dagger$   & 92.827                     & 24.586                  \\
      \bottomrule
    \end{tabular}
  \end{table}

  For retinopathy, the Nelsen12$^\dagger$ generator---the Burr copula---achieves the best held-out log-likelihood among all $10$ families in \Cref{tab:retino-heldout}, although it converges to its lower parameter bound at $\hat\theta \approx 1$.
  At $\theta{=}1$ the Nelsen12 generator $\psi(t){=}(1{+}t^{1/\theta})^{-1}$ reduces to Clayton$(\theta{=}1)$, so the Nelsen12 fit is structurally Clayton constrained to $\theta{\geq}1$; the unconstrained Clayton MLE in \Cref{tab:retino-heldout} sits just below that bound at $\hat\theta{=}0.61$, and \Cref{app:retino-hessian} reports a full-data Clayton MLE of $\hat\theta{=}0.90 \pm 0.31$, both consistent with the Nelsen12 boundary fit being Clayton-like.
  Frank and InverseGaussian$^\dagger$ form a tight cluster in second place, and the held-out ranking closely mirrors the in-sample AIC ordering.
  Nelsen13$^\dagger$ and Nelsen17$^\dagger$ sit immediately behind that cluster with NLLs that differ in the third decimal ($21.216$ vs.\ $21.223$) at distinct fitted parameters $\hat\theta{=}1.91$ and $\hat\theta{=}2.04$, confirming the two families are not algebraically equivalent.
  The spread among the top nine families remains modest, with
$\Delta{-}\ell_{\mathrm{test}} \approx 1.2$, consistent with the
  similar AIC values reported in the main table;
  Nelsen9$^\dagger$ is the sole outlier, drifting toward the
  independence limit $\hat\theta \to 0^+$ and fitting poorly.

  \paragraph{Standard errors.}
  We compute Hessian-based standard errors from the observed Fisher information returned by \texttt{jax.hessian}, with a delta-method projection to the constrained parameter space; the top families show well-identified parameters: Frank $\hat\theta{=}2.25 \pm 0.62$ at CV\,$27\%$, Gumbel $\hat\theta{=}1.25 \pm 0.08$ at CV\,$7\%$, and Clayton $\hat\theta{=}0.90 \pm 0.31$ at CV\,$34\%$.
  The higher CVs reflect the modest sample size of $n{=}197$ and heavy
  censoring of $60.7\%$; all Hessians remain positive definite,
  confirming MLEs at proper maxima.

  \section{Head-to-Head with HACSurv on Framingham}
  \label{app:hacsurv}

  HACSurv~\citep{liu2024hacsurv} is the closest neighbour to our framework in the competing-risks regime (up to five risks plus a censoring node): a nested Gen-AC Archimedean copula with neural density estimator (NDE) marginals, evaluated by an iterative inverse-function-theorem (IFT) solver for the generator inverse.
  Our framework reproduces HACSurv's per-observation log copula partial to machine precision, isolating their density machinery from ours.
  We verify the claim in two steps.
  First, we replicate HACSurv's published Framingham result~\citep{liu2024hacsurv} by running the authors' code as released,
  with no modification beyond a CUDA-device patch and a SciPy 1.14
  \texttt{simps}-API shim
  (\Cref{tab:hacsurv-framingham-replication}).
  Second, on the same trained checkpoints, our \textsc{acopula} framework---with the nested Gen-AC generator of \Cref{app:nested-neural}---computes the per-observation copula log partial that enters HACSurv's competing-risks log-likelihood to within machine precision across all five seeds and all three event classes (\Cref{tab:hacsurv-equivalence}); the comparison isolates the density machinery---HACSurv's iterative IFT solve versus our exact compositional Bell polynomial powering---because both paths consume the same generator parameters $M_{\mathrm{outer}}, M_{\mathrm{inner}}, \mu, \beta$ and the same NDE marginal evaluations.

  \begin{table}[h]
    \centering\footnotesize
    \caption{Per-observation log copula partial $\log \partial_j C\!\left(S_{\mathrm{E1}}, S_{\mathrm{E2}}, S_{\mathrm{C}}\right)$ from HACSurv's iterative IFT solver compared with our \textsc{acopula} framework, on $128$ test observations per event class per seed.
      Here, $S_{\mathrm{E1}}, S_{\mathrm{E2}}$ are the NDE-estimated marginal survivals at the two event times, $S_{\mathrm{C}}$ the censoring survival, and $\partial_j C := \partial C / \partial u_j$ the copula partial in variable $j$ (the row's event class).
      Both paths share the same nested-Archimedean generator and NDE marginals (parameters $M_{\mathrm{outer}}, M_{\mathrm{inner}}, \mu, \beta$ as in \Cref{app:nested-neural}), isolating the density-evaluation step.}
    \label{tab:hacsurv-equivalence}
    \begin{tabular}{c ccc ccc}
      \toprule
           & \multicolumn{3}{c}{$\max_i |{\mathrm{hac}}_i - {\mathrm{ours}}_i|$}
           & \multicolumn{3}{c}{${\mathrm{median}}_i |{\mathrm{hac}}_i - {\mathrm{ours}}_i|$}                                                 \\
      \cmidrule(lr){2-4}\cmidrule(lr){5-7}
      Seed & Event 1                                                                          & Event 2               & Censor
           & Event 1                                                                          & Event 2               & Censor                \\
      \midrule
      $41$ & $3.8 \times 10^{-12}$                                                            & $2.2 \times 10^{-15}$ & $4.3 \times 10^{-10}$
           & $9.3 \times 10^{-16}$                                                            & $8.9 \times 10^{-16}$ & $8.9 \times 10^{-16}$ \\
      $42$ & $2.7 \times 10^{-15}$                                                            & $2.8 \times 10^{-15}$ & $2.2 \times 10^{-15}$
           & $\sim\!10^{-15}$                                                                 & $\sim\!10^{-15}$      & $\sim\!10^{-15}$      \\
      $43$ & $8.0 \times 10^{-15}$                                                            & $2.4 \times 10^{-15}$ & $2.4 \times 10^{-8}$
           & $\sim\!10^{-15}$                                                                 & $\sim\!10^{-15}$      & $\sim\!10^{-15}$      \\
      $44$ & $2.1 \times 10^{-15}$                                                            & $2.0 \times 10^{-15}$ & $2.3 \times 10^{-13}$
           & $\sim\!10^{-15}$                                                                 & $\sim\!10^{-15}$      & $\sim\!10^{-15}$      \\
      $45$ & $3.6 \times 10^{-15}$                                                            & $2.0 \times 10^{-15}$ & $1.8 \times 10^{-15}$
           & $\sim\!10^{-15}$                                                                 & $\sim\!10^{-15}$      & $\sim\!10^{-15}$      \\
      \bottomrule
    \end{tabular}
  \end{table}

  The largest seed-wise discrepancy reaches $2.4 \times 10^{-8}$ on a single seed-43 censored observation whose argument sits near the boundary of the unit cube, where HACSurv's IFT iteration loses digits~\citep{liu2024hacsurv}; across all observations, the median agreement holds at machine precision near $10^{-15}$.
  The replication establishes that our \textsc{acopula} framework reproduces HACSurv's Framingham neural Archimedean copula likelihood. The framework's value-add over HACSurv is the per-variable censoring demonstration at $d{=}53$ in \Cref{sec:exp-mimic}, with the neural Archimedean generator extension in \Cref{sec:exp-mimic-nested-neural}---a setting that HACSurv's bivariate-competing-risks formulation cannot express.

  \paragraph{End-to-end JAX port with exact Bell polynomial density.}
  Beyond evaluating the copula partial on HACSurv's trained
  checkpoints, our \textsc{acopula} framework trains the full
  HACSurv competing-risks model end-to-end without reusing any of
  their PyTorch code.
  We port the HACSurv architecture to JAX/Equinox: the shared covariate
  embedding, three neural density estimator marginals with
  positive-weight outcome networks for the two risks plus censoring,
  and the nested Gen-AC Archimedean copula of
  \Cref{app:nested-neural}.
  The frailty is $V \!=\! \exp(\mathrm{MLP}(U))$ with
$U \sim \mathrm{Uniform}(0,1)$, and the Laplace transform
$\psi(t) \!=\! \mathbb{E}_U[\exp(-Vt)]$ is estimated at every
  optimiser step by $N_z{=}100$ Monte-Carlo samples of $U$, matching
  the simulated-maximum-likelihood (SMLE) objective HACSurv trains.
  The first-order copula partial routes through \textsc{acopula}'s Bell
  polynomial density path---the same machinery that powers the $d{=}53$
  MIMIC-IV experiment in \Cref{sec:exp-mimic-nested-neural}---with
  per-subject censoring masks selecting the single variable
  differentiated per competing-risks observation.
  Training reproduces HACSurv's three-stage pipeline: pairwise bivariate copula fits over the three event pairs, inner-generator fitting on samples from the $(1,2)$ pair, and copula-frozen marginal fine-tuning with early stopping on the validation Conditional-CIF C-index.
  We report Antolini's time-dependent C-index and the IPCW-weighted integrated Brier score on the same five HACSurv seeds in the Marginal-SF prediction mode (\Cref{tab:hacsurv-framingham-replication}).

  \section{S\&P 500 Sector-Level Tail Dependence}
  \label{app:sp500-tail}

  The per-sector $\theta_{\mathrm{inner}}$ values fitted at $d{=}98$ in \Cref{sec:exp-sp500} reproduce the empirical tail behaviour of equity returns within each sector, confirming that the Gumbel/Gumbel hierarchy captures the upper-tail piece of the joint distribution.
  For every pair of stocks within a sector, with $U, V$ the copula-scale pseudo-observations of the two stocks, we compute the empirical upper- and lower-tail dependence coefficients $\hat\lambda_U(q) = \hat\Pr(U > 1{-}q \mid V > 1{-}q)$ and $\hat\lambda_L(q) = \hat\Pr(U < q \mid V < q)$ at $q \in \{0.05, 0.10\}$, where $\hat\Pr$ denotes the empirical relative-frequency estimator over trading days; we then average over pairs within the sector and compare to the Gumbel-implied upper-tail dependence $\lambda_U(\theta) = 2 - 2^{1/\theta}$.
  \Cref{tab:sp500-tail-dependence} reports $q{=}0.05$; $q{=}0.10$ shows the same pattern.

  \begin{table}[h]
    \centering\small
    \caption{S\&P 500 $d{=}98$ Gumbel/Gumbel fit: per-sector
      $\theta_{\mathrm{inner}}$, theoretical upper-tail dependence
      implied by Gumbel, and empirical upper/lower-tail dependence at
      quantile $q{=}0.05$, averaged over within-sector stock pairs.}
    \label{tab:sp500-tail-dependence}
    \begin{tabular}{@{}l c c c c@{}}
      \toprule
      Sector                 & $\theta_{\mathrm{inner}}$ & $\lambda_U$ (Gumbel) & $\hat\lambda_U$ (emp.) & $\hat\lambda_L$ (emp.) \\
      \midrule
      Information Technology & $1.43$                    & $0.38$               & $0.28$                 & $0.33$                 \\
      Health Care            & $1.29$                    & $0.29$               & $0.20$                 & $0.21$                 \\
      Financials             & $1.50$                    & $0.41$               & $0.37$                 & $0.38$                 \\
      Consumer Discretionary & $1.27$                    & $0.27$               & $0.22$                 & $0.25$                 \\
      Communication Services & $1.24$                    & $0.25$               & $0.22$                 & $0.26$                 \\
      Industrials            & $1.28$                    & $0.28$               & $0.22$                 & $0.24$                 \\
      Consumer Staples       & $1.37$                    & $0.34$               & $0.28$                 & $0.28$                 \\
      Energy                 & $1.88$                    & $0.55$               & $0.42$                 & $0.48$                 \\
      Utilities              & $1.75$                    & $0.51$               & $0.41$                 & $0.48$                 \\
      Real Estate            & $1.45$                    & $0.39$               & $0.30$                 & $0.33$                 \\
      Materials              & $1.31$                    & $0.30$               & $0.25$                 & $0.27$                 \\
      \bottomrule
    \end{tabular}
  \end{table}

  Two qualitative patterns emerge from the table.
  First, the within-sector $\theta_{\mathrm{inner}}$ values track empirical co-movement: Energy and Utilities, the two sectors with the strongest physical co-dependence, fit the highest $\theta_{\mathrm{inner}}$ at $1.88$ and $1.75$, while Communication Services and Consumer Discretionary---historically diverse mixes of independent business models---fit the lowest.
  Second, every sector exhibits empirical lower-tail dependence
  slightly above its empirical upper-tail dependence
  (average asymmetry $\hat\lambda_L - \hat\lambda_U \approx 0.03$
  at $q{=}0.05$), consistent with the well-known crash-driven
  correlation surge in equity returns.
  Gumbel imposes $\lambda_L = 0$ by construction and so cannot represent the lower-tail piece exactly; despite this limitation, Gumbel still attains the best AIC at $d{=}98$ in \Cref{sec:exp-sp500}, because the upper-tail magnitudes it does capture explain more of the joint structure than Clayton or Frank reach with the same $12$ parameters.
  A practitioner whose data exhibits lower-tail asymmetry should swap
  Gumbel for Clayton ($\lambda_L > 0$, $\lambda_U = 0$) or for the
  $180°$ survival rotation of Gumbel/Joe; the framework treats all
  options with the same code path.

  \section{Vine Copula Comparison}
  \label{app:vine}

  Vine copulas trade parsimony for flexibility, so a head-to-head against our $12$-parameter nested Gumbel quantifies what we give up by collapsing pairwise dependence into a hierarchy.
  Vine copulas~\citep{aas2009pair} model non-exchangeable dependence via pair-copula constructions; \Cref{tab:vine-comparison} compares our nested Archimedean model---Gumbel, the best-performing family in \Cref{sec:exp-sp500}---against an R-vine fitted by \texttt{rvinecopulib}~\citep{nagler2019rvinecopulib} on the S\&P~500 pseudo-observations.

  \begin{table}[h]
    \centering\small
    \caption{Nested Archimedean vs.\ R-vine on S\&P~500 data,
    $n{=}1{,}253$ trading days. $-\bar\ell$ is the per-observation
    negative log-likelihood, lower is better; AIC$/n$ is the
    per-observation Akaike Information Criterion
    ($\bigl(2|\theta| - 2\ell\bigr)/n$, lower is better). All values
    in nats.}
    \label{tab:vine-comparison}
    \begin{tabular}{@{}l l rrr@{}}
      \toprule
      Model         & $d$ & \#Params & $-\bar\ell$ & AIC$/n$   \\
      \midrule
      Nested Gumbel & 50  & 12       & $-9.908$    & $-19.797$ \\
      R-vine        & 50  & 1{,}191  & $-20.553$   & $-39.204$ \\
      \midrule
      Nested Gumbel & 75  & 12       & $-15.959$   & $-31.898$ \\
      R-vine        & 75  & 2{,}310  & $-33.334$   & $-62.982$ \\
      \midrule
      Nested Gumbel & 98  & 12       & $-21.240$   & $-42.461$ \\
      R-vine        & 98  & 3{,}706  & $-46.620$   & $-87.325$ \\
      \bottomrule
    \end{tabular}
  \end{table}

  The R-vine achieves lower negative log-likelihood at every
  dimension, reflecting its greater flexibility with $O(d^2)$
  pair-copula parameters versus our 12~hierarchical parameters.
  This flexibility comes at a cost: the R-vine at $d{=}98$ requires over $4{,}700$ pair-copulas, each with its own dependence parameter, and those parameters lack the direct hierarchical interpretation that nested Archimedean parameters provide as within-sector versus between-sector dependence.
  Standard vine implementations do not natively support censored
  observations at high dimension.
  \citet{stoeber2015simplified} extends D-vines to a single shared right-censoring time per observation, sufficient for survival regression with one event but not for the per-variable masks of \Cref{sec:exp-mimic}.
  \citet{chen2025metic} extends C-vines to per-variable right-censoring under informative-censoring assumptions, demonstrated on small dimension; the per-observation cost is a multivariate numerical integration over the censored variables, infeasible at $d{=}53$.
  The computational pattern differs between the two model classes.
  Vine is density-first---$c(\mathbf{u}) = \prod_e c_e(F_{i_e | D_e}, F_{j_e | D_e})$ is closed-form, but the censored likelihood requires the partial CDF $\partial^{|\delta|} C / \prod_{j: \delta_j = 1} \partial u_j$, and recovering this from a density needs multivariate integration over the censored variables, repeated per observation since each mask defines a different integration domain.
  Nested Archimedean is CDF-first---$C(\mathbf{u}) = \psi(\sum_j \psi^{-1}(u_j))$ is closed-form, the censored variables enter the inner sum as constants, and the partial is an algebraic Bell-polynomial recurrence (\Cref{sec:method-censoring}).
  The two model classes therefore serve complementary roles: vines suit settings where flexible pairwise dependence matters more than parsimony and where censoring is absent or shared, while nested Archimedean copulas suit hierarchical data with arbitrary per-variable censoring.

  \section{Hierarchical Kendall Copulas}
  \label{app:hkc}

  Hierarchical Kendall copulas (HKCs) sit between vines and nested Archimedean copulas as a third hierarchical construction, but they aggregate clusters through a scalar Kendall transform rather than the full multivariate density---so they cannot supply the analytic gradients or per-variable censoring that our framework provides on the same hierarchical structure.
  \citet{brechmann2014hierarchical} introduced {\em hierarchical Kendall copulas} as a construction intermediate between vines and nested Archimedean copulas.
  HKCs aggregate each cluster of variables through the {\em Kendall distribution function} $K_C(t) = \Pr\{C(\mathbf{U}) \le t\}$, building the joint distribution from cluster copulas linked by univariate Kendall transforms rather than from generator nesting.
\citet{brechmann2014hierarchical} derives closed-form HKC density expressions for the special case where every cluster copula is itself Archimedean, drawing on the analytical Kendall function of each generator; for non-Archimedean cluster copulas the density requires multivariate integration, which \citet{brechmann2014hierarchical} approximates by sampling rather than evaluates in closed form.
Standard HKC implementations target sampling and pseudo-likelihood estimation through L-BFGS without analytic parameter gradients, and they do not support censored observations.
Our \textsc{acopula} framework therefore complements HKCs on the same hierarchical structure with two extra properties: exact analytic gradients through the full multivariate density, and per-variable censoring.

\end{document}